\documentclass{article}

\PassOptionsToPackage{numbers, compress}{natbib}

\usepackage[preprint]{neurips_2026}

\usepackage[utf8]{inputenc} 
\usepackage[T1]{fontenc}    
\usepackage[colorlinks=true]{hyperref}    
\usepackage{url}            
\usepackage{booktabs}       
\usepackage{amsfonts}       
\usepackage{nicefrac}       
\usepackage{microtype}      
\usepackage[dvipsnames]{xcolor}         
\usepackage{pgfplots}

\usepackage{graphicx}
\usepackage{subcaption}
\usepackage{wrapfig}
\usepackage{algorithm2e}
\usepackage{bbm}
\RestyleAlgo{ruled}
\usepackage{subcaption} 

\pgfplotsset{compat=1.18}
\definecolor{linkblue}{HTML}{505DA0}  
\definecolor{citemauve}{HTML}{655390}                           
\definecolor{darkred}{RGB}{130, 40, 40}
\definecolor{darkgreen}{RGB}{40, 100, 40}
                     
\definecolor{swmcyan}{rgb}{0,0.7,0.9}
\hypersetup{                                                                                                    
linkcolor=linkblue,                                                                                             
urlcolor=linkblue,                                                                                              
citecolor=linkblue,                                   
}                                                                                                 

\usepackage{listings}

\usepackage{amsmath,amsfonts,bm}

\def\eqref#1{equation~\ref{#1}}

\def\1{\bm{1}}

\def\va{{\bm{a}}}

\def\vo{{\bm{o}}}

\def\vs{{\bm{s}}}

\DeclareMathAlphabet{\mathsfit}{\encodingdefault}{\sfdefault}{m}{sl}
\SetMathAlphabet{\mathsfit}{bold}{\encodingdefault}{\sfdefault}{bx}{n}

\newcommand{\R}{\mathbb{R}}

\DeclareMathOperator*{\argmin}{arg\,min}

\definecolor{dkgreen}{rgb}{0,0.6,0}
\definecolor{gray}{rgb}{0.5,0.5,0.5}
\definecolor{mauve}{rgb}{0.58,0,0.82}

\usepackage[most]{tcolorbox}
\usepackage{xparse}
\usepackage{framed}

\makeatletter

\@ifundefined{c@algocf}{\newcounter{algocf}}{}

\@ifundefined{SetCommentSty}{}{\SetCommentSty{texttt}}

\newlength{\BalgRuleThickness}
\newlength{\BalgBeforeSkip}
\newlength{\BalgAfterSkip}
\newlength{\BalgCaptionRuleSkip}
\newlength{\BalgBodyTopSkip}
\newlength{\BalgBodyBottomSkip}
\newlength{\BalgIndent}
\newcommand{\balg@rule}{%
  \par\nointerlineskip
  \hrule height \BalgRuleThickness
  \par\nointerlineskip
}

\newcommand{\balg@commentsty}[1]{%
  \@ifundefined{CommentSty}{%
    \texttt{#1}%
  }{%
    \CommentSty{#1}%
  }%
}

\newenvironment{balgbar}{%
  \par\begingroup
  \setlength{\FrameSep}{0pt}%
  \setlength{\OuterFrameSep}{0pt}%
  \MakeFramed{%
    \advance\hsize-\width
    \FrameRestore
    \setlength{\parindent}{0pt}%
    \setlength{\parskip}{0pt}%
  }%
}{%
  \endMakeFramed
  \endgroup\par
}

\newenvironment{breakablealgorithm}[1][]{%
  \par\vspace{\BalgBeforeSkip}%
  \refstepcounter{algocf}%
  \begin{tcolorbox}[
    enhanced,
    breakable,
    blanker,       
    left=0pt,
    right=0pt,
    top=0pt,
    bottom=0pt,
    boxsep=0pt,
    before skip=0pt,
    after skip=0pt,
    #1
  ]%
  \begingroup
  \setlength{\parindent}{0pt}%
  \setlength{\parskip}{0pt}%

  \RenewDocumentCommand{\caption}{O{} +m}{%
    \balg@rule
    \vspace{\BalgCaptionRuleSkip}%
    \noindent\textbf{Algorithm~\thealgocf: }##2\par
    \vspace{\BalgCaptionRuleSkip}%
    \balg@rule
    \vspace{\BalgBodyTopSkip}%
  }%

  \let\DontPrintSemicolon\relax

  \let\KwIn\relax
  \NewDocumentCommand{\KwIn}{+m}{%
    \noindent\textbf{Input:} ##1\par
  }%

  \let\KwOut\relax
  \NewDocumentCommand{\KwOut}{+m}{%
    \noindent\textbf{Output:} ##1\par
  }%

  \let\KwTo\relax
  \NewDocumentCommand{\KwTo}{}{%
    \ifmmode
      \mathrel{\mathrm{to}}%
    \else
      \ \textbf{to}\ %
    \fi
  }%

  \let\BlankLine\relax
  \NewDocumentCommand{\BlankLine}{}{%
    \par\vspace{0.45em}%
  }%

  \let\Return\relax
  \NewDocumentCommand{\Return}{+m}{%
    \par\noindent\textbf{return} ##1\par
  }%

  \let\For\relax
  \NewDocumentCommand{\For}{m +m}{%
    \par\noindent\textbf{for} ##1 \textbf{do}\par
    \begin{balgbar}##2\end{balgbar}%
  }%

  \let\While\relax
  \NewDocumentCommand{\While}{m +m}{%
    \par\noindent\textbf{while} ##1 \textbf{do}\par
    \begin{balgbar}##2\end{balgbar}%
  }%

  \let\If\relax
  \NewDocumentCommand{\If}{m +m}{%
    \par\noindent\textbf{if} ##1 \textbf{then}\par
    \begin{balgbar}##2\end{balgbar}%
  }%

  \let\tcp\relax
  \NewDocumentCommand{\tcp}{s O{r} +m}{%
    \IfBooleanTF{##1}{%
      \ifhmode
        \hfill\balg@commentsty{// ##3}%
      \else
        \noindent\balg@commentsty{// ##3}%
      \fi
    }{%
      \ifhmode
        \hfill\balg@commentsty{// ##3}%
        \par\noindent\ignorespaces
      \else
        \noindent\balg@commentsty{// ##3}%
        \par\noindent\ignorespaces
      \fi
    }%
  }%

  \let\tcc\relax
  \NewDocumentCommand{\tcc}{+m}{%
    \par\noindent\balg@commentsty{/* ##1 */}%
    \par\noindent\ignorespaces
  }%

  \let\balg@origsemi\;
  \def\;{%
    \ifmmode
      \balg@origsemi
    \else
      \par\noindent\ignorespaces
    \fi
  }%
}{%
  \par\vspace{\BalgBodyBottomSkip}%
  \balg@rule
  \endgroup
  \end{tcolorbox}
  \par\vspace{\BalgAfterSkip}%
}

\makeatother
\title{\texttt{stable-worldmodel}: A Platform for Reproducible\\World Modeling Research and Evaluation}

\author{%
Lucas Maes\textnormal{*\textsuperscript{1}}~Quentin Le Lidec\textnormal{*\textsuperscript{2}}~Luiz Facury\textnormal{\textsuperscript{3}}~Nassim Massaudi\textnormal{\textsuperscript{4}}~Ayush Chaurasia\textnormal{\textsuperscript{5}}\\
\textbf{
~Francesco Capuano\textnormal{\textsuperscript{6}}~Richard Gao\textnormal{\textsuperscript{7}}~Taj Gillin\textnormal{\textsuperscript{7}}~Dan Haramati\textnormal{\textsuperscript{7}}~Damien Scieur\textnormal{\textsuperscript{1}}}\\
\textbf{~Yann LeCun\textnormal{\textsuperscript{2}}~Randall Balestriero\textnormal{\textsuperscript{7}}}
\\[4pt]
$^{1}$Mila \& Université de Montréal~~$^{2}$New York University~~$^{3}$Universidade Federal de Minas Gerais\\
~~$^{4}$Independent Researcher~~$^{5}$LanceDB~~$^{6}$University of Oxford~~$^{7}$Brown University}

\usepackage{cleveref}

\begin{document}

\renewcommand{\thefootnote}{}
\footnotetext{* Equal contribution. Correspondence to \texttt{lucas.maes@mila.quebec}}
\renewcommand{\thefootnote}{\arabic{footnote}}

\maketitle

\vspace{-0.5em}
\begin{abstract}
World models are central to building agents that can reason, plan, and generalize beyond their training data. However, research on world models is currently fragmented, with disparate codebases, data pipelines, and evaluation protocols hindering reproducibility and fair comparison. Current practice is further limited by three key bottlenecks: fragile one-off codebases, slow video data loading, and the lack of standardized generalization benchmarks. We present \texttt{stable-worldmodel} (\texttt{swm}), an open-source platform for standardized and reproducible world modeling research and evaluation. It delivers (1) a high-performance Lance-based data layer with native support and conversion tools for MP4, HDF5, and LeRobot datasets, (2) clean, well-tested implementations of modern world model baselines and planning solvers, and (3) a broad suite of environments and tasks extended with controllable visual, geometric, and physical factors of variation for systematic in-silico evaluation of dynamics understanding, control performance, representation quality, and out-of-distribution generalization. By unifying the full pipeline under a single, scalable framework, \texttt{swm} dramatically reduces research overhead and accelerates trustworthy progress toward reliable world models. Code available \href{https://github.com/galilai-group/stable-worldmodel}{here}.
\end{abstract}

\section{Introduction}

\begin{wrapfigure}{r}{0.5\textwidth}
    \vspace{-4.6em}
    \centering
    \includegraphics[width=0.5\textwidth]{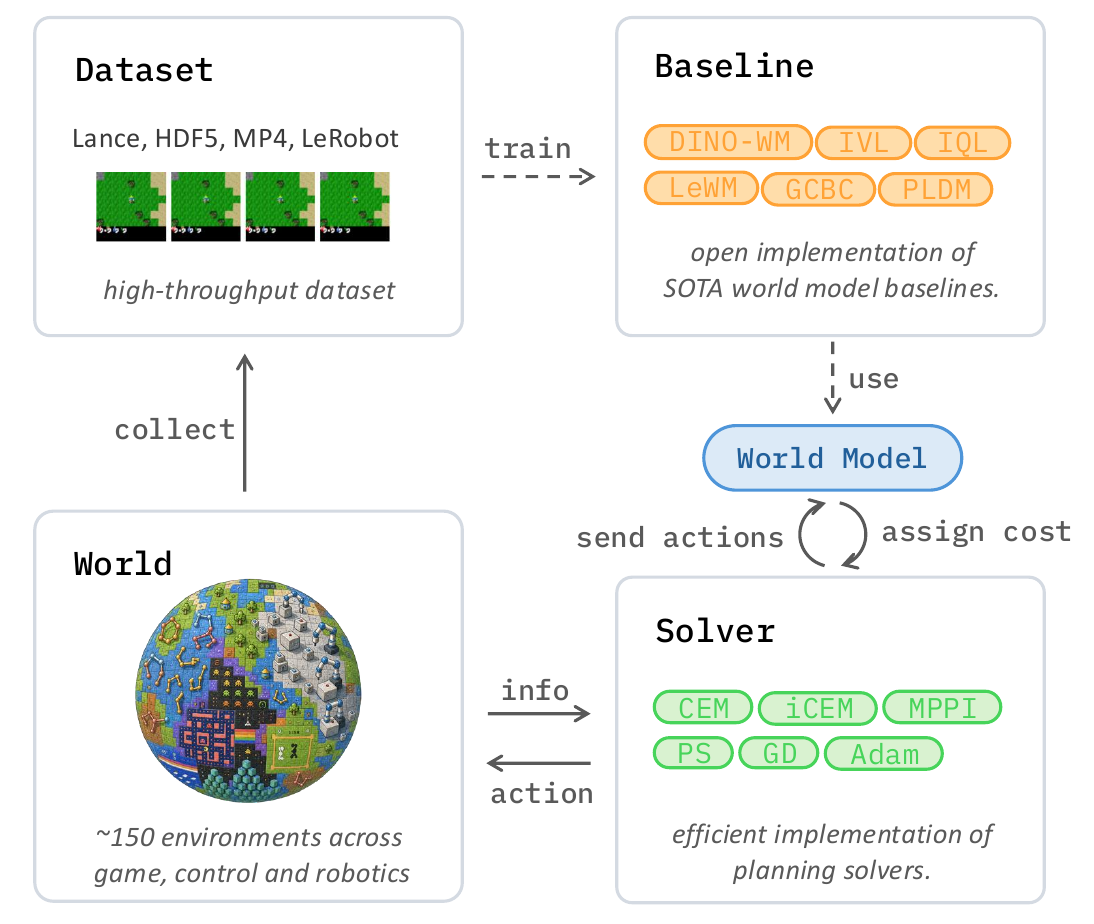}
    \caption{Overview of \texttt{stable-worldmodel}: data is efficiently collected from a world and used to train world models via provided baselines, then leveraged by solvers for control.}
    \label{fig:swm}
\end{wrapfigure}

The idea of using predictive models to guide decision-making dates back to the 1960s-1970s from the control theory community~\cite{propoi1963use,richalet1993industrial}. Most of these approaches relied on analytical, closed-form models or hand-crafted simulators to predict the outcome of actions (control variables)~\cite{kouvaritakis2016model,rawlings2020model}. Yet, recent advances in deep learning have instead learned to build these predictive models directly from raw data, leveraging neural networks ~\cite{ha2018world, hafner2019learning, hafner2023mastering,hansen2023td, bardes2023v}, an approach more commonly referred to as \textit{world models}. These world models enable the capture of far more complex real-world phenomena~\cite{assran2025v, goswami2025world} without deriving dynamics equations explicitly. However, as for any learned system, strengths and weaknesses audits before real-world deployment are paramount. Yet, despite exciting progress, the world modeling community still lacks reliable baselines, standardized benchmarks, and reproducible evaluation protocols. Most research relies on custom, often fragile codebases, forcing researchers to repeatedly re-implement the same algorithms. As the history of computer science has repeatedly shown, this approach is a recipe for hidden bugs, inconsistent results, leading to reduced credibility and trustworthiness in the results reported in publications~\cite{brooks1995mythical,raff2019step,islam2017reproducibility}.

To address these challenges, we present \texttt{stable-worldmodel} (\texttt{swm}), an open-source platform for reproducible world modeling research and evaluation. Built on PyTorch~\citep{paszke2019pytorch} and Gymnasium~\citep{towers2024gymnasium}, \texttt{swm} provides a complete, modular test-bed that supports researchers across the entire world model pipeline: from dataset collection and training utilities to comprehensive evaluation (including control, representation probing, out-of-distribution, and zero-shot generalization). It enables systematic assessment of new algorithms, clear identification of \textit{in-silico} model limitations, and fair comparisons against strong baselines.

Our contributions are the following:
\begin{itemize}
\setlength{\itemsep}{0pt}
\setlength{\parsep}{0pt}
\setlength{\topsep}{0pt}
\setlength{\partopsep}{0pt} 
\item A high-performance Lance-based data layer with native support and conversion tools for MP4, HDF5, and LeRobot datasets, eliminating I/O bottlenecks in multimodal world model training.
\item Clean, well-tested, and reproducible implementations of modern world model baselines (DINO-WM, LeWorldModel, PLDM, TD-MPC2, etc.) together with efficient and reliable implementations of planning solvers (CEM, MPPI, GD, etc.).
\item A diverse benchmarking suite spanning classic control, MuJoCo, Atari, robotics, and open-world environments, augmented with systematic controllable factors of variation (visual, geometric, and physical) enabling standardized zero-shot generalization and robustness evaluation.
\end{itemize}

\begin{figure}[tbp]
    \centering
    \begin{subfigure}[b]{0.195\textwidth}
        \centering
        \includegraphics[width=\textwidth]{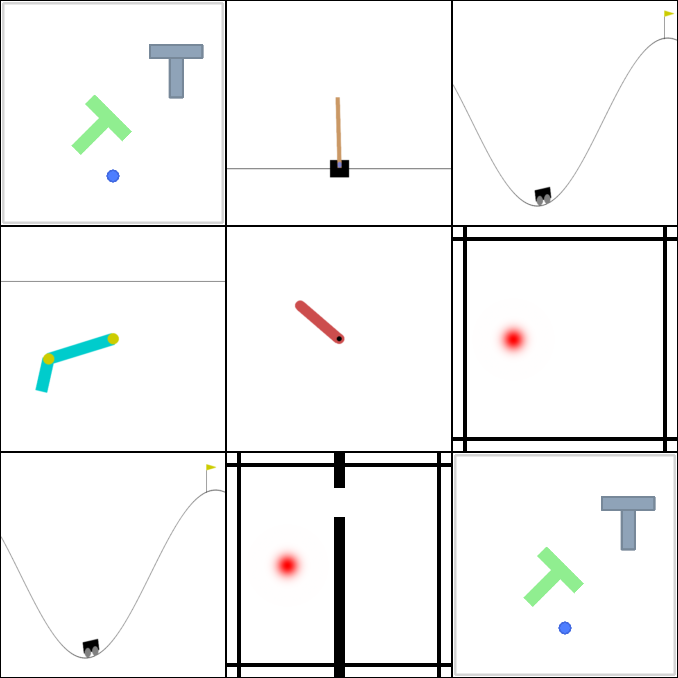}
    \end{subfigure}
    \hfill
    \begin{subfigure}[b]{0.195\textwidth}
        \centering
        \includegraphics[width=\textwidth]{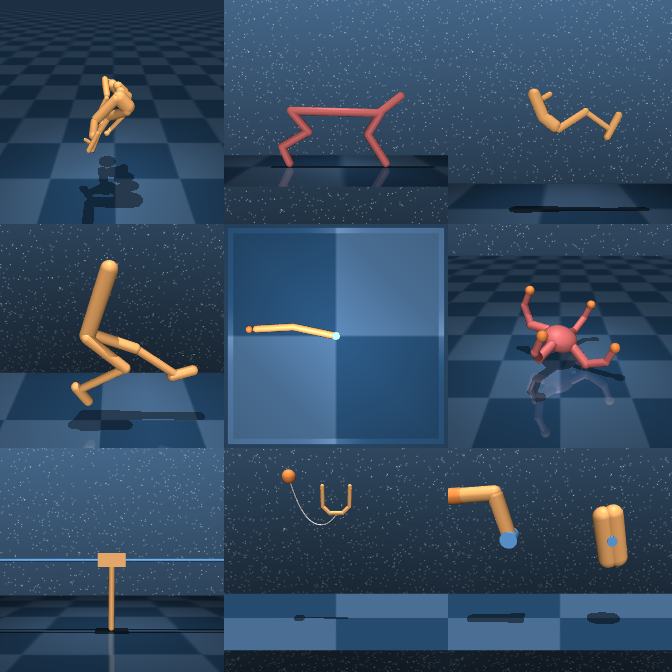}
    \end{subfigure}
    \hfill
    \begin{subfigure}[b]{0.195\textwidth}
        \centering
        \includegraphics[width=\textwidth]{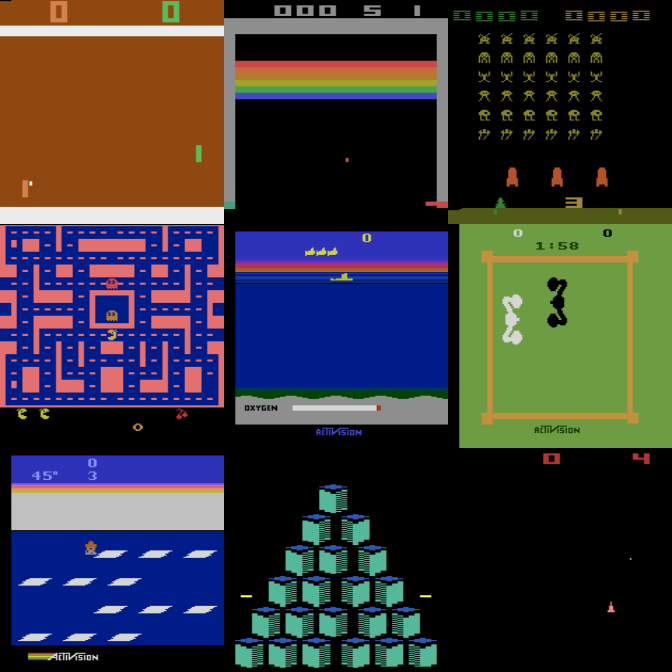}
    \end{subfigure}
    \hfill
    \begin{subfigure}[b]{0.195\textwidth}
        \centering
        \includegraphics[width=\textwidth]{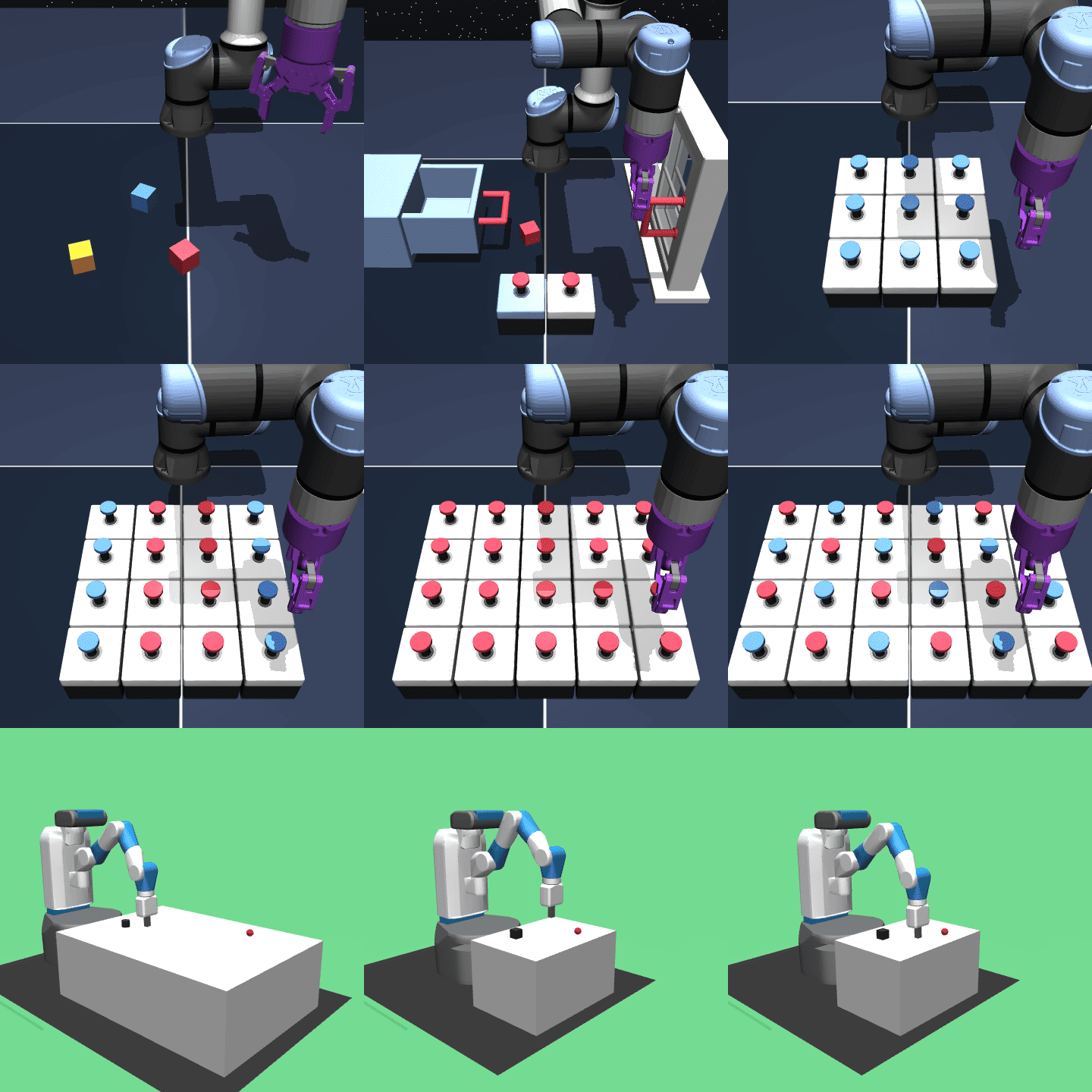}
    \end{subfigure}
    \hfill
    \begin{subfigure}[b]{0.196\textwidth}
        \centering
        \includegraphics[width=\textwidth]{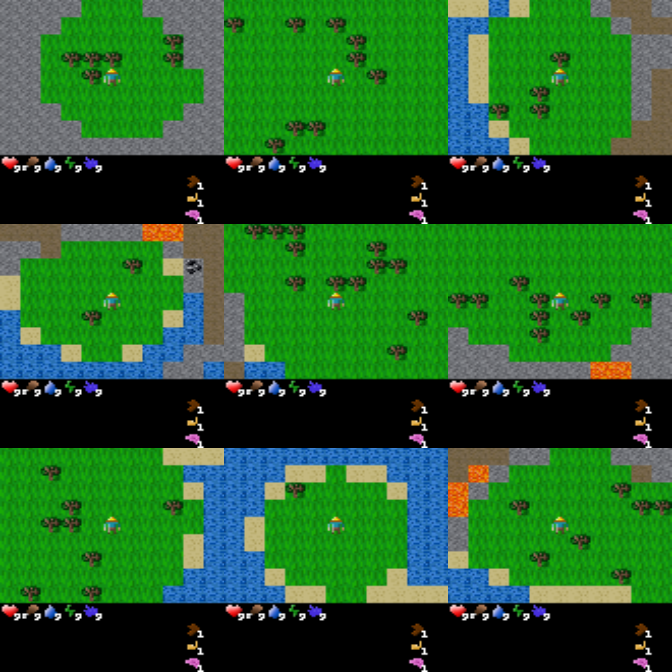}
    \end{subfigure}

    \vspace{0.10em}
    
    \begin{subfigure}[b]{0.195\textwidth}
        \centering
        \includegraphics[width=\textwidth]{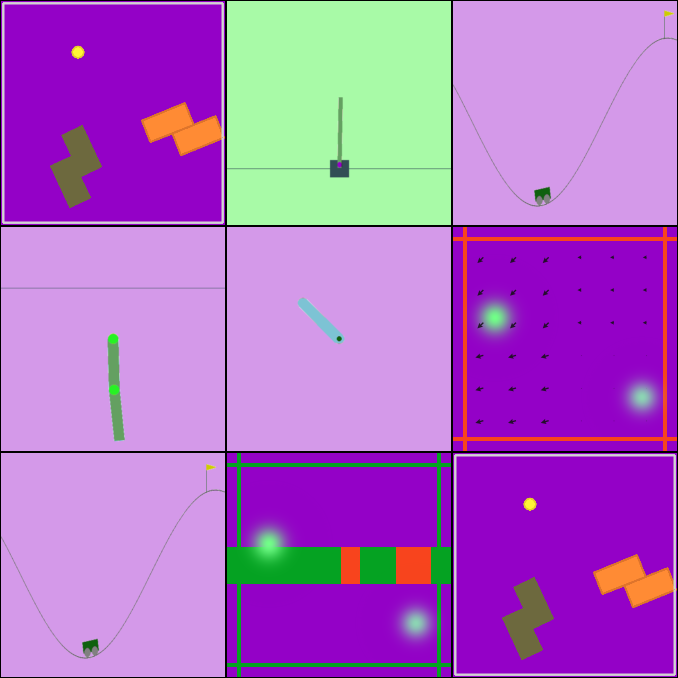}
        \caption{Classic Control}
        \label{fig:wrapper_classic}
    \end{subfigure}
    \hfill
    \begin{subfigure}[b]{0.195\textwidth}
        \centering
        \includegraphics[width=\textwidth]{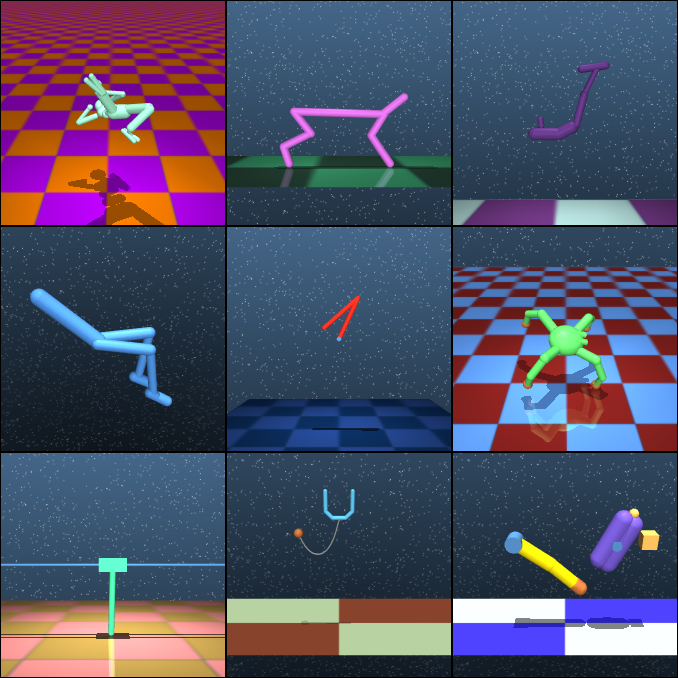}
        \caption{MuJoCo}
        \label{fig:wrapper_mujoco}
    \end{subfigure}
    \hfill
    \begin{subfigure}[b]{0.195\textwidth}
        \centering
        \includegraphics[width=\textwidth]{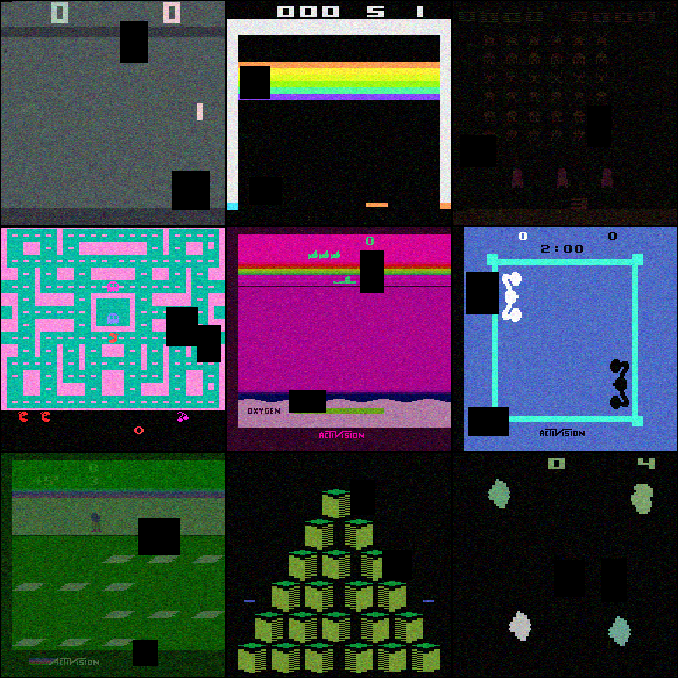}
        \caption{Arcade Game}
        \label{fig:wrapper_arcade}
    \end{subfigure}
    \hfill
    \begin{subfigure}[b]{0.195\textwidth}
        \centering
        \includegraphics[width=\textwidth]{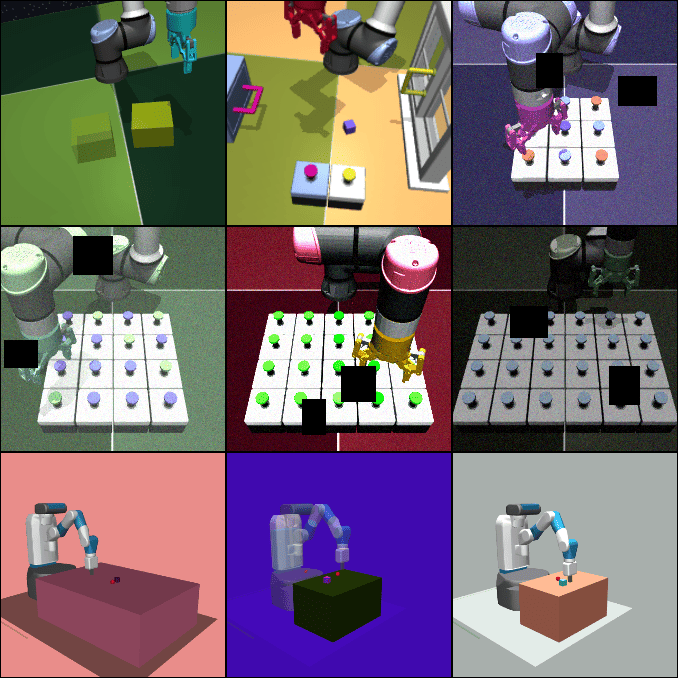}
        \caption{Robotics}
        \label{fig:wrapper_robotics}
    \end{subfigure}
    \hfill
    \begin{subfigure}[b]{0.196\textwidth}
        \centering
        \includegraphics[width=\textwidth]{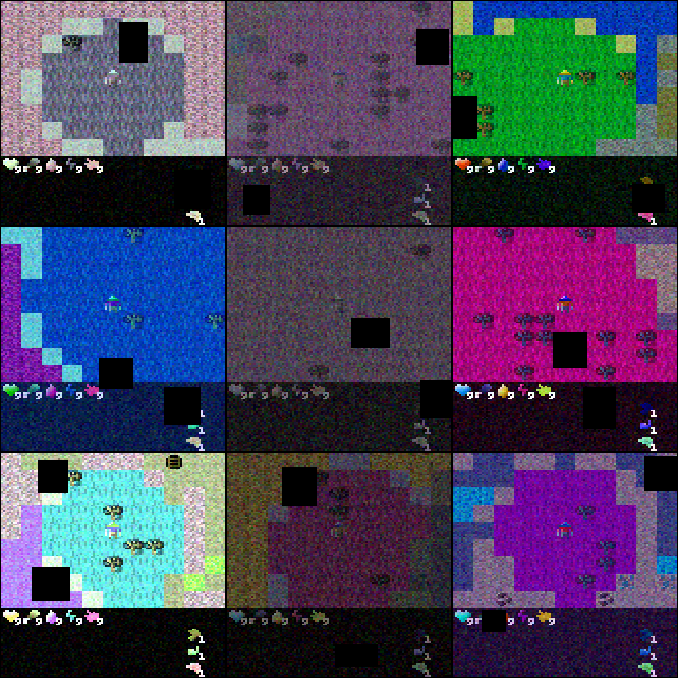}
        \caption{Open-World Game}
        \label{fig:wrapper_openworld}
    \end{subfigure}
    \caption{Environment families supported by \texttt{swm}. \textbf{Top row:} default (unperturbed) renderings of each environment. \textbf{Bottom row:} all visual factors of variation (e.g., agent, object, scene, geometry, lighting) jointly perturbed. Dynamic physical parameters (e.g., mass, density, gravity, or friction) can also be modified, but are omitted here as they are not visible in a single frame.}
    \label{fig:env_overview}
\end{figure}

\section{Background: the World Model Pipeline}
\label{sec:background}

In this section, we provide the necessary background on world models and control, and highlight key challenges that arise in their design, training, and evaluation.


\paragraph{Training World Models.} A world model is mostly the association of two core components, an encoder and a predictor. The encoder $\mathcal{E}: \mathbb{R}^n \rightarrow \mathbb{R}^D  $ maps a (possibly partial) observation $\mathbf{o}_t \in \mathbb{R}^n$ to a latent state $\mathbf{s}_t \in \mathbb{R}^D$. The predictor $\mathcal{P}: \mathbb{R}^D \times \mathbb{R}^A \rightarrow \mathbb{R}^D$ learns the dynamics by predicting the next state $\mathbf{s}_{t+1}$ given the current state $\mathbf{s}_t$ and an action $\mathbf{a}_t$. In practice, learning such representations from high-dimensional sensory inputs, such as videos, has become ubiquitous, yet it requires compressing observations into compact latent states that retain only task-relevant features. This has led to a wide range of representation learning strategies, including reconstruction-based objectives \cite{hafner2025training, hansen2023td, goswami2025world} or JEPAs \cite{lecun2022path, sobal2025learning, balestriero2025lejepa, maes2026leworldmodel}.

World modeling requires latent representations that faithfully capture a system's underlying dynamics, since accurate long-horizon prediction and planning depend on understanding how states evolve over time. This places strong demands on the representations: they must support rich temporal reasoning and generalize robustly to out-of-distribution scenarios. 



\paragraph{Planning with World Models.}

Once learned, a world model can be used to determine a sequence of actions that drives the system toward a desired goal state $\vs_g \in \R^D$. This problem can be framed as optimal control, where the objective is to find a policy $\pi$ which minimizes the expected cumulative cost induced by the learned dynamics. Formally, given an initial state $\vs_0$, we seek a mapping $\pi: \R^D \rightarrow \R^A$ such that

\begin{equation}
\label{eq:planning}
\begin{aligned}
    \pi^* = & \argmin_{\pi} \sum_{t} c(\vs_t, \va_t) \quad \text{s.t.} && \vs_{t+1} = \mathcal{P}(\vs_t, \va_t), \quad \va_t = \pi(\vs_t),
\end{aligned}
\end{equation}

where $\mathcal{P}$ denotes the world model predictor and $c$ is a task-dependent stage cost measuring progress toward the goal $\vs_g$.

Approaches to solving \eqref{eq:planning} typically fall into two categories. Reinforcement Learning (RL) learns the policy $\pi$ offline (or online) by optimizing it over many trajectories  \cite{hafner2023mastering, hafner2025training}, while Model Predictive Control (MPC) solves a finite-horizon ($T$ steps) version of this problem at test time by directly optimizing predicted trajectories over action sequences \cite{zhou2024dino, maes2026leworldmodel}. Unlike RL, MPC leverages the world model directly as part of the agent. For the rest of this work, we will rely on MPC to solve control, yet our platform is also compatible with RL as both approaches ultimately aim to solve the same underlying control problem.

\section{\texttt{stable-worldmodel}: A Unified Framework for World Model Research}
\label{sec:overview}

We introduce \texttt{stable-worldmodel} (Fig.~\ref{fig:swm}), an open-source platform for reproducible world model research and evaluation that supports researchers across the entire pipeline. As detailed in the following subsections, \texttt{swm} is motivated by key challenges in current practice, including fragmented implementations, data handling bottlenecks, and limitations in evaluation protocols. \texttt{swm} is entirely open-source and designed to integrate smoothly with existing code bases.


                       
                                    
                            

\subsection{Challenges in World Modeling Research}
\label{sec:challenges_wm}

We highlight key challenges across the world-modeling pipeline that motivate the need for a shared, standardized platform enabling efficient, reproducible, and rigorous evaluation and iteration.

\paragraph{Reproducibility and Implementation Fragmentation.} As depicted before, the plurality of approaches and motivations to tackle the world modeling task has led most recent works to operate in silos: each lab designs its own end-to-end pipeline for data collection, model training, and evaluation. In addition, these works routinely re-implement the same baselines, environments, and core algorithms from scratch. For example, the Cross-Entropy Method (CEM)~\citep{rubinstein2004cross} planner has been independently re-implemented (with varying degrees of fidelity) in at least five recent papers, including TDMPC~\cite{hansen2023td}, PLDM~\cite{sobal2025learning}, DINO-WM~\cite{zhou2024dino}, LeWM~\cite{maes2026leworldmodel}, and V-JEPA2~\cite{assran2025v}.
Repeatedly re-implementing the same components across different codebases often introduces subtle inconsistencies that undermine fair comparison and reproducibility~\citep{brooks1995mythical,islam2017reproducibility}. The resulting fragmentation reduces the trustworthiness of reported results and makes it difficult to isolate whether performance gains stem from genuine methodological advances or from implementation differences.

\paragraph{Data Bottlenecks in WMs learning.} World models rely on an encoder $  \mathcal{E}  $ to infer latent states $  \vs_t  $ from raw observations $  \vo_t  $. Unlike traditional pipelines, training such models requires loading multimodal data as contiguous temporal blocks, including video frames, actions, proprioception, and other sensor streams. 
This requirement creates a fundamental trade-off between random access efficiency and I/O throughput. Storing trajectories as individual frames (standard in computer vision) enables fast random sampling but incurs prohibitive I/O overhead and storage costs due to the high redundancy of file header decoding. Conversely, encoding trajectories as compressed video clips (e.g., MP4) drastically reduces disk footprint, yet severely degrades random access: retrieving a later frame often requires decoding all preceding frames in the clip. Consequently, neither approach scales effectively. 
Data loading is a real bottleneck, leading to GPU starvation, as loading fails to keep the accelerator utilized. 

\paragraph{Evaluating Robust World Models.}

Current world model research mostly depends on simulator evaluation because transferring models and controllers to real environments is expensive, time-consuming, and technically demanding. As a result, Gym-style simulation benchmarks have become the dominant proxy for measuring generalization and control performance~\citep{towers2024gymnasium}. However, standard benchmarks often evaluate models under conditions close to the training distribution, making it difficult to determine whether a world model has learned reusable environment dynamics or merely exploitable correlations in the data. This distinction is central to world modeling: a world model is not itself a policy, but a learned representation of how observations, states, and actions evolve over time, which can then be used by MPC, model-based RL, or other downstream planners. Recent methods such as DINO-WM~\citep{zhou2024dino}, PLDM~\citep{sobal2025learning}, LeWM~\citep{maes2026leworldmodel}, TD-MPC2~\citep{hansen2023td}, and Genie~\citep{bruce2024genie} illustrate the diversity of architectures and objectives used to learn such predictive structure. Yet predictive accuracy or task success in-distribution does not necessarily imply that the learned latent dynamics are temporally stable, intervention-robust, or useful for counterfactual reasoning and long-horizon planning. Recent analyses of learned physical systems further show that high-capacity sequence models can fit trajectories accurately while failing to recover the local dynamical laws that generated them~\citep{liu2026keplernewtoninductivebiases}. This limitation is especially problematic because many important dynamical modes, such as contact interactions, occlusions, or changes in physical parameters, only appear under specific behaviors or environment configurations that may be absent from the training data. Consequently, a model may appear successful on standard benchmarks while still harboring fundamental misunderstandings of the underlying dynamics. Zero-shot and out-of-distribution evaluation help expose such failures by testing whether learned dynamics remain useful under controlled changes in appearance, geometry, object properties, memory requirements, or physical parameters.

\subsection{Design Principles and Interface}
\label{sec:design}

The core philosophy behind \texttt{swm} is to impose as few restrictions as possible on the user's model architecture and training code, while providing strong standardization for data collection, evaluation, and control. Researchers typically have their own preferred training frameworks and model designs; forcing a particular training loop would limit adoption. In contrast, data collection, environment interaction, and evaluation protocols benefit greatly from standardization, enabling fair and reproducible comparisons.

To achieve this balance, \texttt{swm} is built around three minimal yet powerful abstractions (Fig.~\ref{fig:swm}):

\begin{itemize}
    \item \textbf{\texttt{World}}: A unified environment wrapper that supports data collection, policy execution, and evaluation across diverse simulators (Gymnasium-compatible). It handles vectorized execution, rendering, and controllable intervention on the environment's visual, geometric, and physical properties, which we call \textit{factors of variation} (FoV).
    
    \item \textbf{\texttt{Policy}}: A simple interface that maps observations (or latent states) to actions. This includes random policies, expert policies (e.g., SAC), and learned policies from RL or MPC. In particular, \texttt{MPCPolicy} wraps any world model and planner solver: at each timestep it encodes the current observation into a latent state and delegates action selection to the underlying solver by solving the finite-horizon optimal control problem (Equation~\ref{eq:planning}).
    
    \item \textbf{\texttt{Solver}}: A self-contained and robust implementation of various single-shooting planning algorithms for MPC. Supported solvers include the Cross-Entropy Method (CEM), Model Predictive Path Integral (MPPI), Gradient Descent, Projected Gradient Descent, and many more. Each solver simply expects their associated world model to implement a \texttt{get\_cost} method and use that signal to return an optimized finite-horizon action sequence. All solvers are extensively tested and benchmarked for numerical stability and performance.
\end{itemize}

Overall, our non-invasive design enables a clean, reproducible research loop with minimal boilerplate, as illustrated in Algorithm~\ref{alg:swmloop}. Furthermore, by separating model training from the evaluation infrastructure, \texttt{swm} lets researchers focus on algorithmic innovation while benefiting from battle-tested data pipelines, planning solvers, and standardized benchmarks. New environments can be added with minimal effort, as long as they follow the Gymnasium API.

\begingroup
\RestyleAlgo{plain}

\begin{algorithm}[ht]
\caption{Typical \texttt{stable-worldmodel} loop illustrating the full pipeline: \textbf{(1)} instantiating a world composed of 8 Push-T environments
\textbf{(2)} collecting a dataset using a given policy (here random, but expert policies are also supported), 
\textbf{(3)} training or loading a user-defined world model, 
\textbf{(4)} creating a planner solver and wrapping it into an MPC policy, which and run evaluation on a modified environment with a random agent size and background color. The same interface works with any supported solver and environment.}
\label{alg:swmloop}
\rule{\linewidth}{1pt}
\begin{lstlisting}
import stable_worldmodel as swm
# 1] Create the world
world = swm.World(
    env_name="swm/PushT-v1", num_envs=8, max_episode_steps=1000)
# 2] Collect data
world.set_policy(swm.policy.RandomPolicy(seed=42))
world.collect(dataset_path="pusht.lance", episodes=5000, seed=42)
# 3] Train/Load your world model (user-defined)
model = MyWorldModel(...)   # Any PyTorch model
# 4] Evaluate with planning
planner = swm.solver.CEMSolver(model, ...)
wm_policy = swm.policy.MPCPolicy(planner)
world.set_policy(wm_policy)
metrics = world.evaluate(
    episodes=100, seed=0, video="videos/",
    options={'variation': ['agent.size', 'background.color']})
print(metrics["success_rate"])
\end{lstlisting}
\rule{\linewidth}{1pt}
\end{algorithm}

\endgroup

\subsection{System components}

Here, we describe the concrete choices made in \texttt{swm} to directly address the challenges outlined in Sec.~\ref{sec:challenges_wm}. We begin with the data layer, detailing the decisions that enable efficient collection and loading of multimodal trajectories. We then present the baseline world models and planning solvers currently implemented in the platform. Finally, we introduce \texttt{swm}'s core contribution: its evaluation testbed. This includes a broad suite of supported environments together with a rich set of customizable factors of variation (visual, geometric, and physical) that can be applied on-the-fly to any environment.


\begin{figure}[tbp]
    \centering
    \begin{minipage}[c]{0.55\textwidth}
        \centering
        \begin{tabular}{lrr}
        \toprule
                       & \multicolumn{2}{c}{\textbf{Throughput} (sample/sec)} \\
        \cmidrule(lr){2-3}
        \textbf{Format}         & w/o caching & w/ caching \\
        \midrule
        HDF5 (local)   & 1{,}416    & 1{,}474 \\
        HDF5 (S3)      &         9  &    757  \\
        Lance (local)  & 4{,}815    & 4{,}431 \\
        Lance (S3)     & 3{,}184    & 3{,}253 \\
        Video (local)  & 1{,}331    & 1{,}348 \\
        \bottomrule
        \end{tabular}
    \end{minipage}
    \hfill
    \begin{minipage}[c]{0.4\textwidth}
        \centering
        \includegraphics[width=\textwidth]{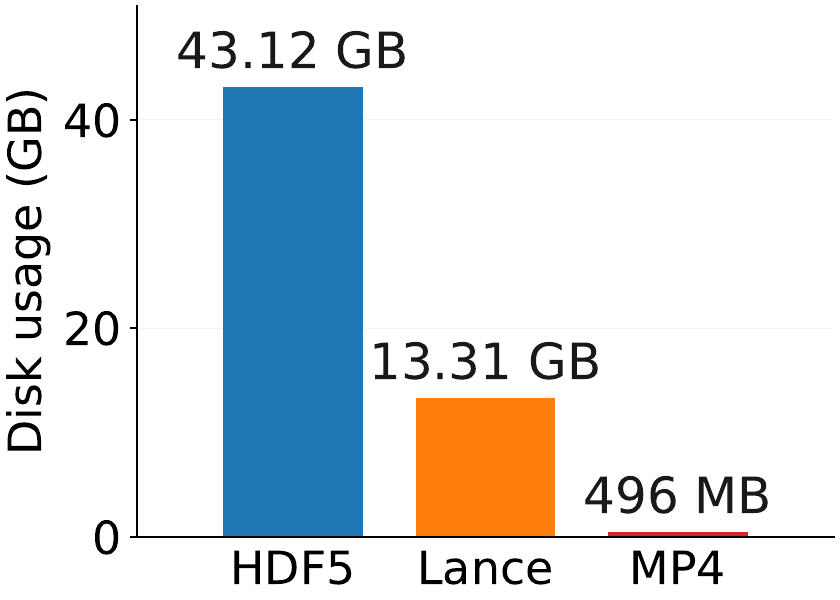}
    \end{minipage}
    \caption{Performance comparison of different data formats for a dataset from the Push-T environment. {\bf(Left)} Data loading throughput (samples/sec) with and without caching, for both local storage and remote (S3) streaming. {\bf(Right)} Disk storage usage. Results demonstrate that Lance is the most efficient format in terms of throughput while remaining a good trade-off in compression for storage.}
    \label{fig:pusht-data}
\end{figure}

\paragraph{Data Layer.}
\texttt{swm} provides a flexible data collection pipeline supporting both offline and online regimes. Users can generate trajectories with random policies for broad exploration, pre-trained expert policies, policies loaded from model checkpoints, or any mixture thereof. This design makes it straightforward to study the influence of data composition on world model performance. To perform online world model training as in TD-MPC2~\citep{hansen2023td}, the platform natively supports iterative alternation between data collection and model training or evaluation by simply repeating the steps of Alg.~\ref{alg:swmloop}. All collections are performed through the unified \texttt{World} interface and can be combined on-the-fly with controllable factors of variation (FoV) for visual, geometric, or physical perturbations (Fig.~\ref{fig:env_overview}).

As discussed in Sec.~\ref{sec:challenges_wm}, efficient loading of multimodal, temporally contiguous trajectories is a major bottleneck in world model training. 
To address this, \texttt{swm} adopts Lance~\citep{pace2025lance} as its primary storage format. Lance is a modern, columnar, ML-optimized format that offers fast random access, high compression ratios, zero-copy operations, native versioning, and seamless streaming from cloud object stores. However, for maximum compatibility, \texttt{swm} also supports common alternatives (MP4, HDF5, and LeRobot~\citep{cadenelerobot}) and provides one-click conversion tools to Lance. This allows researchers to immediately use existing datasets while benefiting from standardized, high-performance storage. In particular, LeRobot datasets include a large and growing collection of real-robot trajectories, which are seamlessly integrated via an adapter and can be automatically converted to Lance for optimal throughput. Benchmarks on Push-T (Fig.~\ref{fig:pusht-data}) and Two-Room (Fig.~\ref{fig:tworoom-data}) validate that Lance achieves the highest loading throughput, outperforming locally stored HDF5, MP4, and LeRobot, and even surpassing remote HDF5 when streaming online from S3.

\paragraph{Control Layer.} As noted in Sec.~\ref{sec:challenges_wm}, commonly used solvers have been independently reimplemented across many recent world model papers. \texttt{swm} addresses this by providing a collection of planning solvers, easily pluggable with any trained world model through an \texttt{MPCPolicy} interface. At every environment step, the active solver receives the current latent state and uses the world model to evaluate candidate action sequences by rolling out their trajectories over a planning horizon. It then returns an optimized action sequence, of which the $K$ first actions are executed before replanning at the next step.
The implemented solvers cover two broad families of approaches. Sampling-based and Gradient-based methods. Sampling methods include Predictive Sampling~\cite{howell2022predictive}, Cross-Entropy Method (CEM)~\cite{rubinstein2004cross}, improved CEM (iCEM)~\cite{pinneri2021sample}, and Model Predictive Path Integral (MPPI)~\cite{williams2016aggressive}. They iteratively sample, evaluate, and refine populations of candidate action sequences without requiring any information on the model (e.g., gradient). Gradient-based methods, including Gradient Descent, Projected Gradient Descent~\citep{henaff2017model}, and GRASP~\citep{psenka2026parallel}, instead exploit differentiability of the predictor to optimize actions through back-propagation.
All solvers are tested and validated end-to-end: when paired with their original world models, our implementations reproduce the planning success rates reported in DINO-WM~\cite{zhou2024dino} and PLDM~\cite{sobal2025learning} (see Sec.~\ref{sec:exp_planning}), confirming that the unified interface introduces no performance regression. Pseudocode for all solvers is provided in App.~\ref{app:solvers}.

\paragraph{Model Layer.}
\texttt{swm} implements multiple baselines that mainly span two paradigms: goal-conditioned reinforcement learning (GCRL) and latent world models with test-time planning. Both instantiate control through a subclass of the common \texttt{Policy} interface. Concretely, GCRL methods directly parameterize a \texttt{FeedForwardPolicy} that maps observations (and goals) to actions, whereas world model approaches wrap a learned dynamics model inside an \texttt{MPCPolicy}, which selects actions by solving the finite-horizon planning problem described in Eq.~\ref{eq:planning} at each timestep. As described in Sec.~\ref{sec:design}, all methods share a common data management and evaluation loop, removing any ambiguity from comparisons.
On the GCRL side, goal-conditioned behavioral cloning (GCBC \cite{ghosh2020learningreachgoalsiterated}) is a supervised learning approach that trains policies on expert trajectories. Goal-conditioned implicit Q-learning methods (e.g., GCIVL and GCIQL \cite{kostrikov2021offline}) instead learn Q functions via expectile regression and extract policies through advantage-weighted updates.
On the planning side, world models follow the structure described in Sec.~\ref{sec:background}, consisting primarily of an encoder $\mathcal{E}$ and a predictor $\mathcal{P}$. At test time, these models are used within an \texttt{MPCPolicy}, which encodes the current observation into a latent state and optimizes actions with respect to a goal observation by solving the planning objective. The different world model baselines share this abstraction and differ mainly in their training objectives and architectural choices. DINO-WM \cite{zhou2024dino} freezes a pretrained DINOv2 encoder \cite{oquab2023dinov2} and learns a ViT~\citep{dosovitskiy2021an} predictor over spatial patch features, serving as a baseline for latent world models from foundation models. Planning with a Latent Dynamics Model (PLDM)~\citep{sobal2025learning} learns latent representations and dynamics jointly from offline trajectories using a JEPA-style objective, with additional regularization to stabilize training. LeWorldModel (LeWM)~\citep{maes2026leworldmodel} is also a JEPA-based world model that simplifies the training objective while maintaining stable learning, and is used here as the fastest latent world model baseline. Finally, we include TD-MPC2 ~\citep{hansen2023td}, a decoder-free world model that performs local trajectory optimization guided by discrete reward and value predictions.

\subsection{Benchmarking World Models}
\label{sec:benchmark_wm}

\texttt{swm} supports online evaluation with boundary conditions sampled either randomly from the state space or from selected trajectories, following \cite{zhou2024dino, sobal2025learning, maes2026leworldmodel}; see App.~\ref{app:evaluation_protocol} for details. We report success rate as the primary metric for control and planning performance.

This protocol applies across environments spanning multiple regimes (Fig.~\ref{fig:env_overview}): low-dimensional and 2D control (CartPole~\citep{barto2012neuronlike}, PushT~\citep{chi2023diffusionpolicy}), 3D visual environments (OGBench~\citep{park2025ogbench}), discrete domains (Atari~\citep{bellemare2013arcade}), continuous control (MuJoCo~\citep{todorov2012mujoco}), and partial observability (Craftax~\citep{matthews2024craftax}). The suite spans reactive dynamics (Atari, MuJoCo), long-horizon planning (OGBench manipulation, Craftax), and varied action and state spaces.

To probe robustness and zero-shot generalization, we introduce controlled factors of variation (FoV) (Fig.~\ref{fig:env_overview}) targeting visual factors (colors, lighting, textures, occlusions) and physical properties (masses, friction, dynamics). When simulator internals are inaccessible (e.g., Atari ROMs), lightweight visual wrappers can be applied at the observation level (Fig.~\ref{fig:wrapper}), enabling systematic measurement of OOD transfer, continual learning, or generalization. More details can be found in App.~\ref{app:env}

\section{Case Study: Planning Performance and Zero-Shot Evaluation of WMs}
\label{sec:experiments}

We showcase the practical utility of \texttt{swm} on the Push-T manipulation benchmark, a standard testbed for evaluating world models. Using the platform’s unified evaluation pipeline, we demonstrate how \texttt{swm} enables seamless assessment of planning performance and zero-shot generalization across diverse settings with minimal additional code. These experiments highlight common failure modes in existing models while illustrating how \texttt{swm} facilitates systematic diagnosis and comparison. Detailed experimental setup and additional results are provided in App.~\ref{app:exp_details} and App.~\ref{app:additional_exp}.

\subsection{In-distribution Planning Performance}
\label{sec:exp_planning}

\begin{wraptable}{r}{0.4\columnwidth}
    \vspace{-1em} 
    \centering
    \caption{Baseline comparison.}
    \begin{tabular}{lcc}
        \toprule
        Method & Push-T & OGB-Cube \\
        \midrule
        TD-MPC2       & 12 & 4 \\
        GCBC          & 75 & 84 \\
        LeWM          & 94 & 72 \\
        PLDM          & 78 & 62 \\
        DINO-WM       & 92 & 86 \\
        \bottomrule
    \end{tabular}
    \label{tab:baselines_sr}
\end{wraptable}

We first report the in-distribution planning performance of representative state-of-the-art world models on Push-T: TD-MPC2~\citep{hansen2023td}, GCBC~\citep{ghosh2020learningreachgoalsiterated}, LeWorldModel~\citep{maes2026leworldmodel}, PLDM~\citep{sobal2025learning}, and DINO-WM~\citep{zhou2024dino}. All models were trained on the same expert dataset used in DINO-WM and evaluated under identical planning configurations. Tab.~\ref{tab:baselines_sr} summarizes their success rates. We recover performance values consistent with those originally reported for baselines evaluated on Push-T. However, we observed that TD-MPC2 performs poorly in an offline setting; we conjecture that this stems from generating OOD actions, fooling the predictor (see Fig.~\ref{fig:pca-tdmpc} in App.~\ref{app:additional_exp} for a direct visualization of this drift). We verify our TD-MPC2 implementation by benchmarking it online on the DeepmindControl tasks used in the original paper and compare against a Soft-Actor-Critic (SAC) baseline from Stable-Baselines3 (Tab.~\ref{tab:online_training_rewards}).

\subsection{World Models Are Still Brittle Under Distribution Shift}

We assess the robustness of world models under distribution shift through progressive shifts and controlled visual perturbations.

\begin{figure}[tbp]
    \centering
    \includegraphics[width=\textwidth]{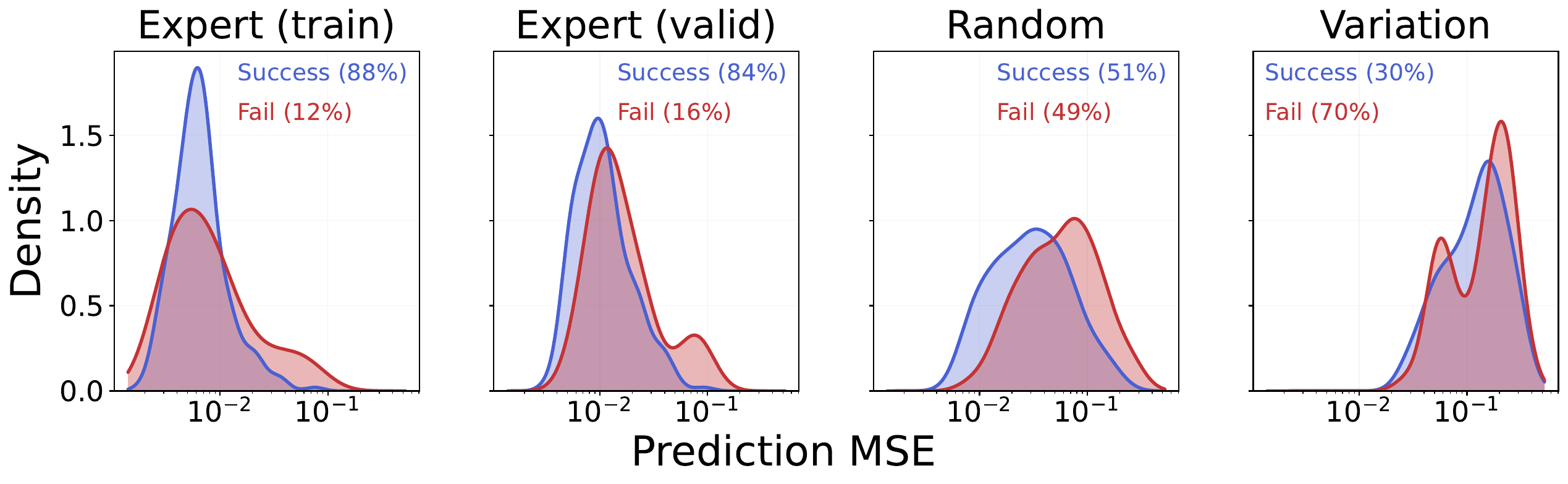}
    \caption{Distribution of trajectory-level prediction MSE for successful (blue) and failed (red) plans on Push-T using LeWM, across four levels of increasing distribution shift. From left to right: expert training trajectories, held-out expert validation trajectories, random-policy trajectories, and random-policy trajectories with full variations (color, shape, size, background, etc.). Prediction error does not seem to correlate with SR. The same qualitative result holds for PLDM (see Fig.~\ref{fig:pldm_pred_vs_sr}).}
    \label{fig:lewm_pred_vs_sr}
\end{figure}

We first consider progressive distribution shift by performing planning on expert training data, held-out expert validation data, random-policy trajectories, and finally random-policy trajectories with all \texttt{swm} factors of variation (color, shape, size, background, etc.). For each of the four settings, we run 256 trajectories and plot the distribution of trajectory-level prediction MSE for successful and failed planning (Fig.~\ref{fig:lewm_pred_vs_sr} and ~\ref{fig:pldm_pred_vs_sr}).
Prediction error correlates poorly with planning success for both models. While more OOD settings produce higher prediction error, the success and failure distributions largely overlap even under strong distribution shifts. This suggests that out-of-distribution inputs, rather than raw prediction error magnitude, are the primary driver of planning failures.

\begin{figure*}[t]
\centering

\begin{subfigure}[t]{0.52\textwidth}
    \vspace{0pt}
    \centering
    \includegraphics[width=\linewidth]{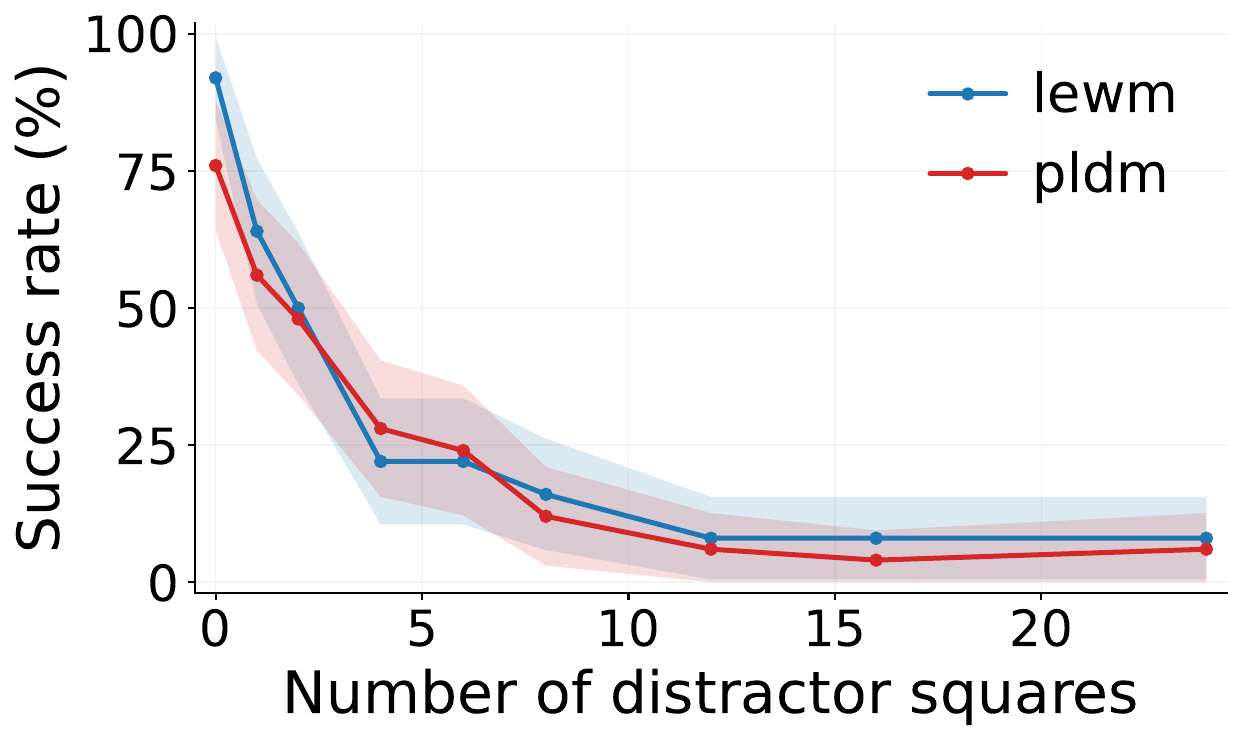}
    \caption{SR evolution wrt visual distractors on Push-T.}
    \label{fig:pusht-distractor}
\end{subfigure}
\hfill
\begin{subtable}[t]{0.44\textwidth}
    \vspace{0pt}
    \centering
    \setlength{\tabcolsep}{1.5pt}
    \begin{tabular}{@{}l l c c c@{}}
        \toprule
        \textbf{FoV} & \textbf{Entity} & \multicolumn{3}{c}{\textbf{SR \% ($\uparrow$)}} \\
        \cmidrule(lr){3-5}
                 &             & \textbf{LeWM} & \textbf{PLDM} & \textbf{DINO-WM} \\
        \midrule
        \textbf{None }    &             &\textbf{50.8} & \textbf{50.8} & \textbf{20.0} \\
        \midrule
        Color    & Agent       & 12.0 &  8.0 & 18.0 \\
                 & Block       & 22.0 & 18.0 & 18.0 \\
                 & Canvas      &  6.0 &  6.0 & 10.0 \\
        \addlinespace
        Size     & Agent       & 22.0 & 18.0 &  4.0 \\
                 & Block       & 20.0 & 18.0 & 16.0 \\
        \addlinespace
        Shape    & Agent       & 26.0 & 52.0 & 18.0 \\
                 & Block       & 12.0 & 14.0 &  8.0 \\
        \bottomrule
    \end{tabular}
    \caption{Planning success under factor variations.}
    \label{tab:factor_variation_short}
\end{subtable}

\caption{Robustness analysis on Push-T. Left: effect of visual distractors. Right: planning success under factor variations.}
\label{fig:combined_pushT}
\end{figure*}

To more directly isolate the effect of distribution shift, we next introduce targeted visual perturbations using \texttt{swm}'s factors-of-variation interface. As shown in Tab.~\ref{tab:factor_variation_short}, planning success rates drop sharply for most models even under mild perturbations, confirming that current world models remain brittle outside their training distribution. Fig.~\ref{fig:lewm-wheel} reports the variations of SR as a function of color wheel background intensities. Additionally, using the occlusion wrapper introduced in Fig.~\ref{fig:wrapper}, we report in Fig.~\ref{fig:pusht-distractor} the planning success rate as a function of the number of visual distractor squares. We observe a quadratic decay: models tolerate small numbers of distractors but degrade rapidly beyond that point. The pattern holds across all baselines. Overall, these results indicate that current world models exhibit limited zero-shot generalization: even modest shifts outside the training distribution lead to substantial degradation in planning performance.
\texttt{swm} makes such systematic robustness studies trivial to run and reproduce. 

These case studies demonstrate how \texttt{swm} turns previously ad-hoc analyses into standardized, reproducible experiments, accelerating the identification of current limitations and the development of more robust world models.



\section{Related Work}
\label{sec:related-work}

\paragraph{Benchmarking World Models.} Learning predictive models of the environment has become a central AI paradigm, driving a steady increase in the development of world models. 
Generative latent approaches, such as Dreamer \citep{hafner2019dream, hafner2020mastering, hafner2023mastering, hafner2025training, bruce2024genie} rely on pixel reconstruction, while implicit methods like TD-MPC2 \citep{hansen2022temporal,hansen2023td} optimize for reward prediction. 
A recent alternative trains purely via self-supervised prediction in learned representation spaces to model dynamics~\citep{sobal2025learning,garrido2024learning,zhou2024dino,bar2025navigation,maes2026leworldmodel,assran2025v}.
Despite their shared philosophy, standalone codebases lead to duplicated implementations and inconsistent evaluation, hindering research progress.
Recent benchmarks such as WorldMark ~\citep{xu2026worldmark} and comprehensive frameworks like WorldTest~\citep{warrier2025benchmarking} standardize the evaluation of interactive video and behavior-based models, yet their training infrastructure remains fragmented. Conversely, works like EB-JEPA \citep{terver2026lightweight} are restricted to JEPA-style architectures, are mainly developed for educational purposes, and lack scalable research components, such as efficient data-loading, training-agnostic evaluation, and modeling baselines.
\texttt{swm} addresses these shortcomings by providing a model-agnostic, complete ecosystem for efficient training and evaluation.

\paragraph{Benchmarking RL.} World model evaluation is closely tied to Reinforcement Learning (RL), where standardized benchmarks have driven substantial progress ~\citep{schrittwieser2020mastering, mnih2013playing, hafner2019dream, haarnoja2018soft, schulman2017proximal, alonso2024diffusion}.
Datasets such as \texttt{D4RL}~\citep{fu2020d4rl} and \texttt{OGBench}~\citep{park2025ogbench} dominate offline and goal-conditioned RL evaluations, but focus mainly on model-free approaches using static trajectories in fixed environments. Concurrently, visual robustness has been studied via the \texttt{Natural RL Environment}~\citep{zhang2018natural}, \texttt{Distracting Control Suite}~\citep{stone2021distracting}, \texttt{DMControl Generalization Benchmark}~\citep{hansen2021stabilizing}, and \texttt{DMC-VB}~\citep{ortiz2024dmc}. 
However, these remain tied to specific environment families and only address visual perturbations. Recent efforts have also highlighted the importance of rapid adaptation to structurally novel environments with different dynamic mechanics (e.g., \textit{different games})~\citep{ying2025assessing}.
To address this, \texttt{swm} reuses environments from these benchmarks and introduces systematic variations to evaluate robustness, zero-shot adaptation, and generalization.

\paragraph{Complementary Frameworks.} The open-source ecosystem already supports algorithm development outside of the field of world modeling.
For instance, \texttt{stable-baselines3}~\citep{raffin2021stable} and \texttt{CleanRL}~\citep{huang2022cleanrl} standardize model-free baselines, while \texttt{mbrl-lib}~\citep{pineda2021mbrl} provides an older, less-maintained model-based counterpart.
Robotics frameworks such as \texttt{LeRobot}~\citep{cadenelerobot}, \texttt{RoboHive}~\citep{kumar2023robohive}, 
\texttt{robosuite}~\citep{zhu2020robosuite}, and \texttt{RLBench}~\citep{james2020rlbench} focus on imitation learning and policy optimization, leaving world-model infrastructure largely unaddressed. 
Rather than replacing these ecosystems, \texttt{swm} integrates seamlessly with frameworks like \texttt{stable-baselines3} and \texttt{LeRobot} to supply missing world-model components, like planning or evaluation. \texttt{swm} aims to become a unified, standard reference library for the research community.

\section{Conclusion and Future Work}

We introduced \texttt{stable-worldmodel}, an open-source platform for reproducible research and evaluation of world models. \texttt{swm} provides efficient, modular, and well-tested tools to reduce the friction in the full world modeling pipeline, including high-performance data management, planning solvers, and standardized evaluation benchmark extended with customisable visual and physical properties. 
Our experiments highlight that current world models remain far from achieving robust zero-shot generalization, even under mild distribution shifts, underscoring the need for advances in both visual and physical robustness as well as long-horizon consistency. These findings suggest that improved generalization may require both architectural advances and systematic scaling. By enabling a reliable and scalable evaluation workflow, \texttt{swm} makes it possible to rigorously study scaling trends and their impact on generalization, supporting more controlled and reproducible investigations of existing and future architectures. Looking forward, we aim to extend the platform beyond simulation toward improved sim-to-real transfer, support for asynchronous real-time interaction, and efficient online training, as well as broaden its applicability to a wider range of scientific domains.
We hope this unified framework will foster fairer comparisons, more reliable generalization studies, and accelerated progress toward robust and trustworthy world models for the real world.

\bibliography{ref}
\bibliographystyle{unsrtnat}


\appendix

\section*{Appendix}

Our Appendix complements the main paper with a walkthrough of the \texttt{stable-worldmodel} platform. We begin with installation and environment setup, before turning to the core abstractions of the library: \texttt{World}, \texttt{Policy}, \texttt{Solver}, factors of variation, wrappers, and the dataset API. Building on these foundations, we detail the supported environment families along with the two complementary intervention mechanisms used to construct distribution shifts, namely the factors of variation and the wrappers. We then review the implemented baselines, presenting their training objectives and accompanying pseudo-code, followed by the evaluation protocols supported by the platform. Next, we cover the supported planning solvers, spanning both sampling-based and gradient-based methods, each illustrated, again, with their own pseudocode. We also document the ready-to-use scripts and command-line interface that make the platform immediately usable in practice. Finally, we outline open research problems that the platform makes tractable, provide concrete guidelines for contributors adding new environments, solvers, or baselines, and close with additional experimental results.

\section{Installation \& Setup}

This section walks through the process of installing \texttt{stable-worldmodel} and preparing your environment for development and experimentation. We recommend using \texttt{uv} as your Python package manager and Linux as the operating system for the best experience. The library should also work on macOS or Windows, and an alternative setup using Conda is possible.

\texttt{swm} is designed to be easy to set up while supporting both local development and large-scale distributed training.

\textbf{Development setup:}
\begin{lstlisting}
[assuming downloaded code]
cd stable-worldmodel
uv venv --python=3.10 && source .venv/bin/activate
uv sync --all-extras --group dev
\end{lstlisting}

{\bf Note: }\texttt{swm} uses the environment variable \texttt{\textcolor{ForestGreen}{\$STABLEWM\_HOME}} (default: \texttt{\textcolor{ForestGreen}{\textasciitilde/.stable\_worldmodel}}) to locate datasets and checkpoints. Datasets are stored in \texttt{\textcolor{ForestGreen}{\$STABLEWM\_HOME/datasets}} and checkpoints in \texttt{\textcolor{ForestGreen}{\$STABLEWM\_HOME/checkpoints}}.

\section{Core Abstractions}

The design of \texttt{stable-worldmodel} revolves around a small set of clean, composable abstractions that unify data collection, training, and evaluation of world models. These abstractions are deliberately lightweight, compatible with the PyTorch deep learning framework and Gymnasium-based environments, while adding powerful features tailored for world model research.

The central abstraction is the \textcolor{ForestGreen}{\texttt{World}} class. A \textcolor{ForestGreen}{\texttt{World}} wraps one or more vectorized Gymnasium environments through an efficient internal \textcolor{ForestGreen}{\texttt{EnvPool}} and provides a unified interface for simulation, dataset recording, visualization, and evaluation. Unlike the standard Gymnasium API, \textcolor{mauve}{\texttt{World.reset()}} and \textcolor{mauve}{\texttt{World.step()}} do not return observations or rewards. Instead, they update an internal dictionary called \texttt{world.infos} in place. This dictionary contains all information returned by the underlying environments (RGB frames, states, rewards, terminations, truncations, etc.), giving immediate and consistent access to the complete state of all environments at every timestep. Note that each call to \textcolor{mauve}{\texttt{reset()}} or \textcolor{mauve}{\texttt{step()}} replaces the previous \texttt{world.infos} content.

Action selection is decoupled via a \textcolor{ForestGreen}{\texttt{Policy}} object. After calling \textcolor{ForestGreen}{\texttt{world.set\_policy(policy)}}, every call to \textcolor{mauve}{\texttt{World.step()}} automatically queries the associated policy for actions. This design makes it trivial to swap between expert policies, random policies, learned policies, or planning-based controllers at any time without modifying the world loop. A policy only needs to implement a \textcolor{mauve}{\texttt{get\_actions(info)}} method that receives the current \texttt{infos} dictionary. \texttt{swm} provides several ready-to-use policies, including heuristic policies, \textcolor{ForestGreen}{\texttt{FeedForwardPolicy}} for reinforcement learning agents, and \textcolor{ForestGreen}{\texttt{MPCPolicy}} for test-time planning.

Planning and model-predictive control are supported via \textcolor{ForestGreen}{\texttt{Solver}} implementations (e.g., \textcolor{ForestGreen}{\texttt{CEMSolver}}, \textcolor{ForestGreen}{\texttt{MPPISolver}}) and the \textcolor{ForestGreen}{\texttt{MPCPolicy}} abstraction. Any world model that implements a \textcolor{mauve}{\texttt{get\_cost}} method can be directly turned into a deployable controller.

Another key abstraction is the notion of \textbf{Factors of Variation (FoV)}. Every supported environment exposes a set of controllable properties (colors, shapes, sizes, physical parameters, lighting, etc.) through a dedicated \texttt{variation\_space}. These factors can be fixed, randomly sampled, or systematically varied during data collection or evaluation, enabling rigorous studies of robustness, domain shift, continual learning, and zero-shot generalization.

The \textcolor{ForestGreen}{\texttt{World}} interface also supports \textcolor{ForestGreen}{\texttt{Wrapper}} objects that can be easily attached to modify parts of the \texttt{infos} dictionary or alter default behavior (e.g., adding observations, changing rewards, applying visual transformations, or injecting custom logic). For more details on wrappers, see Appendix~\ref{app:env} or Section~3 of the paper.

Data handling is managed through the \textcolor{ForestGreen}{\texttt{Dataset}} API. \texttt{swm} supports multiple storage formats: Lance, HDF5, MP4 (video), and LeRobot, through a flexible format registry. Each format provides its own reader and writer implementation, making it straightforward to swap formats or add new ones. A convenient \textcolor{mauve}{\texttt{swm.data.convert()}} utility allows one-line conversion between any supported formats (provided the corresponding reader and writer exist).

Together, these abstractions create a highly modular pipeline where researchers can focus on their algorithm evaluation while relying on \texttt{swm} for reliable environments, efficient data pipelines, and solver algorithms. For a more detailed description of each abstraction, we refer the reader to the documentation.

\section{Environments and Factors of Variation (FoV)}
\label{app:env}

The main strength of \texttt{swm} lies in its large, standardized, and highly customizable suite of environments. All environments follow the standard \textcolor{ForestGreen}{\texttt{Gymnasium}} interface while being deeply integrated with the \textcolor{ForestGreen}{\texttt{World}} abstraction, enabling high-throughput data collection, visualization, and evaluation as mentioned earlier.

\texttt{swm} supports environments from a wide range of domains:
\begin{itemize}
    \item \textbf{DeepMind Control Suite (DMC)}: locomotion, control and manipulation tasks (continuous)
    \item \textbf{OGBench}: 3D object manipulation and scene understanding (continuous)
    \item \textbf{Classic Control Suite}: Classic control tasks (mountain car, inverted pendulum, etc.) (continuous/discrete)
    \item \textbf{Fetch-Suite}: 3D robotic manipulation tasks (continuous)
    \item \textbf{Craftax}: open-world procedural 2D survival game (pixel \& symbolic) (discrete)
    \item \textbf{Arcade Learning Environment (ALE)}: 100+ Atari games (discrete)
    \item \textbf{Extra Environments}: \texttt{swm/PushT-v1} (2D manipulation) and \texttt{swm/TwoRoom-v1} (simple navigation) (continuous)
\end{itemize}

A \textcolor{ForestGreen}{\texttt{World}} can be instantiated with any registered environment ID. For example:
\begin{lstlisting}
import stable_worldmodel as swm
world = swm.World("swm/PushT-v1", num_envs=8, image_shape=(64, 64))
\end{lstlisting}

{\bf Note:} \texttt{swm} can load by default any environment registered through Gymnasium and work with them. However, unless explicitly added, no Factors of Variation will be available for these non-supported environments.

\subsection{Environment Registry}

Every environment supported by \texttt{swm} is registered under a Gymnasium-compatible ID and can be instantiated directly through the \texttt{World} interface. The registry distinguishes two tiers of environments based on the level of integration with \texttt{swm}'s intervention machinery.

\paragraph{Native environments} (registered under the \texttt{swm/*} namespace) expose a structured state dictionary in their \texttt{info} output and declare a typed \texttt{variation\_space} listing all available factors of variation (FoV). This allows simulator-level interventions: scene properties, geometry, and physical parameters can be edited at reset time through the standard Gymnasium \texttt{options} argument.

\paragraph{External environments} (e.g., ALE, Craftax) are imported as-is and run through \texttt{World} without modification. They do not natively expose a \texttt{variation\_space}; interventions on these environments are instead applied at the observation boundary through \texttt{Visual Wrappers} (Sec.~\ref{app:fov}), or via a dedicated adapter when one is provided.

\subsection{Factors of Variation (FoV) \& Wrappers}
\label{app:fov}

\texttt{swm} provides two parallel mechanisms for these interventions, described next: a simulator-level mechanism for environments whose internal state we can edit, and a boundary-level mechanism based on Gymnasium wrappers for environments we treat as black boxes (Atari, Craftax). Both feed into the same evaluation pipeline and can be active on a single \texttt{World} instance.

\paragraph{Factor of Variation.}
Every native \texttt{swm} environment exposes its interventions through a hierarchical \texttt{variation\_space}. The hierarchy follows the structure of the scene: Push-T exposes factors for the agent, block, goal, and background, while MuJoCo-based environments expose factors tied to bodies, geoms, joints, cameras, and physics parameters. FoVs are configured through the standard Gymnasium \texttt{options} argument, which is forwarded by \textcolor{ForestGreen}{\texttt{World.reset()}}, \textcolor{ForestGreen}{\texttt{World.collect()}}, \textcolor{ForestGreen}{\texttt{World.evaluate()}}, \textcolor{ForestGreen}{\texttt{evaluate\_from\_dataset()}}, and related methods. Factors are referenced by dot-key names such as \texttt{agent.color}, \texttt{block.scale}, or \texttt{physics.floor.friction}, and setting \texttt{variation} to \texttt{"all"} samples every available native factor.

FoVs can be sampled randomly or pinned to fixed values. Random sampling produces train-time diversity or evaluation-time distribution shifts; fixed values give controlled ablations, reproducible sweeps, and debugging targets. In both cases the sampled values are applied at reset and held constant for the duration of the episode. That matters for evaluation: a failure can be attributed to a persistent change in the environment, not to frame-wise noise injected independently at every timestep.

\begin{lstlisting}
# Example when collecting a dataset
world.collect(
    "data/pusht_demo.h5",
    episodes=100,
    options={
        "variation": ["agent.color", "block.scale"],
        "variation_values": {
            "background.color": [0.1, 0.2, 0.9]  # fix specific value
        }
    }
)

# Example during evaluation
world.evaluate(
    policy=your_policy,
    options={"variation": ["all"]}
)
\end{lstlisting}

\begin{figure}[t]
    \centering
    \includegraphics[width=\textwidth]{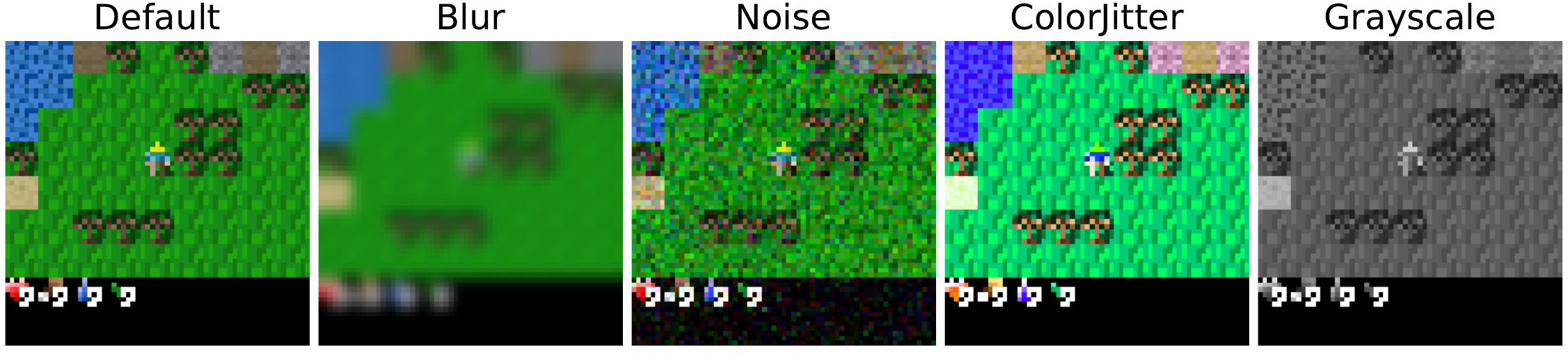}
    \includegraphics[width=\textwidth]{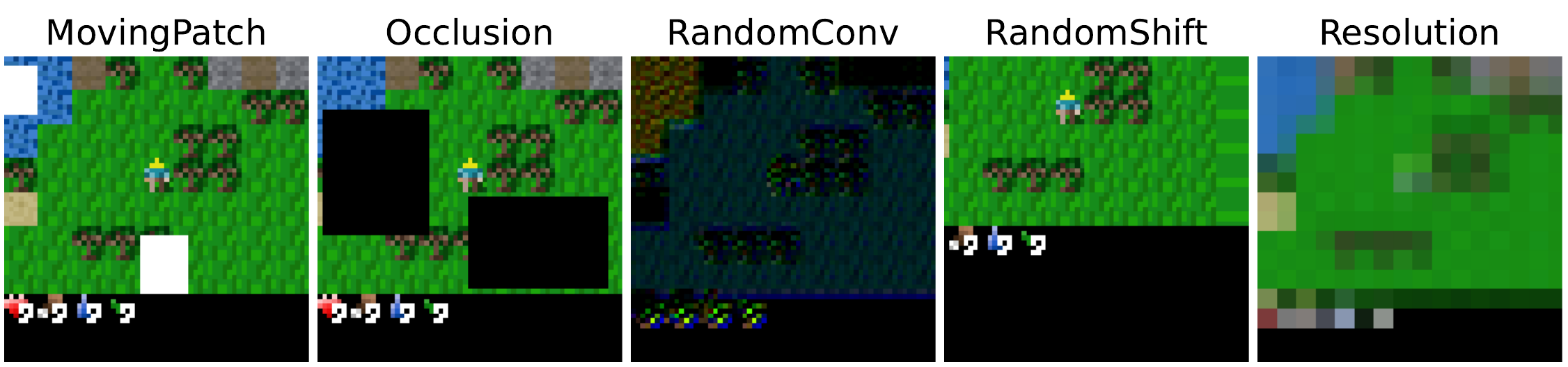}
    \caption{Examples of visual perturbations that can be applied on-the-fly to any supported environment without modifying its source code. These perturbations enable systematic robustness and generalization studies even for environments with limited source access (e.g., Atari ROMs).}
    \label{fig:wrapper}
\end{figure}

\paragraph{Boundary-level FoV via Visual Wrappers.}
Not all environments expose editable simulator internals. Atari games run from ROMs, and Craftax exposes only the observation pipeline without programmatic hooks into rendering or transitions. \texttt{swm} handles these cases through Gymnasium wrappers, used as boundary-level interventions: a wrapper sits between the environment and the agent and overwrites the relevant channel (frames, rewards, actions, or auxiliary metadata) without touching the original source. We focus here on visual wrappers, listed in Table~\ref{tab:registry-visual-wrappers}: the environment renders its RGB frame, and the wrapper rewrites that frame before it reaches the agent, policy, or world model. Figure~\ref{fig:wrapper} shows the resulting perturbations on a Craftax frame.

\begin{table}[ht]
\centering
\small
\caption{Visual wrappers available in \texttt{swm}. These wrappers operate on rendered frames and can therefore be applied even when simulator internals are inaccessible.}
\label{tab:registry-visual-wrappers}
\begin{tabular}{@{}lp{0.68\linewidth}@{}}
\toprule
Wrapper & Intervention \\
\midrule
\texttt{ChromaKeyWrapper} & Replaces pixels matching a target color with a user-provided image or video background. \\
\texttt{NoiseWrapper} & Adds Gaussian or salt-and-pepper pixel noise. \\
\texttt{BlurWrapper} & Applies Gaussian blur with a controllable kernel size. \\
\texttt{ColorJitterWrapper} & Perturbs brightness, contrast, saturation, and hue. \\
\texttt{GrayscaleWrapper} & Removes color information while preserving spatial structure. \\
\texttt{RandomShiftWrapper} & Translates the frame by a random offset. \\
\texttt{CutoutWrapper} & Masks random rectangular regions of the frame. \\
\texttt{OcclusionWrapper} & Places solid occluding patches at random locations. \\
\texttt{MovingPatchWrapper} & Adds occluding patches with independent temporal motion. \\
\texttt{RandomConvWrapper} & Applies a fixed random convolution to the frame. \\
\texttt{ResolutionWrapper} & Downsamples and upsamples frames to vary spatial fidelity. \\
\bottomrule
\end{tabular}
\end{table}

Wrappers are passed to \texttt{World} through the \texttt{extra\_wrappers} argument as a list of factories, applied in order on top of \texttt{swm}'s default \texttt{MegaWrapper}. The same call site instruments a native MuJoCo task and a closed-source ROM; only the contents of the list change.

\begin{lstlisting}
from functools import partial
import stable_worldmodel as swm
from stable_worldmodel.wrapper.visual import (
    NoiseWrapper, ColorJitterWrapper, OcclusionWrapper,
)

# Robustness sweep on Atari Pong: pixel noise, color jitter, occluders.
world = swm.World(
    "ALE/Pong-v5",
    num_envs=8,
    image_shape=(84, 84),
    extra_wrappers=[
        partial(NoiseWrapper, std=8.0),
        partial(ColorJitterWrapper, brightness=0.3, hue=0.1),
        partial(OcclusionWrapper, num_patches=2, size=(0.1, 0.25)),
    ],
)
\end{lstlisting}

Each wrapper takes its own constructor arguments (kernel size, noise standard deviation, occlusion fraction, and so on). Wrappers that resample per episode, such as \texttt{ColorJitterWrapper}, \texttt{OcclusionWrapper}, and \texttt{RandomShiftWrapper}, redraw their internal parameters at every \texttt{reset}, so a sweep produces a fresh draw per episode rather than a single fixed perturbation. The two FoV mechanisms compose: a single \texttt{World} can stack visual wrappers over a native FoV environment by passing \texttt{extra\_wrappers} at construction together with an \texttt{options=\{"variation": [...]\}} dictionary to \texttt{World.reset()}.

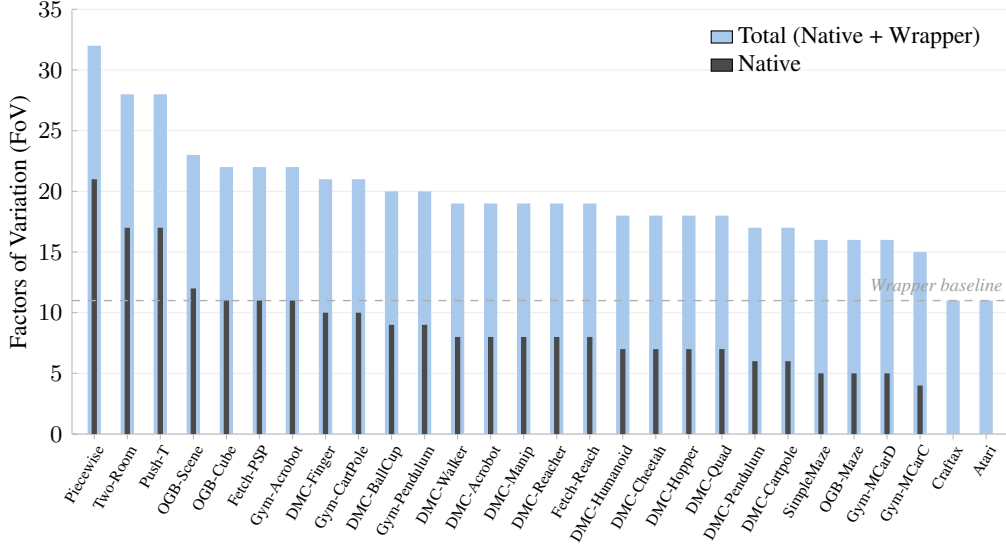
\begin{figure}[htbp]
    \centering
    \definecolor{wrapperblue}{HTML}{A8C8EC}
    \definecolor{nativegray}{HTML}{4A4A4A}
    
    \begin{tikzpicture}
        \begin{axis}[
            ybar,
            bar shift=0pt,
            width=\linewidth,
            height=7.2cm,
            ymin=0,
            ymax=35,
            ylabel={Factors of Variation (FoV)},
            ylabel style={font=\small, yshift=-4pt},
            symbolic x coords={Piecewise, Two-Room, Push-T, OGB-Scene, OGB-Cube, Fetch-PSP, Gym-Acrobot, DMC-Finger, Gym-CartPole,  DMC-BallCup, Gym-Pendulum, DMC-Walker, DMC-Acrobot, DMC-Manip, DMC-Reacher, Fetch-Reach, DMC-Humanoid, DMC-Cheetah, DMC-Hopper, DMC-Quad, DMC-Pendulum, DMC-Cartpole, SimpleMaze, OGB-Maze, Gym-MCarD, Gym-MCarC, Craftax, Atari},
            xtick=data,
            x tick label style={rotate=60, anchor=north east, inner sep=1pt, font=\tiny},
            y tick label style={font=\small},
            ytick={0,5,10,15,20,25,30,35},
            enlarge x limits=0.025,
            legend style={
                at={(0.985,0.985)},
                anchor=north east,
                draw=none,
                fill=none,
                font=\small,
                row sep=-1pt,
                /tikz/every even column/.append style={column sep=4pt}
            },
            legend cell align={left},
            legend image code/.code={
                \draw [#1] (0cm,-0.08cm) rectangle (0.28cm,0.08cm);
            },
            axis x line*=bottom,
            axis y line*=left,
            axis line style={gray!55, line width=0.4pt},
            tick style={gray!55, line width=0.4pt},
            tickwidth=2pt,
            ymajorgrids=true,
            grid style={solid, gray!15, line width=0.3pt},
        ]
        
        \addplot[fill=wrapperblue, draw=none, bar width=5pt] coordinates {
            (Piecewise, 32)
            (Two-Room, 28)
            (Push-T, 28)
            (OGB-Scene, 23)
            (OGB-Cube, 22)
            (Fetch-PSP, 22)
            (Gym-Acrobot, 22)
            (DMC-Finger, 21)
            (Gym-CartPole, 21)
            (DMC-BallCup, 20)
            (Gym-Pendulum, 20)
            (DMC-Walker, 19)
            (DMC-Acrobot, 19)
            (DMC-Manip, 19)
            (DMC-Reacher, 19)
            (Fetch-Reach, 19)
            (DMC-Humanoid, 18)
            (DMC-Cheetah, 18)
            (DMC-Hopper, 18)
            (DMC-Quad, 18)
            (DMC-Pendulum, 17)
            (DMC-Cartpole, 17)
            (SimpleMaze, 16)
            (OGB-Maze, 16)
            (Gym-MCarD, 16)
            (Gym-MCarC, 15)
            (Craftax, 11)
            (Atari, 11)
        };
        \addlegendentry{Total (Native + Wrapper)}

        \addplot[fill=nativegray, draw=none, bar width=2pt] coordinates {
            (Piecewise, 21)
            (Two-Room, 17)
            (Push-T, 17)
            (OGB-Scene, 12)
            (OGB-Cube, 11)
            (Fetch-PSP, 11)
            (Gym-Acrobot, 11)
            (DMC-Finger, 10)
            (Gym-CartPole, 10)
            (DMC-BallCup, 9)
            (Gym-Pendulum, 9)
            (DMC-Walker, 8)
            (DMC-Acrobot, 8)
            (DMC-Manip, 8)
            (DMC-Reacher, 8)
            (Fetch-Reach, 8)
            (DMC-Humanoid, 7)
            (DMC-Cheetah, 7)
            (DMC-Hopper, 7)
            (DMC-Quad, 7)
            (DMC-Pendulum, 6)
            (DMC-Cartpole, 6)
            (SimpleMaze, 5)
            (OGB-Maze, 5)
            (Gym-MCarD, 5)
            (Gym-MCarC, 4)
            (Craftax, 0)
            (Atari, 0)
        };
        \addlegendentry{Native}

        \draw[dashed, gray!65, line width=0.5pt]
            ({rel axis cs:0,0} |- {axis cs:Atari,11}) --
            ({rel axis cs:1,0} |- {axis cs:Atari,11});
        
        \node[anchor=south east, font=\scriptsize\itshape, text=gray!75, inner sep=2pt]
            at ({rel axis cs:1,0} |- {axis cs:Atari,11}) {Wrapper baseline};
        
        \end{axis}
    \end{tikzpicture}
    \caption{Factors of variation supported per environment. Light blue bars indicate total factors available (native + 11 universal visual wrappers); inset dark bars show only those exposed natively. Closed-source simulators (Atari, Craftax) lack accessible internals and rely entirely on the wrapper layer; Atari denotes the ALE suite (100+ games), which contributes no native simulator FoV here.}
    \label{fig:fov_envlist}
\end{figure}

{\bf Note:} \texttt{swm} can load by default any environment registered through Gymnasium and work with them. However,
  unless explicitly added, no Factors of Variation will be available for these non-supported environments.
  Fig.~\ref{fig:fov_envlist} summarizes the number of native and wrapper-based FoV exposed by each supported
  environment.

\section{Available Baselines}

\texttt{swm} provides six implementations of world model baselines. These cover two major paradigms for solving goal-conditioned environments: goal-conditioned reinforcement learning (GCRL) and latent world models. All baselines share a common structure, making it simple to create direct comparisons. For training, all baselines share a common Hydra-configured training entry point and matching config. All algorithms described learn offline from a dataset $D$ consisting of trajectories $(o_0, a_0, o_1, \dots, o_T)$ where $o_i$ and $a_i$ are the observation and action at time $i$ respectively. These data are collected before training and can be shared across baselines, allowing fair comparisons of offline methods. As described earlier, checkpoints live in \textcolor{ForestGreen}{\texttt{\$ STABLEWM\_HOME/checkpoints}}. GCRL baselines implement $\textcolor{ForestGreen}{\texttt{Actionable}}$ and are wrapped by a common $\textcolor{ForestGreen}{\texttt{FeedForwardPolicy}}$ with no solver. Latent world models implement $\textcolor{ForestGreen}{\texttt{Costable}}$ and are wrapped by $\textcolor{ForestGreen}{\texttt{MPCPolicy}}$, along with any solver described in Appendix \ref{app:solvers}.

\subsection{Goal-Conditioned Reinforcement Learning}
Goal-conditioned reinforcement learning follows a similar structure to classical reinforcement learning, with learned functions having an extra parameter of the current goal. The GCRL baselines implemented by \texttt{swm} share a common $\textcolor{ForestGreen}{\texttt{GCRL}}$ class which combines a DINOv2 head with, depending on the algorithm, an action predictor, a value predictor, and a critic predictor. While not in the original implementations, in \texttt{swm} observations and goals are first encoded into DINOv2 \cite{oquab2023dinov2} patch embeddings before being used as states.  All GCRL baselines act through $\textcolor{ForestGreen}{\texttt{get\_action}}$ which takes a current state and goal and returns an action. GCRL baselines do not use any predictors and, as such, do not use any solvers.

\paragraph{Goal-conditioned Behavioural Cloning} Goal-Conditioned Behavioural Cloning (GCBC) \cite{ghosh2019learning} is an imitation learning baseline that learns a goal-conditioned policy $\pi(a | s, g)$. It is trained to reproduce expert actions given a state and goal.

\begin{algorithm}[ht]
\caption{GCBC Training}
\label{alg:gcbc-training}
\DontPrintSemicolon
\KwIn{Dataset $D$ of expert trajectories, frozen encoder $\mathcal{E}$ (DINOv2), goal-conditioned policy $\pi_\theta$, history $H$, learning rate $\eta$.}
\KwOut{Policy $\pi_\theta$.}
\While{not converged}{
    Sample $(o_0, \dots, o_{H-1},\; a_0, \dots, a_{H-1},\; g) \sim D$\;
    \For{$i \leftarrow 0$ \KwTo $H - 1$}{
        $z_i \leftarrow \mathcal{E}(o_i)$\;
    }
    $z_g \leftarrow \mathcal{E}(g)$\;
    \For{$i \leftarrow 0$ \KwTo $H - 1$}{
        $\hat{a}_i \leftarrow \pi_\theta(z_i,\; z_g)$\;
    }
    $\mathcal{L} \leftarrow \dfrac{1}{H} \displaystyle\sum_{i=0}^{H-1} \|\hat{a}_i - a_i\|_2^2$\;
    $\theta \leftarrow \theta - \eta\,\nabla_\theta \mathcal{L}$\;
}
\end{algorithm}

\paragraph{Implicit Q-Learning} In implicit q-learning (IQL), a state value function is learned through expectile regression. This is done through jointly learning a critic and value function. After these networks are trained, a policy is extracted via advantage-weighted regression (AWR). \texttt{swm} implements two variations: implicit q-learning and implicit v-learning. IQL learns both a Q-function $Q_\psi(s_t,a_t,g)$ and a value function $V_\theta(s_t,g)$. The Q-network is trained with Bellman regression, bootstrapping from the target value network $V_{\bar{\theta}}$:
$$ \mathcal{L}_{Q} = \mathbb{E}_{(s_t, a_t, s_{t+1}, g) \sim \mathcal{D}} \left[ \left( Q_\psi(s_t, a_t, g) - \left( r(s_t, g) + \gamma \, m_t \, V_{\bar{\theta}}(s_{t+1}, g) \right) \right)^2 \right] $$

where $m_t = 0$ if $s_t = g$ (terminal) and $m_t = 1$ otherwise. The value network is trained with expectile regression against targets from the target Q-network $Q_{\bar{\psi}}$:

$$ \mathcal{L}_{V} = \mathbb{E}_{(s_t, a_t, g) \sim \mathcal{D}} \left[ L_\tau^2 \!\left( Q_{\bar{\psi}}(s_t, a_t, g) - V_\theta(s_t, g) \right) \right] $$

where $L_\tau^2(u) = |\tau - \mathbbm{1}(u < 0)| \, u^2$ is the asymmetric expectile loss. The total critic-phase loss is $\mathcal{L}_{\text{critic}} = \mathcal{L}_{Q} + \mathcal{L}_{V}$.

\begin{breakablealgorithm}
\caption{GCIQL Training}
\label{alg:gciql-training}
\DontPrintSemicolon
\KwIn{Dataset $D$, frozen encoder $\mathcal{E}$ (DINOv2), value head $V_\theta$ with target $V_{\bar{\theta}}$, critic $Q_\psi$ with target $Q_{\bar{\psi}}$, actor $\pi_\phi$, discount $\gamma$, expectile $\tau$, AWR temperature $\alpha$, EMA rate $\rho$, learning rate $\eta$.}
\KwOut{Value $V_\theta$, critic $Q_\psi$, actor $\pi_\phi$}
\BlankLine
\tcp{Phase 1: joint value and critic learning}
\While{not converged}{
    Sample $(o_t, a_t, o_{t+1}, g) \sim D$\;
    $z_t, z_{t+1}, z_g \leftarrow \mathcal{E}(o_t), \mathcal{E}(o_{t+1}), \mathcal{E}(g)$\;
    $m \leftarrow 1 - \mathbbm{1}[o_t = g]$\;
    $r \leftarrow -m$\;
    $q_\mathrm{tgt} \leftarrow r + \gamma\,m\,V_{\bar{\theta}}(z_{t+1}, z_g)$ 
      
    $\mathcal{L}_Q \leftarrow \big(Q_\psi(z_t, a_t, z_g) - q_\mathrm{tgt}\big)^2$\;
    \BlankLine
    $\mathcal{L}_V \leftarrow L_\tau^2\!\big(Q_{\bar{\psi}}(z_t, a_t, z_g) - V_\theta(z_t, z_g)\big)$ 
    \BlankLine
    $\mathcal{L}_\mathrm{critic} \leftarrow \mathcal{L}_Q + \mathcal{L}_V$\;
    $(\theta, \psi) \leftarrow (\theta, \psi) - \eta\,\nabla \mathcal{L}_\mathrm{critic}$\;
    $(\bar{\theta}, \bar{\psi}) \leftarrow \rho\,(\bar{\theta}, \bar{\psi}) + (1-\rho)\,(\theta, \psi)$\;
}
\BlankLine
\tcp{Phase 2: policy extraction with $V_\theta, Q_\psi$ frozen}
\While{not converged}{
    Sample $(o_t, a_t, g) \sim D$\;
    $z_t, z_g \leftarrow \mathcal{E}(o_t), \mathcal{E}(g)$\;
    $A \leftarrow Q_\psi(z_t, a_t, z_g) - V_\theta(z_t, z_g)$\;
    $w \leftarrow \exp(\alpha\,A)$\;
    $\mathcal{L}_\pi \leftarrow w \cdot \|\pi_\phi(z_t, z_g) - a_t\|_2^2$\;
    $\phi \leftarrow \phi - \eta\,\nabla_\phi \mathcal{L}_\pi$\;
}
\end{breakablealgorithm}

\paragraph{Implicit V-Learning} Implicit v-learning is an extension of IQL that learns only a value function and no q-function. It does this by training $V_\theta(s_t, g)$ directly against bootstrapped targets from a target network $V_{\bar{\theta}}$:

$$ \mathcal{L}_{V} = \mathbb{E}_{(s_t, s_{t+1}, g) \sim \mathcal{D}} \left[ L_\tau^2 \!\left( r(s_t, g) + \gamma \, V_{\bar{\theta}}(s_{t+1}, g) - V_\theta(s_t, g) \right) \right] $$

with the same expectile loss $L_\tau^2$, discount $\gamma$, and reward $r$ as IQL.

\begin{breakablealgorithm}
\caption{GCIVL Training}
\label{alg:gcivl-training}
\DontPrintSemicolon
\KwIn{Dataset $D$, frozen encoder $\mathcal{E}$ (DINOv2), value head $V_\theta$ with target $V_{\bar{\theta}}$, actor $\pi_\phi$, discount $\gamma$, expectile $\tau$, AWR temperature $\alpha$, EMA rate $\rho$, learning rate $\eta$}
\KwOut{Value $V_\theta$, critic $Q_\psi$, actor $\pi_\phi$.}
\BlankLine
\tcp{Phase 1: joint value and critic learning}
\While{not converged}{
    Sample $(o_t, a_t, o_{t+1}, g) \sim D$\;
    $z_t, z_{t+1}, z_g \leftarrow \mathcal{E}(o_t), \mathcal{E}(o_{t+1}), \mathcal{E}(g)$\;
    $m \leftarrow 1 - \mathbbm{1}[o_t = g]$\;
    $r \leftarrow -m$\;
    $q \leftarrow r + \gamma\,m\,V_{\bar{\theta}}(z_{t+1}, z_g)$\;
    \BlankLine
    $\mathcal{L}_V \leftarrow L_\tau^2\!\big(q - V_\theta(z_t, z_g)\big)$\;
    $\theta \leftarrow \theta - \eta\,\nabla_\theta \mathcal{L}_V$\;
    $\bar{\theta} \leftarrow \rho\,\bar{\theta} + (1 - \rho)\,\theta$\;

}
\BlankLine
\tcp{Phase 2: policy extraction with $V_\theta$ frozen}
\While{not converged}{
    Sample $(o_t, a_t, g) \sim D$\;
    $z_t, z_g \leftarrow \mathcal{E}(o_t), \mathcal{E}(g)$\;
    $A \leftarrow V_\theta(z_{t+1}, z_g) - V_\theta(z_t, z_g)$\;
    $w \leftarrow \exp(\alpha\,A)$\;
    $\mathcal{L}_\pi \leftarrow w \cdot \|\pi_\phi(z_t, z_g) - a_t\|_2^2$\;
    $\phi \leftarrow \phi - \eta\,\nabla_\phi \mathcal{L}_\pi$\;
}
\end{breakablealgorithm}

\subsection{Latent World Models}
Latent world models consist of an encoded $\mathcal{E}$ that maps observations to a latent space and a predictor $\mathcal{P}$ that models the latent dynamics. The encoder is a function of observations, and produces an embedding. The predictor is a function of a history of $H$ embeddings, concatenated with action embeddings, and predicts the embedding $k$ time steps into the future, where $k$ is the offset. At test time, all baselines are wrapped by a common \textcolor{ForestGreen}{\texttt{MPCPolicy}} that includes a solver Appendix \ref{app:solvers}. Each world model must implement \textcolor{ForestGreen}{\texttt{encode}}, \textcolor{ForestGreen}{\texttt{predict}}, \textcolor{ForestGreen}{\texttt{rollout}}, \textcolor{ForestGreen}{\texttt{criterion}}, and \textcolor{ForestGreen}{\texttt{get\_cost}}. The solver uses these functions to evaluate candidate action sequences by rolling out the predictor and scoring the resulting trajectories, then returns the best found action sequence. This process is described in more depth in Appendix \ref{app:solvers}. 
 
\paragraph{DINO-WM} DINO-WM \cite{zhou2024dino} is a latent world model with a frozen, pretrained DINOv2 \cite{oquab2023dinov2} encoder. Only the predictor is learned, operating over patch-level visual tokens produced by the encoder, as well as optionally proproiception embeddings. The predictor is a causal Transformer with a windowed attention mask of size $H$. It is trained by teacher-forced $\ell_2$-regression toward the DINOv2 embedding of the next frame. By freezing the encoder, DINOv2 avoids collapse common in JEPAs without regularization. DINO-WM serves as proof of latent planning using visual features, even without end-to-end encoder-predictor training.

\begin{breakablealgorithm}
\caption{DINO-WM Training}
\label{alg:dino-wm-training}
\DontPrintSemicolon
\KwIn{Dataset $D$, frozen encoder $\mathcal{E}$ (DINOv2), predictor $\mathcal{P}_\theta$, action encoder $\mathcal{A}$; history $H$, offset $k$, learning rate $\eta$}
\KwOut{Predictor $\mathcal{P}_\theta$.}
\While{not converged}{
    Sample $(o_0, \dots, o_{H+k-1},\; a_0, \dots, a_{H-1}) \sim D$\;
    \For{$i \leftarrow 0$ \KwTo $H + k - 1$}{
        $z_i \leftarrow \mathcal{E}(o_i)$ 
    }
    \For{$i \leftarrow 0$ \KwTo $H - 1$}{
        $e_i \leftarrow \mathrm{concat}(z_i,\; \mathcal{A}(a_i))$\;
    }
    $(\hat{z}_k, \dots, \hat{z}_{H+k-1}) \leftarrow \mathcal{P}_\theta(e_0, \dots, e_{H-1})$ 
    
    $\mathcal{L} \leftarrow \dfrac{1}{H} \displaystyle\sum_{i=0}^{H-1} \big\| \hat{z}_{i+k} - z_{i+k} \big\|_2^2$\;
    
    $\theta \leftarrow \theta - \eta\,\nabla_\theta \mathcal{L}$\;
}
\end{breakablealgorithm}

\paragraph{Planning with Latent Dynamics Models} Planning with Latent Dynamics Models (PLDM) \cite{sobal2025learning} is an end-to-end JEPA. The encoder and predictor are trained jointly from raw observations. The encoder produces a single \texttt{[CLS]}-token embedding, as opposed to full patch-level visual tokens, that is used by the predictor. To prevent representational collapse, PLDM uses a multi-term objective, combining teacher-forced prediction loss with VICReg-style variance and covariance regularizers \cite{bardes2021vicreg}, a temporal alignment term, and an inverse dynamics term, each with their own independent weight.

\begin{breakablealgorithm}
\caption{PLDM Training}
\label{alg:pldm-training}
\DontPrintSemicolon
\KwIn{Dataset $D$, encoder $\mathcal{E}_\theta$, predictor $\mathcal{P}_\theta$, action encoder $\mathcal{A}_\theta$, inverse dynamics model $\mathrm{IDM}_\psi$, history $H$, loss weights $\alpha, \beta, \delta, \omega$, learning rate $\eta$}
\KwOut{Encoder $\mathcal{E}_\theta$ and predictor $\mathcal{P}_\theta$.}
\While{not converged}{
    Sample $(o_0, \dots, o_T,\; a_0, \dots, a_{T-1}) \sim D$\;
    \For{$i \leftarrow 0$ \KwTo $T$}{
        $z_i \leftarrow \mathcal{E}_\theta(o_i)$\;
    }
    $(\hat{z}_1, \dots, \hat{z}_T) \leftarrow \mathcal{P}_\theta\big(z_{0:T-1},\; \mathcal{A}_\theta(a_{0:T-1})\big)$\;
    \BlankLine
    $\mathcal{L}_\mathrm{sim} \leftarrow \dfrac{1}{T} \displaystyle\sum_{i=1}^{T} \|\hat{z}_i - z_i\|_2^2$\;
    \BlankLine
    $\mathcal{L}_\mathrm{std} \leftarrow \mathrm{VarLoss}(z_{0:T})$\;
    \BlankLine
    $\mathcal{L}_\mathrm{cov} \leftarrow \mathrm{CovLoss}(z_{0:T})$\;
    \BlankLine
    $\mathcal{L}_\mathrm{temp} \leftarrow \dfrac{1}{T} \displaystyle\sum_{i=0}^{T-1} \|z_i - z_{i+1}\|_2^2$\;
    \For{$i \leftarrow 0$ \KwTo $T - 1$}{
        $\hat{a}_i \leftarrow \mathrm{IDM}_\psi(z_i,\, z_{i+1})$\;
    }
    $\mathcal{L}_\mathrm{idm} \leftarrow \dfrac{1}{T} \displaystyle\sum_{i=0}^{T-1} \|\hat{a}_i - a_i\|_2^2$\;
    \BlankLine
    $\mathcal{L} \leftarrow \mathcal{L}_\mathrm{sim} + \alpha\,\mathcal{L}_\mathrm{std} + \beta\,\mathcal{L}_\mathrm{cov} + \delta\,\mathcal{L}_\mathrm{temp} + \omega\,\mathcal{L}_\mathrm{idm}$\;
    $(\theta, \psi) \leftarrow (\theta, \psi) - \eta\,\nabla\mathcal{L}$\;
}
\end{breakablealgorithm}

\paragraph{LeWorldModel} LeWorldModel (LeWM) \cite{maes2026leworldmodel} follows the same architecture as PLDM. It is an end-to-end JEPA that jointly trains its encoder and predictor. Instead of the five-term anti-collapse objective, LeWM uses a single regularizer: SIGReg, which pushes the distribution of the latent embeddings to an isotropic Gaussian. As a result, in practice,  it only has one effective hyperparameter that requires tuning: the weight of regularization.

\begin{breakablealgorithm}
\caption{LeWM Training}
\label{alg:lewm-training}
\DontPrintSemicolon
\KwIn{Dataset $D$, encoder $\mathcal{E}_\theta$, predictor $\mathcal{P}_\theta$, action encoder $\mathcal{A}_\theta$, SIGReg weight $\lambda$, learning rate $\eta$}
\KwOut{Encoder $\mathcal{E}_\theta$ and predictor $\mathcal{P}_\theta$.}
\While{not converged}{
    Sample $(o_0, \dots, o_T,\; a_0, \dots, a_{T-1}) \sim D$\;
    \For{$i \leftarrow 0$ \KwTo $T$}{
        $z_i \leftarrow \mathcal{E}_\theta(o_i)$\;
    }
    $(\hat{z}_1, \dots, \hat{z}_T) \leftarrow \mathcal{P}_\theta\big(z_{0:T-1},\; \mathcal{A}_\theta(a_{0:T-1})\big)$\;
    \BlankLine
    $\mathcal{L}_\mathrm{pred} \leftarrow \dfrac{1}{T} \displaystyle\sum_{i=1}^{T} \|\hat{z}_i - z_i\|_2^2$\;
    \BlankLine
    $\mathcal{L}_\mathrm{SIGReg} \leftarrow \mathrm{SIGReg}(z_{0:T})$ 
    \BlankLine
    $\mathcal{L} \leftarrow \mathcal{L}_\mathrm{pred} + \lambda\,\mathcal{L}_\mathrm{SIGReg}$\;
    $\theta \leftarrow \theta - \eta\,\nabla_\theta \mathcal{L}$\;
}
\end{breakablealgorithm}

\paragraph{TD-MPC2} TD-MPC2 \cite{hansen2023td} is a reward-driven implicit world model. It consists of an encoder, latent dynamics model, reward predictor, $Q$ function, and stochastic policy prior, all jointly trained. At test time, actions are selected by sample-based planning that bootstraps with the learned terminal value. The model is trained to match predicted next latents with the (stop-gradded) encodings of the corresponding next observations, as done in other latent world models. Additionally, the model is trained to minimize a cross-entropy loss for reward and value prediction and a maximum-entropy policy loss.

\begin{breakablealgorithm}
\caption{TD-MPC2 Training}
\label{alg:tdmpc2-training}
\DontPrintSemicolon
\KwIn{Dataset $D$ of trajectories with observations, actions, and rewards;
encoder $h$, dynamics $d$, reward $R$, Q-ensemble $\{Q_i\}_{i=1}^{N_Q}$, target ensemble $\{\bar Q_i\}_{i=1}^{N_Q}$, policy prior $p$ (parameters $\theta$);
horizon $H$, temporal weight $\rho$, discount $\gamma$;
loss weights $\beta_c, \beta_r, \beta_v$; entropy coefficient $\beta_\pi$;
EMA rate $\tau$; running scale $S$; learning rate $\eta$.}
\KwOut{Trained world model $\theta$ (encoder, dynamics, reward, Q-ensemble, policy prior).}
\BlankLine
\While{not converged}{
    Sample $(o_0, \dots, o_H,\; a_0, \dots, a_{H-1},\; r_0, \dots, r_{H-1}) \sim D$\;
    \For{$i \leftarrow 0$ \KwTo $H$}{
        $\tilde z_i \leftarrow h(o_i)$ 
    }
    $z \leftarrow \tilde z_0$\;
    $\mathcal{L}_\mathrm{c}, \mathcal{L}_\mathrm{r}, \mathcal{L}_\mathrm{v}, \mathcal{L}_\pi \leftarrow 0$\;
    \BlankLine
    \For{$t \leftarrow 0$ \KwTo $H - 1$}{
        $z'_t \leftarrow d(z, a_t)$\;
        $\hat r_t \leftarrow R(z, a_t)$\;
        \BlankLine
        $\mathcal{L}_\mathrm{c} \mathrel{+}= \rho^t\, \| z'_t - \mathrm{sg}(\tilde z_{t+1}) \|_2^2$\;
        \BlankLine
        $\mathcal{L}_\mathrm{r} \mathrel{+}= \rho^t\, \mathrm{CE}\!\left(\hat r_t,\; \mathrm{TwoHot}(r_t)\right)$\;
        \BlankLine
        \Begin(\nllabel{tdmpc-q-target}){
            $a'_t \sim p(\cdot \mid \tilde z_{t+1})$ \;
            $\{i, j\} \subset \{1, \dots, N_Q\}$ \;
            $q^\star \leftarrow \min\!\big(\bar Q_i(\tilde z_{t+1}, a'_t),\; \bar Q_j(\tilde z_{t+1}, a'_t)\big)$\;
            $y_t \leftarrow r_t + \gamma\, q^\star$\;
        }
        \For{$i \leftarrow 1$ \KwTo $N_Q$}{
            $\mathcal{L}_\mathrm{v} \mathrel{+}= \rho^t\, \mathrm{CE}\!\left(Q_i(z, a_t),\; \mathrm{TwoHot}(y_t)\right)$\;
        }
        \BlankLine

        $a^\pi_t \sim p(\cdot \mid \mathrm{sg}(z))$\;
        $\mathcal{H}_t \leftarrow$ closed-form entropy of $p$\;
        $\bar q^\pi_t \leftarrow \tfrac{1}{2}\!\left(Q_{i'}(z, a^\pi_t) + Q_{j'}(z, a^\pi_t)\right)$ for random $\{i',j'\}$\;
        \If{$t = 0$}{ Update $S$ from quantiles of $\bar q^\pi_t$ }
        $\mathcal{L}_\pi \mathrel{+}= \rho^t\, \big(\!-\bar q^\pi_t / S - \beta_\pi |\mathcal{A}|\, \mathcal{H}_t\big)$\;
        \BlankLine
        $z \leftarrow z'_t$ 
    }
    \BlankLine
    $\mathcal{L} \leftarrow \beta_c\, \mathcal{L}_\mathrm{c} + \beta_r\, \mathcal{L}_\mathrm{r} + \beta_v\, \mathcal{L}_\mathrm{v} / N_Q + \mathcal{L}_\pi$\;
    Update $\theta \leftarrow \theta - \eta\, \nabla_\theta \mathcal{L}$\;
    \BlankLine
    \For{$i \leftarrow 1$ \KwTo $N_Q$}{
        $\bar Q_i \leftarrow (1-\tau)\, \bar Q_i + \tau\, Q_i$\;
    }
}
\end{breakablealgorithm}

\section{Evaluation Protocols}
\label{app:evaluation_protocol}
\label{app:eval}

\paragraph{Goal-observation tasks.}
\texttt{swm} evaluates goal-conditioned agents by specifying tasks through observations rather than through environment-specific reward code. The agent receives the current observation together with a target frame representing the desired state, and is judged on whether its actions bring the environment to that goal from a given initial condition. This convention is shared by planners built on latent world models and by goal-conditioned policies: the policy interface differs between methods, but the evaluation question is the same. Given a start observation and a goal observation, the agent must reach the goal within a fixed interaction budget.

\paragraph{Episodic evaluation.}
\texttt{world.evaluate(episodes=$N$, seed=$s$)} runs the standard online protocol. The environment is reset through the usual Gymnasium reset path, the task is sampled by the environment's default randomiser, and the agent solves each episode from scratch. Terminated environments are auto-reset until $N$ episodes have completed, and the returned dictionary contains \texttt{success\_rate}, per-episode success flags, and the seeds used. This mode is appropriate when the benchmark itself defines the distribution of initial states, goals, and termination conditions, and it is the natural protocol for environments whose goals are generated procedurally at reset time or for robustness sweeps where FoV options and wrappers are applied directly to the live simulator.

\paragraph{Dataset-driven Evaluation.}
\texttt{world.evaluate(dataset=$\mathcal{D}$, episodes\_idx=[$i_1$,\ldots], start\_steps=[$t_1$,\ldots], goal\_offset=$\Delta$, eval\_budget=$B$)} runs the dataset-conditioned protocol used in DINO-WM~\citep{zhou2024dino}, LeWM~\citep{maes2026leworldmodel}, and PLDM~\citep{sobal2025learning}. Instead of sampling an arbitrary initial state and goal online, both are sampled from an existing trajectory: if a dataset trajectory is written as $\tau^i=(o^i_0,\ldots,o^i_T)$, evaluation initialises the environment at $o^i_t$ and uses $o^i_{t+\Delta}$ as the target observation. The offset $\Delta$ controls how far the goal lies from the start, and the dataset split controls the evaluation distribution. Because the target was observed later in the same trajectory, the goal is reachable by construction under the recorded behaviour, which removes a major ambiguity in goal-conditioned planning benchmarks: failure can be interpreted as a planning or modelling failure rather than as an impossible start--goal pair.

This setting is useful for comparing world models under controlled distribution shift. The same policy or planner can be evaluated on expert training trajectories, held-out expert trajectories, random-policy trajectories, or trajectories collected under FoV perturbations, while keeping the start--goal construction fixed. Experiments can vary reachability horizon and distribution shift independently: \texttt{goal\_offset} changes the temporal distance to the goal, and the chosen dataset changes the visual, behavioural, or physical regime from which starts and goals are drawn.

\paragraph{Budget.}
Both protocols cap each episode at \texttt{eval\_budget} environment steps. The cap keeps wall-clock cost comparable across methods that use different planners, control frequencies, or per-step computation, regardless of whether one method (e.g.\ MPPI with large \texttt{num\_samples}) consumes more per-step compute than another.

\paragraph{Reported metrics.}
For each (method, environment, FoV-setting) we report success rate, time-to-goal among successful episodes, and wall-clock latency per planning step. The gap between dataset-driven and episodic success rate serves as a proxy for over-reliance on in-distribution behavioural coverage~\cite{stone2021distracting}.

\section{Planning Solvers}
\label{app:solvers}
In \texttt{swm}, solvers are responsible for computing the optimal action that the policy should execute. To do this, \texttt{swm} implements a suite of planning solvers that optimize action sequences by leveraging a world model to evaluate costs. Specifically, all solvers receive the current latent state $\vs_0$ and use a world model $(\mathcal{E}_\theta,\mathcal{P}_\theta)$ to evaluate the cost of an action sequence $A = (\va_{0},\dots,\va_{H-1})$ over a planning horizon $H$:
\begin{equation}
    J_\theta(\vs_0, A)= \sum^{H-1}_{t=0}c(\vs_t, \va_t),
\end{equation}

where $\vs_{t+1}=\mathcal{P}_\theta(\vs_t,\va_t)$, and $c(\vs_t, \va_t)$ is a task-specific goal-reaching cost. Each solver's objective is to return an action sequence $A^\star$ that minimizes $J_\theta$:
\begin{equation}
    A^\star=\argmin_A J_\theta(\vs_0, A).
\end{equation}

The implemented solvers cover two broad families of approaches, sampling-based (zeroth-order) and gradient-based (first-order), and also span both continuous and discrete action regimes. Each solvers' algorithm is described below. See Table \ref{tab:solvers-comparison} for a complete comparison of each solver.

\begin{table}[ht]
\centering
\small
\caption{Selecting a solver. ``Diff.\ model required'' means the solver back-propagates through \texttt{get\_cost}.}
\label{tab:solvers-comparison}
\begin{tabular}{@{}lllll@{}}
\toprule
Solver & Family & Action space & Diff.\ model required & Constraints \\
\midrule
\texttt{PredictiveSamplingSolver} & sampling & continuous          & no  & continuous box \\
\texttt{CEMSolver}                & sampling & continuous          & no  & continuous box \\
\texttt{ICEMSolver}               & sampling & continuous          & no  & continuous box \\
\texttt{MPPISolver}               & sampling & continuous          & no  & continuous box \\
\texttt{CategoricalCEMSolver}     & sampling & discrete            & no  & categorical simplex \\
\texttt{GradientSolver}           & gradient & continuous          & yes & continuous box \\
\texttt{PGDSolver}                & gradient & discrete            & yes & simplex projection \\
\texttt{GRASP}                    & gradient & continuous          & yes & continuous box \\
\texttt{LagrangianSolver}         & gradient & continuous + ineq.\ & yes & differentiable inequalities \\
\bottomrule
\end{tabular}
\end{table}

\subsection{Sampling-Based Solvers}
Sampling-based methods iteratively draw populations of candidate action sequences. The candidates are evaluated by the cost $J_\theta$, and refined by slowly moving toward low-cost sequences. These methods are zeroth-order: they only require forward evaluations of the world model and do not assume that $J_\theta$ is differentiable. This makes them broadly applicable across learned models, handcrafted costs, and environments with discontinuous or non-smooth objectives.

The main trade-off is the curse of dimensionality. Since sampling-based methods do not use gradient information, they often require a sufficiently large population to explore the action space. In \texttt{swm}, candidate rollouts are executed in parallel, making these methods effective when batched on GPU.

\paragraph{Predictive sampling} Predictive Sampling \cite{howell2022predictive} is a single-shot sampling algorithm that works by repeatedly perturbing a nominal action sequence with Gaussian noise, storing the cost of the perturbed sequences, and returning the lowest-cost candidate.

Unlike CEM, iCEM, or MPPI, Predictive Sampling does not fit a new sampling distribution and does not perform iterative refinement inside a solver call. Its optimization behavior comes from random search around the nominal plan and from repeated MPC replanning over time.

\begin{breakablealgorithm}
\caption{Predictive Sampling Solver}
\label{alg:predictive-sampling-solver}
\DontPrintSemicolon

\KwIn{
World model cost $J_\theta$, current state $s_0$, horizon $H$,
number of candidates $N$, noise scale $\sigma$,
optional nominal sequence $A^{\mathrm{nom}}$
}
\KwOut{
Action sequence $A^\star = (a_0^\star,\dots,a_{H-1}^\star)$
}

$\bar A \leftarrow A^{\mathrm{nom}}$ if provided, otherwise $0_{H \times d_a}$\;
$\epsilon_1 \leftarrow 0$\;

\For{$i \leftarrow 2$ \KwTo $N$}{
    $\epsilon_i \sim \mathcal N(0, \sigma^2 I_{H d_a})$\;
}

\For{$i \leftarrow 1$ \KwTo $N$}{
    $A_i \leftarrow \bar A + \epsilon_i$\;
    $C_i \leftarrow J_\theta(s_0, A_i)$\;
}

$i^\star \leftarrow
\arg\min_{i \in \{1,\dots,N\}} C_i$\;

\Return{$A^\star \leftarrow A_{i^\star}$}\;

\end{breakablealgorithm}

\paragraph{Cross-entropy method} The Cross-Entropy Method (CEM)~\cite{rubinstein2004cross} maintains a sampling distribution over action sequences. In \texttt{swm}, this distribution is a diagonal Gaussian with mean $\mu^\ell$ and coordinatewise standard deviation $\sigma^\ell$ over the full planning horizon. At each CEM iteration, the solver samples candidate action sequences, evaluates them with $J_\theta$, selects the $E$ lowest-cost candidates as elites, and refits the Gaussian parameters to those elites.

This iterative refitting concentrates probability mass around low-cost regions of the action sequences. The mean controls exploitation, while the standard deviation controls exploration and shrinks as the elites converge. After the final iteration, the solver returns the final mean sequence.

\begin{breakablealgorithm}
\caption{Cross-Entropy Method Solver}
\label{alg:cem-solver}
\DontPrintSemicolon

\KwIn{
World model cost $J_\theta$, current state $s_0$, horizon $H$,
number of candidates $N$, iterations $L$, elites $E$,
initial scale $\sigma_0$, optional initial sequence $A^{\mathrm{init}}$
}
\KwOut{
Action sequence $A^\star = (a_0^\star,\dots,a_{H-1}^\star)$
}

$\mu^0 \leftarrow A^{\mathrm{init}}$ if provided, otherwise $0_{H \times d_a}$\;
$\sigma^0 \leftarrow \sigma_0 \mathbf 1$\;

\For{$\ell \leftarrow 0$ \KwTo $L-1$}{
    $\epsilon_1^\ell \leftarrow 0$\;

    \For{$i \leftarrow 2$ \KwTo $N$}{
        $\epsilon_i^\ell \sim \mathcal N(0, I_{H d_a})$\;
    }

    \For{$i \leftarrow 1$ \KwTo $N$}{
        $A_i^\ell \leftarrow
        \mu^\ell + \sigma^\ell \odot \epsilon_i^\ell$\;
        $C_i^\ell \leftarrow J_\theta(s_0, A_i^\ell)$\;
    }

    $\mathcal E^\ell \leftarrow$ Select top $E$ elites that minimize $\{C_i^\ell\}_{i=1}^N$\;

    $\mu^{\ell+1}
    \leftarrow
    \frac{1}{E}
    \sum_{i \in \mathcal E^\ell} A_i^\ell$\;

    $\sigma^{\ell+1}
    \leftarrow
    \operatorname{Std}_{i \in \mathcal E^\ell}
    \left(A_i^\ell\right)$\;
}

\Return{$A^\star \leftarrow \mu^L$}\;
\end{breakablealgorithm}

\paragraph{Model Predictive Path Integral} Model Predictive Path Integral (MPPI) control~\cite{williams2016aggressive} is another sampling-based optimizer, but instead of selecting a hard elite set and refitting a Gaussian, it updates the nominal sequence using a soft cost-weighted average of sampled candidates. Each candidate receives a weight proportional to $\exp\left(-\frac{C_i - C_{\min}}{\lambda}\right)$,
where $C_i$ is the predicted cost and $\lambda$ is a temperature parameter. Lower temperatures place most of the weight on the best candidates, while higher temperatures average over a broader set of samples.

In \texttt{swm}, MPPI samples candidates around a nominal mean sequence and updates only the mean, while the sampling scale remains fixed. It also supports applying the MPPI weighting to only the top-$k$ candidates, which interpolates between MPPI-style soft weighting and more elite-focused behavior as in CEM. This makes MPPI a smooth alternative to CEM: rather than discarding non-elite candidates completely, it assigns them exponentially smaller influence based on their cost.

\begin{breakablealgorithm}
\caption{Model Predictive Path Integral Solver}
\label{alg:mppi-solver}
\DontPrintSemicolon

\KwIn{
Cost $J_\theta$, current state $s_0$, horizon $H$,
number of candidates $N$, iterations $L$, top-$k$ size $E$,
temperature $\lambda$, sampling scale $\sigma_0$,
optional initial sequence $A^{\mathrm{init}}$
}
\KwOut{
Action sequence $A^\star = (a_0^\star,\dots,a_{H-1}^\star)$
}

$\mu^0 \leftarrow A^{\mathrm{init}}$ if provided, otherwise $0_{H \times d_a}$\;
$\sigma \leftarrow \sigma_0 \mathbf 1$\;

\For{$\ell \leftarrow 0$ \KwTo $L-1$}{
    $\epsilon_1^\ell \leftarrow 0$\;

    \For{$i \leftarrow 2$ \KwTo $N$}{
        $\epsilon_i^\ell \sim \mathcal N(0, I_{H d_a})$\;
    }

    \For{$i \leftarrow 1$ \KwTo $N$}{
        $A_i^\ell \leftarrow
        \mu^\ell + \sigma \odot \epsilon_i^\ell$\;
        $C_i^\ell \leftarrow J_\theta(s_0, A_i^\ell)$\;
    }

    $\mathcal E^\ell \leftarrow$ Select top $E$ that minimize $\{C_i^\ell\}_{i=1}^N$\;

    $C_{\min}^\ell
    \leftarrow
    \min_{i \in \mathcal E^\ell} C_i^\ell$\;

    \For{$i \in \mathcal E^\ell$}{
        $w_i^\ell
        \leftarrow
        \dfrac{
        \exp\left(-(C_i^\ell - C_{\min}^\ell)/\lambda\right)
        }{
        \sum_{j \in \mathcal E^\ell}
        \exp\left(-(C_j^\ell - C_{\min}^\ell)/\lambda\right)
        }$\;
    }

    $\mu^{\ell+1}
    \leftarrow
    \sum_{i \in \mathcal E^\ell}
    w_i^\ell A_i^\ell$\;
}

\Return{$A^\star \leftarrow \mu^L$}\;

\end{breakablealgorithm}

\paragraph{Improved Cross-Entropy Method} Improved CEM (iCEM)\cite{pinneri2021sample} is based on CEM and adds colored-noise sampling, elite retention iterations, and momentum on distribution updates to give better sample efficiency, especially under small planning budgets. 

Colored noise produces smoother action sequences by correlating perturbations across time, rather than sampling each action independently. This is useful for control tasks where good behavior often requires planning over several steps. Elite retention reuses a small number of strong candidates from the previous CEM iteration, preventing useful samples from being discarded too quickly. Momentum smooths the updates to $\mu^\ell$ and $\sigma^\ell$, reducing instability when the elite set is small.

\begin{breakablealgorithm}
\caption{Improved Cross-Entropy Method Solver}
\label{alg:icem-solver}
\DontPrintSemicolon

\KwIn{
World model cost $J_\theta$, current state $s_0$, horizon $H$,
number of candidates $N$, iterations $L$, elites $E$,
initial scale $\sigma_0$, colored-noise exponent $\beta$,
momentum $\alpha$, retained elites $E_{\mathrm{keep}}$,
action bounds $\mathcal A$, optional initial sequence $A^{\mathrm{init}}$
}
\KwOut{
Action sequence $A^\star = (a_0^\star,\dots,a_{H-1}^\star)$
}

$\mu^0 \leftarrow A^{\mathrm{init}}$ if provided, otherwise $\mathbf{0}_{H \times d_a}$\;
$\sigma^0 \leftarrow \sigma_0 \mathbf 1$\;
$\mathcal M^0 \leftarrow \emptyset$\;

\For{$\ell \leftarrow 0$ \KwTo $L-1$}{
    $\xi_1^\ell \leftarrow 0$\;

    \For{$i \leftarrow 2$ \KwTo $N$}{
        $\xi_i^\ell \sim \mathcal C_\beta(H,d_a)$\;
    }

    \For{$i \leftarrow 1$ \KwTo $N$}{
        $A_i^\ell
        \leftarrow
        \mu^\ell + \sigma^\ell \odot \xi_i^\ell$\;
    }

    \If{$\ell > 0$}{
        $r \leftarrow \min(E_{\mathrm{keep}}, |\mathcal M^\ell|)$\;
        Replace $A_2^\ell,\dots,A_{1+r}^\ell$
        with the best $r$ retained elites from $\mathcal M^\ell$\;
    }

    \For{$i \leftarrow 1$ \KwTo $N$}{
        $A_i^\ell \leftarrow \operatorname{clip}_{\mathcal A}(A_i^\ell)$\;
        $C_i^\ell \leftarrow J_\theta(s_0, A_i^\ell)$\;
    }

    $\mathcal E^\ell \leftarrow$ Select top $E$ elites that minimize $\{C_i^\ell\}_{i=1}^N$\;

    $\mathcal M^{\ell+1}
    \leftarrow
    \{A_i^\ell : i \in \mathcal E^\ell\}$, ordered by increasing cost\;

    $\hat \mu^\ell
    \leftarrow
    \frac{1}{E}
    \sum_{i \in \mathcal E^\ell} A_i^\ell$\;

    $\hat \sigma^\ell
    \leftarrow
    \operatorname{Std}_{i \in \mathcal E^\ell}
    \left(A_i^\ell\right)$\;

    $\mu^{\ell+1}
    \leftarrow
    \alpha \mu^\ell + (1-\alpha)\hat\mu^\ell$\;

    $\sigma^{\ell+1}
    \leftarrow
    \alpha \sigma^\ell + (1-\alpha)\hat\sigma^\ell$\;
}
\Return{$A^\star \leftarrow \mu^L$}\;
\end{breakablealgorithm}

\paragraph{Categorical CEM.}
Categorical CEM is the discrete-action analog of CEM. Instead of maintaining a Gaussian distribution over continuous action sequences, it maintains an independent categorical distribution over the discrete action set at each planning step. Candidate sequences are sampled from these categorical distributions, converted to one-hot action representations, and evaluated with $J_\theta$. The elite candidates are then used to update the categorical probabilities through empirical frequencies, optionally with smoothing and momentum. This solver is useful when actions are discrete and a continuous relaxation is not desired (in contrast to PGD).

\begin{breakablealgorithm}
\caption{Categorical Cross-Entropy Method Solver}
\label{alg:categorical-cem-solver}
\DontPrintSemicolon

\KwIn{
World model cost $J_\theta$, current state $s_0$, discrete action set $\mathcal A$,
horizon $H$, action blocks $B_{\mathrm{act}}$, number of candidates $N$,
iterations $L$, elites $E$, smoothing $\delta$, momentum $\alpha$
}
\KwOut{
Discrete action sequence $A^\star = (a_0^\star,\dots,a_{H-1}^\star)$
}

Initialize categorical probabilities
$\pi^0_{t,b}(a) \leftarrow 1 / |\mathcal A|$
for all $t = 0,\dots,H-1$, $b = 1,\dots,B_{\mathrm{act}}$, and $a \in \mathcal A$\;

\For{$\ell \leftarrow 0$ \KwTo $L-1$}{
    \For{$i \leftarrow 1$ \KwTo $N$}{
        Sample $A_i^\ell = \{a_{i,t,b}^\ell\}_{t,b}$
        from $\pi^\ell$ using Gumbel-max sampling\;
    }

    Set the first candidate to the current mode:
    
        $a_{1,t,b}^\ell
        \leftarrow
        \arg\max_{a \in \mathcal A}
        \pi^\ell_{t,b}(a)$\;

    \For{$i \leftarrow 1$ \KwTo $N$}{
        $Y_i^\ell \leftarrow \operatorname{OneHot}(A_i^\ell)$\;
        $C_i^\ell \leftarrow J_\theta(s_0, Y_i^\ell)$\;
    }

    $\mathcal E^\ell
    \leftarrow
    \text{indices of the } E \text{ lowest costs in }
    \{C_i^\ell\}_{i=1}^{N}$\;

    Refit categorical probabilities from elite frequencies:
    
        $\hat\pi^\ell_{t,b}(a)
        \leftarrow
        \frac{1}{E}
        \sum_{i \in \mathcal E^\ell}
        \mathbf 1[a_{i,t,b}^\ell = a]$\;

    \If{$\delta > 0$}{
            $\hat\pi^\ell_{t,b}(a)
            \leftarrow
            \frac{
                \hat\pi^\ell_{t,b}(a) + \delta
            }{
                1 + |\mathcal A|\delta
            }$\;
    }

    Update probabilities with momentum:
        $\pi^{\ell+1}
        \leftarrow
        \alpha \pi^\ell + (1-\alpha)\hat\pi^\ell$ \;
}

    $a_{t,b}^\star
    \leftarrow
    \arg\max_{a \in \mathcal A}
    \pi^L_{t,b}(a)$\;

\Return{$A^\star$}\;

\end{breakablealgorithm}

\subsection{Gradient-Based Solvers}
Gradient-based solvers work by directly optimizing the action sequence $A$ with respect to the world model cost $J_\theta$ via first-order gradient information. Importantly, because they require $J_\theta$ to be differentiable, the world model predictor $\mathcal{P}_\theta$ must also be differentiable. Gradient-based solvers can be more sample-efficient than sampling-based methods because each update gives a directed improvement signal.

The main limitation of gradient-based solvers is that learned world models can have poorly conditioned gradients, especially over long horizons. Back-propagating through multiple predicted steps can produce unstable updates and trap the optimizer in local minima. \texttt{swm} includes simple baselines, such as Gradient Descent and PGD, as well as more structured methods like GRASP that modify the planning objective to improve optimization.

\paragraph{Gradient descent} The gradient descent (GD) is a first-order algorithm that optimizes an action sequence by iteratively minimizing the differentiable world model cost $J_\theta$. The gradient with respect to $A$ is repeatedly applied to the action sequence. The solver initializes one or more candidate sequences and repeatedly updates each candidate in the direction of lower cost. After the final gradient step, it evaluates the optimized candidates and returns the lowest-cost sequence. The implementation can use standard optimizers such as Adam, optional gradient clipping, and multiple random initial candidates.

\begin{breakablealgorithm}
\caption{Gradient Descent Solver}
\label{alg:gd-solver}
\DontPrintSemicolon

\KwIn{
Differentiable world model cost $J_\theta$, current state $s_0$, horizon $H$,
number of candidates $N$, iterations $K$, step size $\eta$
}
\KwOut{
Action sequence $A^\star = (a_0^\star,\dots,a_{H-1}^\star)$
}

Initialize candidates $\{A_i^0\}_{i=1}^N$, with $A_i^0 \in \mathbb R^{H \times d_a}$\;

\For{$k \leftarrow 0$ \KwTo $K-1$}{
    \For{$i \leftarrow 1$ \KwTo $N$}{
        $A_i^{k+1}
        \leftarrow
        A_i^k - \eta \nabla_{A^k_i} J_\theta(s_0, A_i^k)$\;
    }
}

$i^\star \leftarrow
\arg\min_{i \in \{1,\dots,N\}} J_\theta(s_0, A_i^K)$\;

\Return{$A^\star \leftarrow A_{i^\star}^K$}\;

\end{breakablealgorithm}

\paragraph{Projected Gradient Descent} Projected Gradient Descent (PGD) extends gradient-based planning to discrete action spaces by optimizing a continuous relaxation of the discrete actions. Instead of choosing a discrete action directly at each timestep, the solver optimizes a probability vector over the action set $\mathcal A$. Thus, the optimization variables lie in a product of simplexes, $P \in \Delta(\mathcal A)^H$.

At each step, PGD takes a gradient step on the relaxed action probabilities and then projects each probability vector back onto the simplex using the projection algorithm of Duchi et al.~\cite{duchi2008projection}. This guarantees that every relaxed action remains a valid distribution over discrete actions. After optimization, the relaxed sequence is converted back to a discrete action sequence, either by taking the maximum probability action at each timestep or by sampling from the optimized distributions.

\begin{breakablealgorithm}
\caption{Projected Gradient Descent Solver}
\label{alg:pgd-solver}
\DontPrintSemicolon

\KwIn{
Differentiable world model cost $J_\theta$, current state $s_0$, discrete action set $\mathcal A$,
horizon $H$, number of candidates $N$, iterations $K$, step size $\eta$,
initial scale $\sigma_0$, optional action-noise scale $\sigma_{\mathrm{act}}$,
optional initial sequence $A^{\mathrm{init}}$
}
\KwOut{
Discrete action sequence $A^\star = (a_0^\star,\dots,a_{H-1}^\star)$
}

$\bar P \leftarrow \operatorname{OneHot}(A^{\mathrm{init}})$ if provided,
otherwise $\bar P \leftarrow \frac{1}{|\mathcal A|}\mathbf 1 \in \Delta(\mathcal A)^H$\;

$P_1^0 \leftarrow \bar P$\;

\For{$i \leftarrow 2$ \KwTo $N$}{
    $\epsilon_i \sim \mathcal N(0, \sigma_0^2 I_{H|\mathcal A|})$\;
    $P_i^0 \leftarrow
    \Pi_{\Delta(\mathcal A)^H}
    \left(\bar P + \epsilon_i\right)$\;
}

\For{$k \leftarrow 0$ \KwTo $K-1$}{
    \For{$i \leftarrow 1$ \KwTo $N$}{
        $\xi_i^k \leftarrow 0$\;

        \If{$\sigma_{\mathrm{act}} > 0$}{
            $\xi_i^k \sim \mathcal N(0, \sigma_{\mathrm{act}}^2 I_{H|\mathcal A|})$\;
        }

        $P_i^{k+1}
        \leftarrow
        \Pi_{\Delta(\mathcal A)^H}
        \left(
        P_i^k
        -
        \eta \nabla_P J_\theta(s_0, P_i^k)
        +
        \xi_i^k
        \right)$\;
    }
}

$i^\star
\leftarrow
\arg\min_{i \in \{1,\dots,N\}}
J_\theta(s_0, P_i^K)$\;

\For{$t \leftarrow 0$ \KwTo $H-1$}{
    $a_t^\star
    \leftarrow
    \arg\max_{a \in \mathcal A}
    \left(P_{i^\star,t,a}^K\right)$\;
}

\Return{$A^\star \leftarrow (a_0^\star,\dots,a_{H-1}^\star)$}\;

\end{breakablealgorithm}

\paragraph{Lagrangian} The Lagrangian solver extends gradient-based planning to inequality-constrained objectives. In addition to a differentiable cost $J_\theta$, the world model may expose constraint functions $g_j(s_0,A) \le 0$. The solver then optimizes an augmented Lagrangian objective of the form
\begin{equation*}
    \mathcal L(A,\lambda,\rho) = J_\theta(s_0,A) + \sum_j \lambda_j g_j(s_0,A) + \rho \sum_j \max(0,g_j(s_0,A))^2.    
\end{equation*}
The inner loop performs gradient-based optimization of the action sequence, while the outer loop updates the nonnegative multipliers $\lambda_j$ by dual ascent and increases the penalty coefficient $\rho$. This solver is useful when a task includes explicit differentiable constraints, such as action-norm limits, safety margins, and other user-defined conditions.

\begin{breakablealgorithm}
\caption{Lagrangian Solver}
\label{alg:lagrangian-solver}
\DontPrintSemicolon

\KwIn{
Differentiable cost $J_\theta$, differentiable constraints $g_\theta$,
current state $s_0$, horizon $H$, number of candidates $N$,
outer iterations $L$, inner gradient steps $K$, optimizer step size $\eta$,
initial sampling scale $\sigma_0$, penalty parameters
$\rho_0,\rho_{\max},\rho_{\mathrm{scale}}$,
optional action-noise scale $\sigma_{\mathrm{act}}$,
optional initial sequence $A^{\mathrm{init}}$
}
\KwOut{
Action sequence $A^\star = (a_0^\star,\dots,a_{H-1}^\star)$
}

Initialize nominal sequence
$\bar A \leftarrow A^{\mathrm{init}}$ if provided, otherwise $0_{H \times d_a}$\;

$A_1 \leftarrow \bar A$\;

\For{$i \leftarrow 2$ \KwTo $N$}{
    $\epsilon_i \sim \mathcal N(0,\sigma_0^2 I_{H d_a})$\;
    $A_i \leftarrow \bar A + \epsilon_i$\;
}

Initialize multipliers $\lambda^0 \leftarrow 0$ and penalty $\rho^0 \leftarrow \rho_0$\;

\For{$\ell \leftarrow 0$ \KwTo $L-1$}{
    \For{$k \leftarrow 0$ \KwTo $K-1$}{
        \For{$i \leftarrow 1$ \KwTo $N$}{
            $C_i \leftarrow J_\theta(s_0, A_i)$\;
            $G_i \leftarrow g_\theta(s_0, A_i)$\;
        }

        Form the augmented Lagrangian objective:

            $\mathcal L_{\mathrm{aug}}
            \leftarrow
            \sum_{i=1}^{N}
            \left[
                C_i
                +
                \lambda^\ell \cdot G_i
                +
                \rho^\ell
                \left\|
                    [G_i]_+
                \right\|_2^2
            \right]$\;

        Update all candidate action sequences:
        
            $A_i
            \leftarrow
                A_i - \eta\nabla_{A_i} \mathcal L_{\mathrm{aug}},\quad i=1,\dots, N
                $\;

        \If{$\sigma_{\mathrm{act}} > 0$}{
            \For{$i \leftarrow 1$ \KwTo $N$}{
                $\xi_i \sim \mathcal N(0,\sigma_{\mathrm{act}}^2 I_{H d_a})$\;
                $A_i \leftarrow A_i + \xi_i$\;
            }
        }
    }

    Re-evaluate constraint violations $G_i \leftarrow g_\theta(s_0,A_i)$ for $i=1,\dots,N$\;

    Update Lagrange multipliers by dual ascent:
    
        $\lambda^{\ell+1}
        \leftarrow
        \left[
            \lambda^\ell
            +
            \rho^\ell
            \frac{1}{N}
            \sum_{i=1}^{N}
            G_i
        \right]_+$\;

    Increase the quadratic penalty:

        $\rho^{\ell+1}
        \leftarrow
        \min
        \left(
            \rho_{\max},
            \rho_{\mathrm{scale}}\rho^\ell
        \right)$\;
}

\For{$i \leftarrow 1$ \KwTo $N$}{
    $C_i \leftarrow J_\theta(s_0,A_i)$\;
}

$i^\star
\leftarrow
\arg\min_{i \in \{1,\dots,N\}} C_i$\;

\Return{$A^\star \leftarrow A_{i^\star}$}\;

\end{breakablealgorithm}

\paragraph{GRASP} GRASP \cite{psenka2026parallel} jointly optimizes an action sequence and intermediate virtual states $z_1,\dots,z_{H-1}$ as additional optimization variables. For each state $z_{t+1}$, it tries to minimize distance to the goal state, while not deviating from the state value predicted by the world model $\hat z_{t+1}^k \leftarrow \mathcal{P}_\theta\!\left(\operatorname{sg}(z_t^k), a_t^k\right)$. Periodically, the implementation synchronizes the action sequence with the standard rollout cost $J_\theta$, refining the current plan under the original MPC objective. Because the one-step dynamics terms can be evaluated in parallel across time, GRASP is especially useful for longer-horizon settings where ordinary rollout-based gradient descent becomes difficult to optimize.

\begin{breakablealgorithm}
\caption{GRASP Solver}
\label{alg:grasp-solver}
\DontPrintSemicolon

\KwIn{
Encoder $E_\theta$; predictor $\mathcal{P}_\theta$; World model cost $J_\theta$;
current observation $o_0$; goal observation $o_g$;
horizon $H$; iterations $K$;
action/state step sizes $\eta_a,\eta_s$;
goal weights $\{\gamma_k\}_{k=0}^{K-1}$;
state-noise scales $\{\sigma_k\}_{k=0}^{K-1}$;
sync interval $K_{\mathrm{sync}}$;
sync operator $\operatorname{Sync}_{J_\theta}$;
optional initial sequence $A^{\mathrm{init}}$
}
\KwOut{
Action sequence $A^\star = (a_0^\star,\dots,a_{H-1}^\star)$
}

$s_0 \leftarrow E_\theta(o_0)$, \quad $s_g \leftarrow E_\theta(o_g)$\;

$A^0 \leftarrow A^{\mathrm{init}}$ if provided, otherwise $0_{H \times d_a}$\;

\For{$t \leftarrow 1$ \KwTo $H-1$}{
    $z_t^0 \leftarrow
    \left(1-\frac{t}{H}\right)s_0
    +
    \frac{t}{H}s_g$\;
}
Fix $z_0^k \equiv s_0$ and $z_H^k \equiv s_g$\;

\BlankLine

\For{$k \leftarrow 0$ \KwTo $K-1$}{
    Predict each transition in parallel:

        $\hat z_{t+1}^k
        \leftarrow
        \mathcal{P}_\theta\!\left(\operatorname{sg}(z_t^k), a_t^k\right),
        \qquad t = 0,\dots,H-1$\;

        $\mathcal L_k
        \leftarrow
        \sum_{t=0}^{H-1}
        \left(
        \left\|\hat z_{t+1}^k - z_{t+1}^k\right\|_2^2
        +
        \gamma_k
        \left\|\hat z_{t+1}^k - s_g\right\|_2^2
        \right)$\;
    
        $A^{k+1}
        \leftarrow
        A^k-
        {\eta_a}
        \nabla_{A^k}\mathcal L_k; $\;

        $z_{1:H-1}^{k+1}
        \leftarrow
        z_{1:H-1}^k-
        {\eta_z}
        \nabla_{z_{1:H-1}^k}\mathcal L_k; $\;

    Add stochastic state exploration:
        
        $\xi_t^k \sim \mathcal N(0,I)$

        $z_t^{k+1}
        \leftarrow
        z_t^{k+1} + \sigma_k \xi_t^k,
        \qquad
        t=1,\dots,H-1$\;

    \If{$K_{\mathrm{sync}} > 0$ and $k > 0$ and $(k+1) \bmod K_{\mathrm{sync}} = 0$}{
        $A^{k+1}
        \leftarrow
        \operatorname{Sync}_{J_\theta}
        \left(A^{k+1}; s_0, s_g\right)$\;
    }
}

\Return{$A^\star \leftarrow A^K$}\;

\end{breakablealgorithm}

\section{Scripts \& Command Line Interface}

To further streamline the daily research workflow, \texttt{swm} provides a comprehensive suite of self-contained scripts alongside a dedicated CLI.
\paragraph{Scripts.}
Rather than requiring users to write repetitive boilerplate from scratch, the platform comes with ready-to-use, Hydra-configured scripts that cover the entire experimental pipeline~\citep{Yadan2019Hydra}. Specifically, we provide scripts for efficient multimodal data collection across any supported environment, as well as scripts to train SAC and TD-MPC2 expert policies for generating high-quality demonstration data. In addition, \texttt{swm} includes standardised scripts to train and evaluate established baselines, ranging from imitation learning (GCBC) and offline RL (IQL, IVL) to modern world models (e.g. TD-MPC2, DINO-WM, PLDM, LeWM).
\paragraph{Command-Line Interface.} 
The \texttt{swm} CLI lets researchers audit their workspace and data directly from the terminal, without writing any Python code. It exposes the following commands:
\begin{itemize}
    \setlength{\itemsep}{0pt}
    \setlength{\parsep}{0pt}
    \setlength{\topsep}{0pt}
    \item \texttt{swm datasets} and \texttt{swm inspect}: report metadata on saved trajectories, including disk footprint, episode lengths, and the exact tensor shapes of stored columns.
    \item \texttt{swm envs} and \texttt{swm fovs}: list all available environments and display their customisable factors of variation.
    \item \texttt{swm checkpoints}: view and manage locally cached model weights.
\end{itemize}

\section{Extending \texttt{stable-worldmodel}}
\label{app:extension}

\texttt{swm} is meant as a community effort. As an open-source project, we strongly encourage contributions: new environments, new world models, or new planning solvers. We also plan to extend the benchmarking of included baselines and maintain a public leaderboard, so the state of the art in WM research is clearly tracked, and practitioners are guided in their choice of baselines.

\paragraph{New environment.} Implement the standard \texttt{gymnasium.Env} interface and register the class under the \texttt{swm/} namespace. To unlock factor-of-variation studies, additionally expose a hierarchical \texttt{variation\_space} built from \texttt{swm}'s extended spaces (\texttt{Box}, \texttt{Discrete}, \texttt{RGBBox}, \texttt{Dict}, etc.), each supporting an \texttt{init\_value} and an optional \texttt{constrain\_fn} for rejection sampling. The \texttt{info} dictionary returned by \texttt{reset} and \texttt{step} should carry any extra signals the user wishes to log or evaluate against; standard Gymnasium quantities (observation, action, termination, reward, frames) are recorded automatically. Once registered, the environment is immediately usable through \texttt{World}, the \texttt{swm} CLI, and every built-in evaluation routine.

\paragraph{New planning solver.} A solver is any object exposing \texttt{configure(action\_space, n\_envs, config)} and \texttt{solve(info\_dict, init\_action)}, the latter returning a dict containing the optimized action sequence. New solvers are added by dropping a single file under \texttt{stable\_worldmodel/solver/}.

\paragraph{New baseline.} New baselines should ship as self-contained code, together with a training entry point under \texttt{scripts/train/}, mirroring the existing baselines (e.g.\ \texttt{pldm.py}, \texttt{gcivl.py}). At evaluation time, the baseline plugs into MPC through the \texttt{Costable} protocol: a single method \texttt{get\_cost(info\_dict, action\_candidates) -> Tensor} of shape \texttt{(B, S)}.

\section{Experimental setup}
\label{app:exp_details}

This section details the specific configurations used in the experiments of Sec.~\ref{sec:experiments}. We follow the dataset-driven evaluation protocol described in App.~\ref{app:eval} (\texttt{evaluate\_from\_dataset}), which specifies how start observations and goal observations are constructed from existing trajectories; here we focus only on the concrete settings. 

\paragraph{Tasks and dataset.} All experiments are conducted on the Push-T benchmark using the expert dataset released with DINO-WM~\citep{zhou2024dino}. Following the protocol of LeWM~\citep{maes2026leworldmodel}, for each evaluation start and goal observations are drawn from the same trajectory with a fixed temporal offset of $\Delta = 25$ environment steps. Distribution-shift sweeps additionally vary the source dataset (expert train, expert validation, random-policy, random-policy with FoV) while keeping $\Delta$ fixed.

\paragraph{Evaluation budget and metrics.} Each episode is capped at $B = 50$ environment steps budget and we report success rate as defined in DINO WM~\citep{zhou2024dino}. In-distribution comparisons (Tab.~\ref{tab:baselines_sr}) use 50 evaluation trajectories; the prediction-vs-success analyses (Fig.~\ref{fig:lewm_pred_vs_sr},~\ref{fig:pldm_pred_vs_sr}) use 256 trajectories per regime for stable distributional estimates.

\paragraph{Planning.} Unless otherwise noted, the planner is CEM (App.~\ref{app:solvers}, Alg.~\ref{alg:cem-solver}) with $L = 30$ iterations, $N = 300$ candidates, $E = 30$ elites, and initial sampling scale $\sigma_0 = 1$. World models are trained with a frameskip of $5$; planning therefore uses a horizon of $H = 5$ model steps ($25$ environment steps) and is executed in open-loop mode, i.e.\ the full optimized action sequence is rolled out before replanning. Each planning experiment runs in a few minutes on a single L40S GPU thanks to \texttt{swm}'s efficient implementation.

\paragraph{Baselines.} The robustness analyses (Tab.~\ref{tab:factor_variation_short}, Fig.~\ref{fig:lewm-wheel}, Fig.~\ref{fig:pusht-distractor}) primarily compare PLDM~\citep{sobal2025learning} and LeWM~\citep{maes2026leworldmodel}, which combine strong in-distribution planning performance with low per-step planning overhead, making systematic FoV sweeps tractable. DINO-WM~\citep{zhou2024dino} is additionally reported in Tab.~\ref{tab:factor_variation_short}. Training hyperparameters for each baseline follow the original publications; the exact configs are released alongside the code

\section{Additional Results}
\label{app:additional_exp}

This section presents supplementary plots from the experiments in Section~\ref{sec:experiments}. All results follow the standardized evaluation protocol of \texttt{stable-worldmodel} on the Push-T benchmark unless otherwise noted.

Table~\ref{tab:factor_variation_long} provides a highly granular breakdown of the zero-shot generalization experiments by isolating individual factors of variation. Rather than reporting a single aggregate metric for "out-of-distribution" performance, this disentangled view reveals exactly which physical or visual properties break the learned representations. For instance, we observe that altering the target anchor's position or the agent's size drastically reduces planning success across all baselines. The models exhibit varying sensitivities: PLDM is remarkably robust to agent shape variations (52.0\% success rate) compared to the others, but fails heavily when the agent color is perturbed.

\begin{figure*}[t]
    \centering
    \setlength{\tabcolsep}{5pt}
    \captionof{table}{Planning success rates (\%) under individual factors of variation on Push-T evaluated on random-policy OOD trajectories.}
    \label{tab:factor_variation_long}
    \begin{tabular}{@{}l l c c c@{}}
        \toprule
        \textbf{FoV} & \textbf{Entity} & \multicolumn{3}{c}{\textbf{SR \%
        ($\uparrow$)}} \\
        \cmidrule(lr){3-5}
                 &             & \textbf{LeWM} & \textbf{PLDM} & \textbf{DINO-WM} \\
        \midrule
        \textbf{None }    &             &\textbf{ 50.8} & \textbf{50.8} & \textbf{20.0} \\
        \midrule
        Color    & Anchor      & 14.0 & 10.0 & 20.0 \\
                 & Agent       & 12.0 &  8.0 & 18.0 \\
                 & Block       & 22.0 & 18.0 & 18.0 \\
                 & Canvas      &  6.0 &  6.0 & 10.0 \\
        \addlinespace
        Size     & Anchor      & 26.0 & 18.0 & 14.0 \\
                 & Agent       & 22.0 & 18.0 &  4.0 \\
                 & Block       & 20.0 & 18.0 & 16.0 \\
        \addlinespace
        Angle    & Anchor      & 20.0 & 24.0 & 12.0 \\
                 & Agent       & 14.0 & 22.0 & 12.0 \\
        \addlinespace
        Position & Anchor      & 32.0 & 18.0 &  4.0 \\
        \addlinespace
        Shape    & Agent       & 26.0 & 52.0 & 18.0 \\
                 & Block       & 12.0 & 14.0 &  8.0 \\
        \addlinespace
        Velocity & Agent       & 16.0 & 16.0 & 14.0 \\
        \bottomrule
    \end{tabular}
\end{figure*}

\begin{figure}[ht]
    \centering
    \begin{subfigure}[b]{0.35\textwidth}
        \centering
        \includegraphics[width=\textwidth]{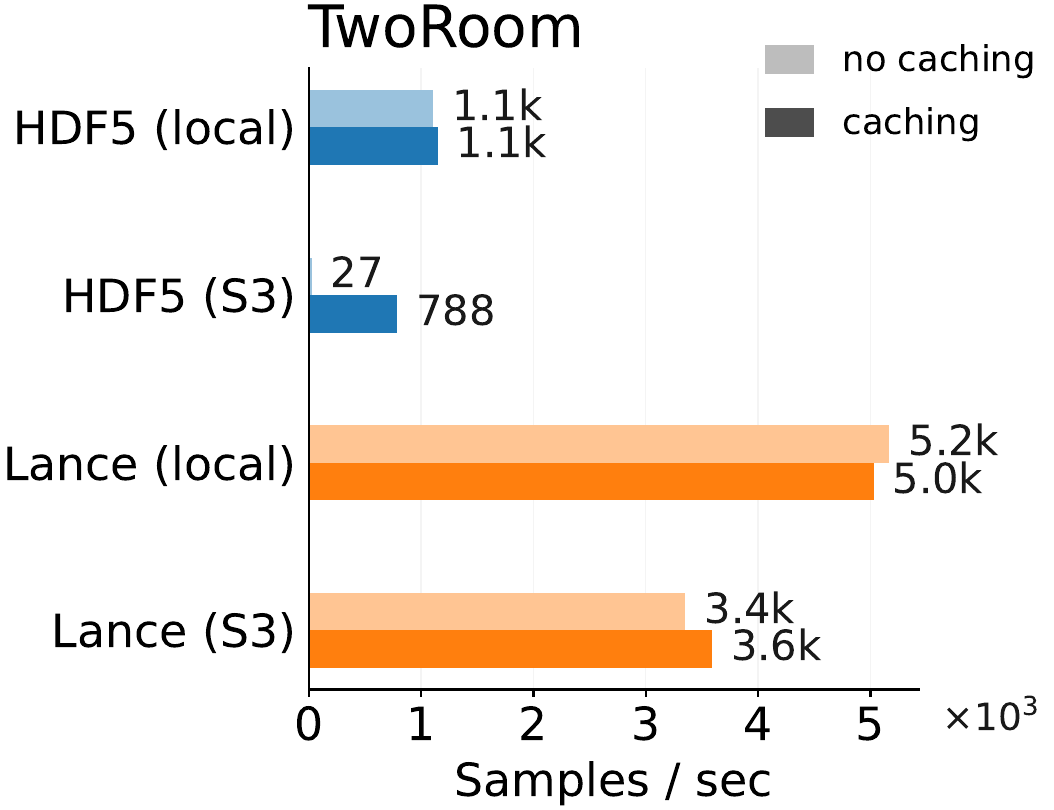}
        \label{fig:sub0}
    \end{subfigure}
    \hfill
    \begin{subfigure}[b]{0.35\textwidth}
        \centering
        \includegraphics[width=\textwidth]{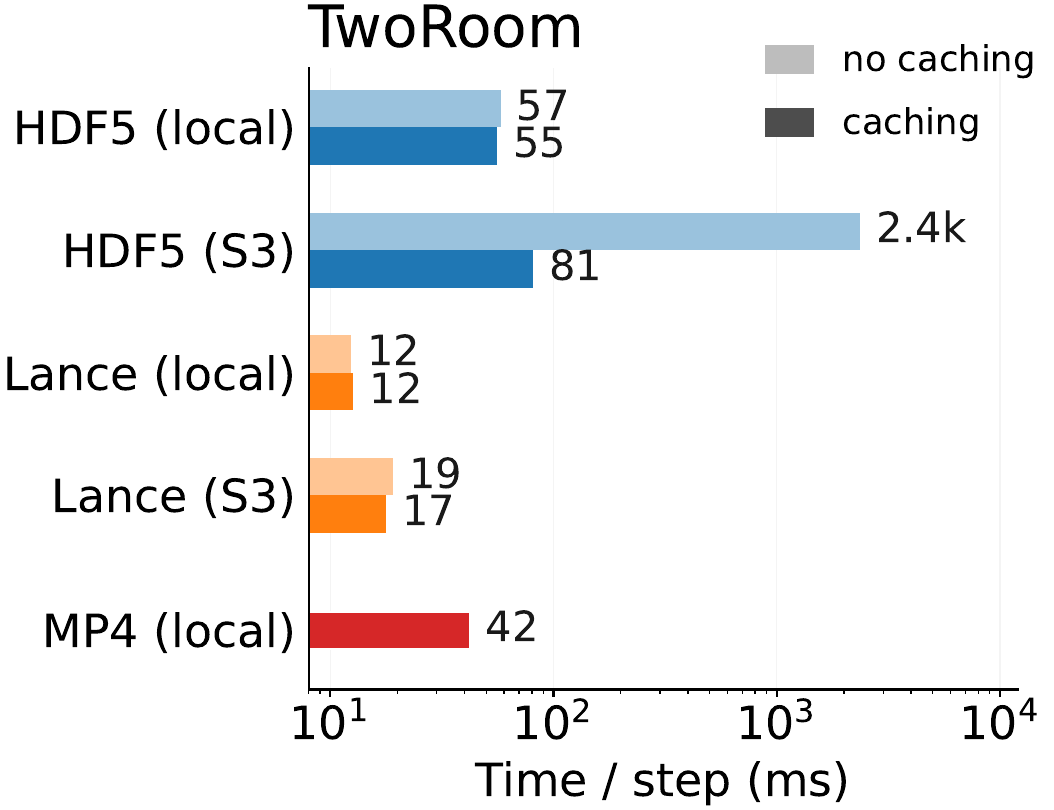}
        \label{fig:sub1}
    \end{subfigure}
    \hfill
    \begin{subfigure}[b]{0.275\textwidth}
        \centering
        \includegraphics[width=\textwidth]{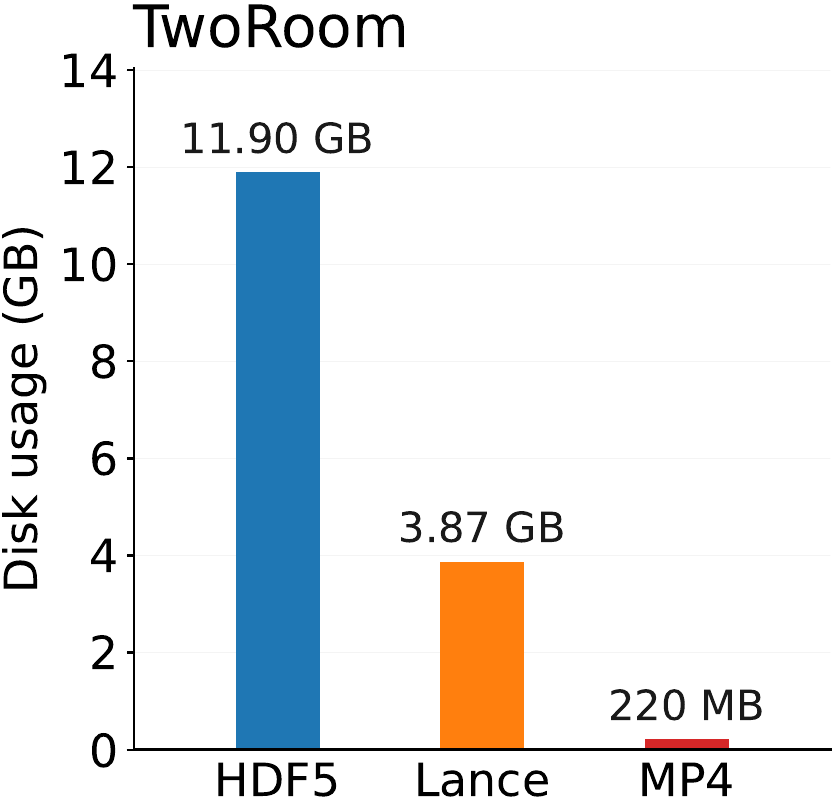}
        \label{fig:sub2}
    \end{subfigure}
    \caption{Performance comparison of different data formats for a dataset from the Two-Room environment. {\bf(Left)} Data loading throughput (samples/sec) with and without caching, for both local storage and remote (S3) streaming. {\bf(Center)} Per-step loading latency (ms). {\bf(Right)} Disk storage usage. Results demonstrate that Lance is the most efficient format in terms of throughput and fetch latency while remaining a good trade-off in compression for storage.}
    \label{fig:tworoom-data}
\end{figure}

The data loading benchmarks for the Two-Room environment are presented in Figure~\ref{fig:tworoom-data}, confirming that our I/O optimizations generalize across different task structures and observation shapes. The results show that the Lance format maintains superior throughput, processing over 5.0k samples per second locally and 3.4k samples per second when streaming from a remote S3 bucket. While the MP4 format achieves the lowest disk storage footprint (220 MB), it severely degrades random access since retrieving a later frame often requires decoding all preceding frames in the clip. Lance thus provides the optimal trade-off between disk compression and high-speed random access.

\begin{figure}[tbp]
    \centering
    \begin{subfigure}[b]{0.45\textwidth}
        \centering
        \includegraphics[width=\textwidth]{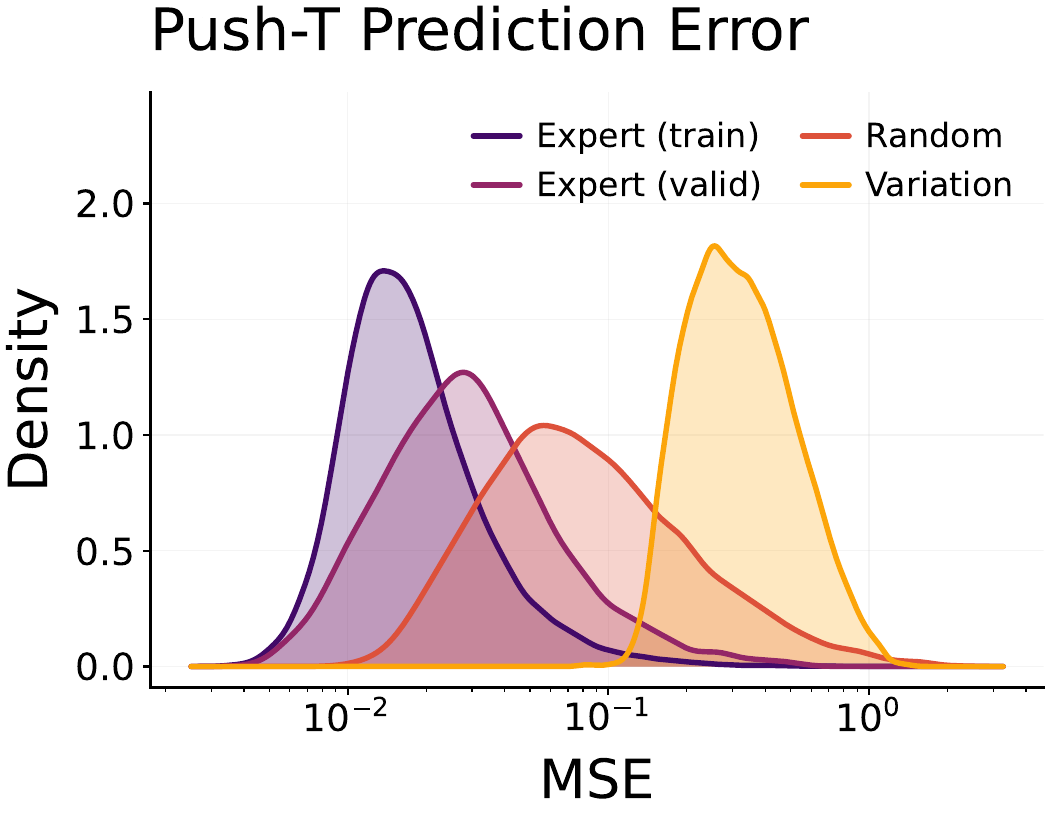}
        \caption{PLDM}
        \label{fig:pldm_pred_error}
    \end{subfigure}
    \hfill
    \begin{subfigure}[b]{0.45\textwidth}
        \centering
        \includegraphics[width=\textwidth]{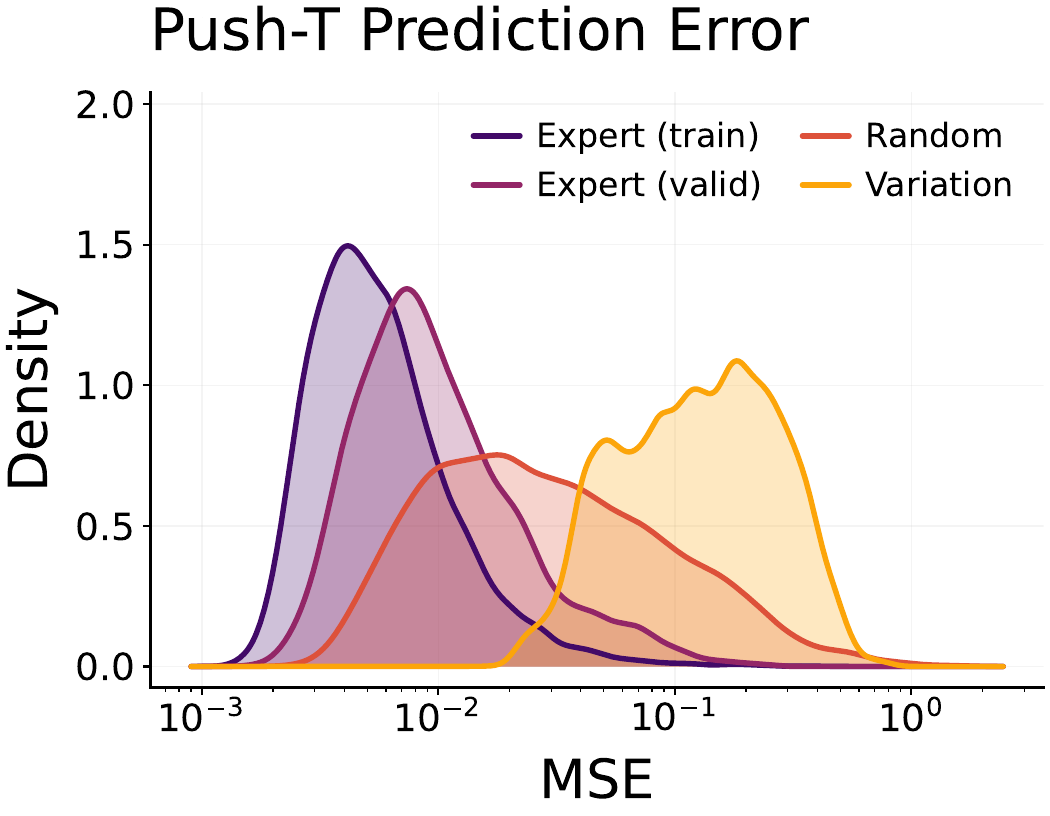}
        \caption{LeWM}
        \label{fig:lewm_pred_error}
    \end{subfigure}
    \caption{Push-T prediction error under increasing distribution shift for \textbf{(a)} PLDM and \textbf{(b)} LeWM . Density of per-trajectory MSE across four regimes of growing OOD severity: expert train, expert validation, random-policy, and random-policy with all \texttt{swm} factors of variation jointly perturbed. Both models exhibit an error increase as the evaluation set shifts from training.}
    \label{fig:pred_error_comparison}
\end{figure}

\begin{figure}[ht]
    \centering
    \includegraphics[width=\textwidth]{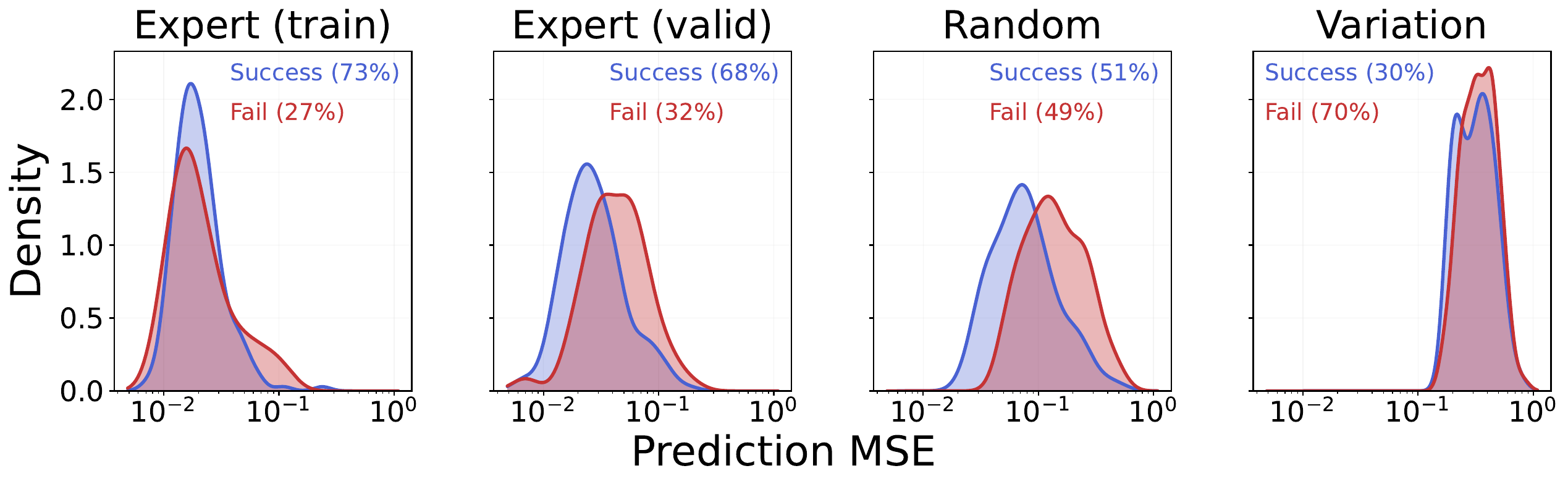}
    \caption{PLDM counterpart of Fig.~\ref{fig:lewm_pred_vs_sr}. Distribution of trajectory-level prediction MSE for successful (blue) and failed (red) plans on Push-T, across four regimes of increasing distribution shift (expert train, expert validation, random-policy, and random-policy with all \texttt{swm} factors of variation). Evaluated on 256 trajectories per regime. As with LeWM, the success and failure distributions overlap heavily even under a strong shift, confirming that prediction error is a poor predictor of planning success.}
    \label{fig:pldm_pred_vs_sr}
\end{figure}

Figures~\ref{fig:pred_error_comparison} and~\ref{fig:pldm_pred_vs_sr} expand on the prediction-vs-control analysis. Figure~\ref{fig:pred_error_comparison} shows that the per-trajectory MSE distributions for both PLDM and LeWM flatten and shift toward higher errors as the evaluation regime drifts from training, moving from expert train to expert validation, then to random policy, and finally to random policy with full variations. Figure~\ref{fig:pldm_pred_vs_sr} provides the corresponding success/failure breakdown for PLDM: even under strong distribution shift, the MSE densities of successful and failed rollouts overlap heavily, so a low prediction error does not guarantee planning success.

\begin{figure}[t]
    \centering
    \includegraphics[width=0.5\textwidth]{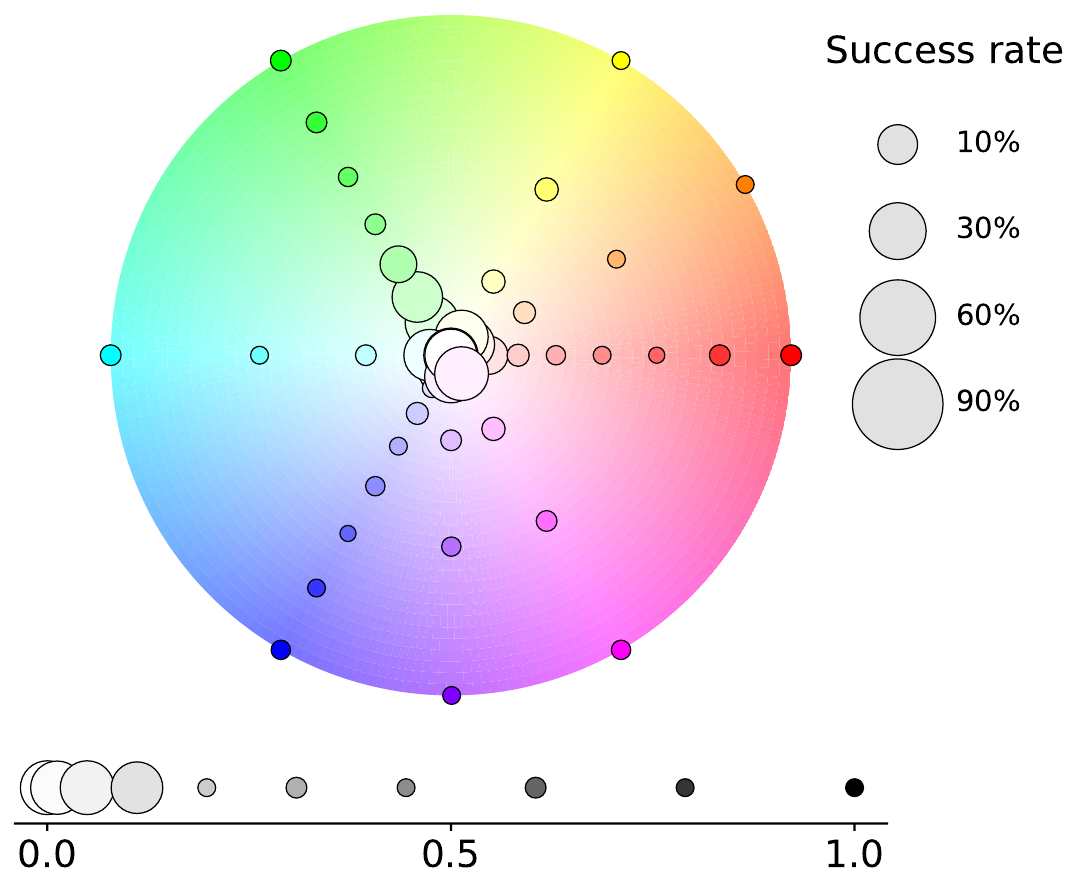}
    \caption{Planning success rate of LeWM on Push-T as a function of background color intensity along the chromatic wheel. Performance remains high near the original white background and is notably more robust to green backgrounds, likely due to the green visual anchor present in the default environment. Success degrades sharply for most other colors.}
    \label{fig:lewm-wheel}
\end{figure}

Figure~\ref{fig:lewm-wheel} provides a continuous view of this brittleness by reporting LeWM's planning success rate across the chromatic wheel. Performance stays high near the center (close to the white background) and along the green axis, which matches the color of the target anchor in the unperturbed Push-T environment. As the background shifts toward red, blue, or purple at higher intensities, the success rate collapses, suggesting that the model latches onto specific background-foreground color contrasts rather than the underlying task geometry.

\begin{figure}[htbp]
    \centering
    \begin{subfigure}[b]{0.32\textwidth}
        \centering
        \includegraphics[width=\textwidth]{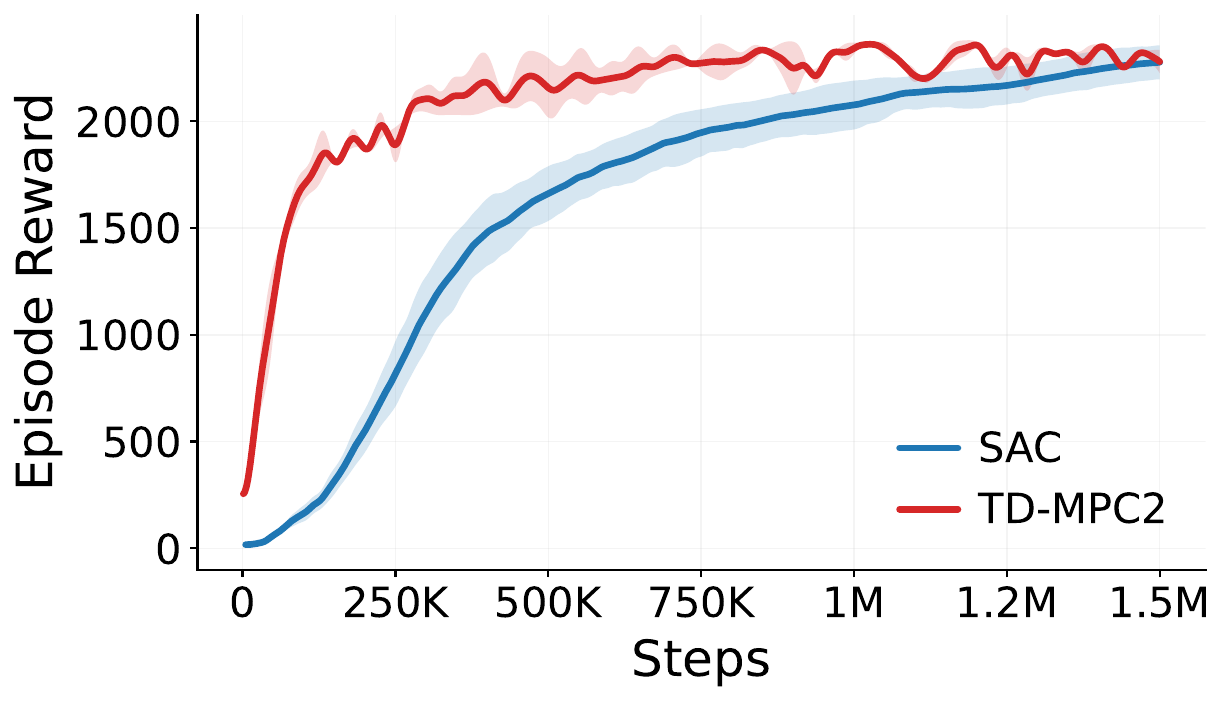}
        \caption{Cheetah Run}
        \label{fig:res_cheetah}
    \end{subfigure}
    \hfill
    \begin{subfigure}[b]{0.32\textwidth}
        \centering
        \includegraphics[width=\textwidth]{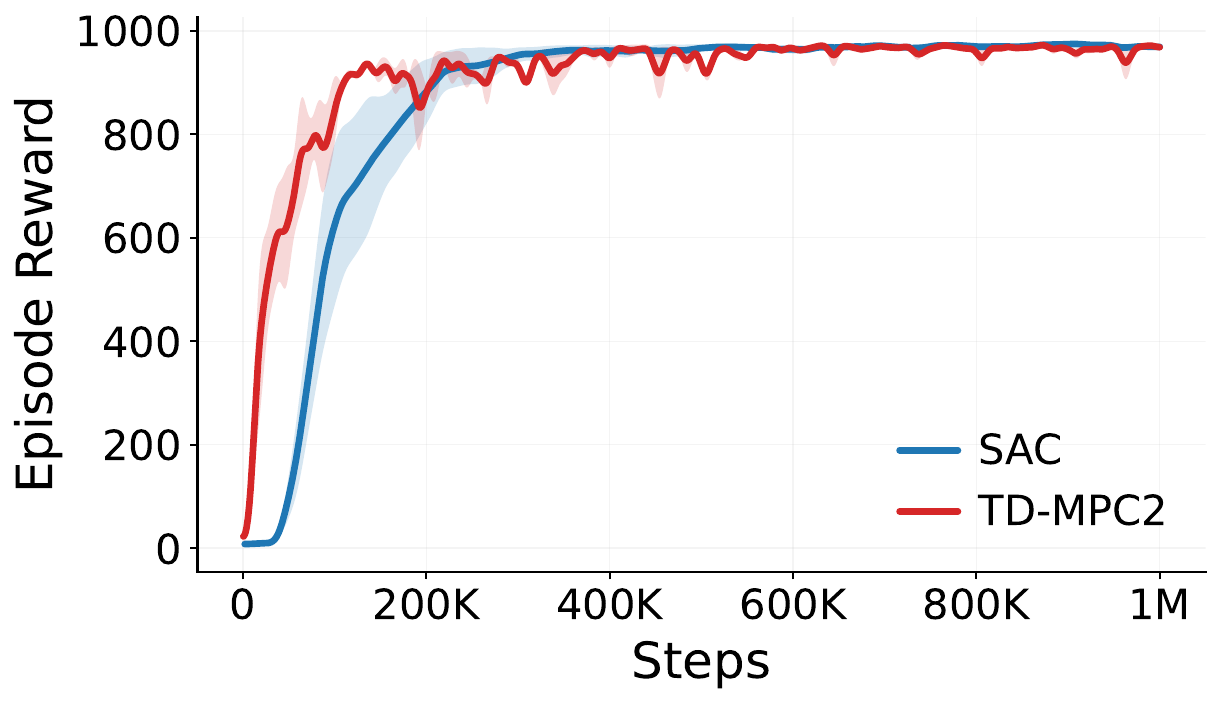}
        \caption{Reacher Hard}
        \label{fig:res_reacher}
    \end{subfigure}
    \hfill
    \begin{subfigure}[b]{0.32\textwidth}
        \centering
        \includegraphics[width=\textwidth]{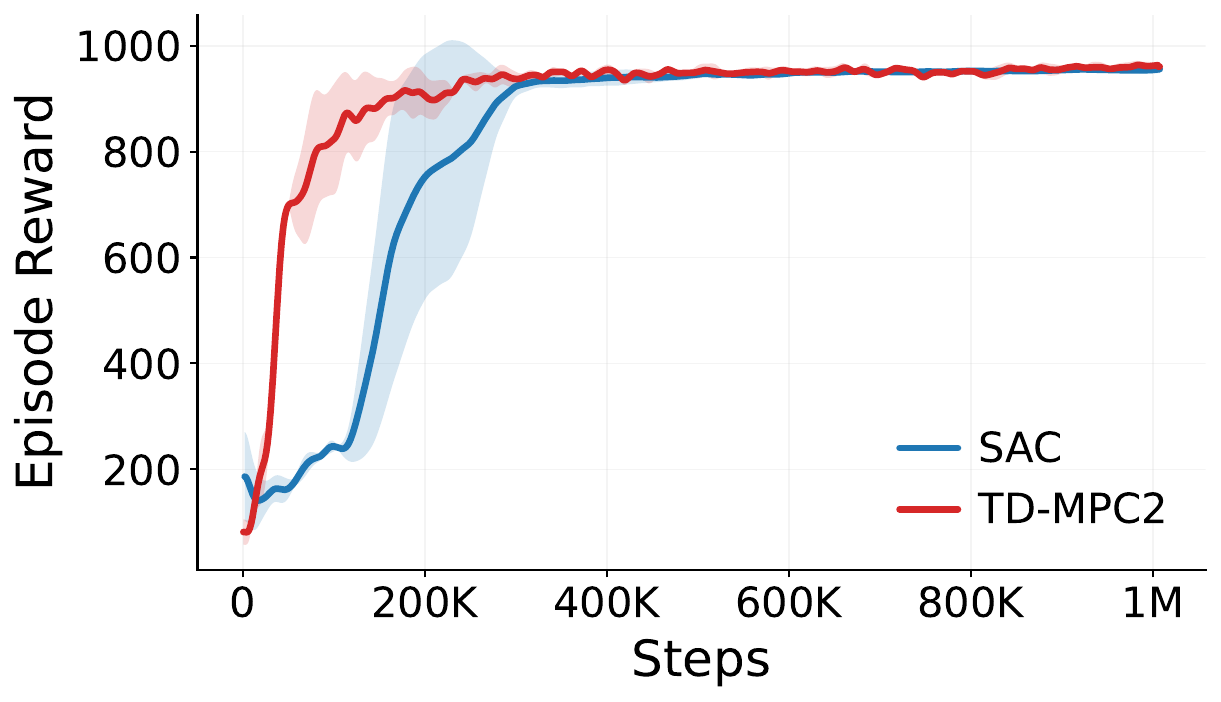}
        \caption{Quadruped Walk}
        \label{fig:res_quadruped}
    \end{subfigure}

    \vspace{0.75em} 

    \begin{subfigure}[b]{0.32\textwidth}
        \centering
        \includegraphics[width=\textwidth]{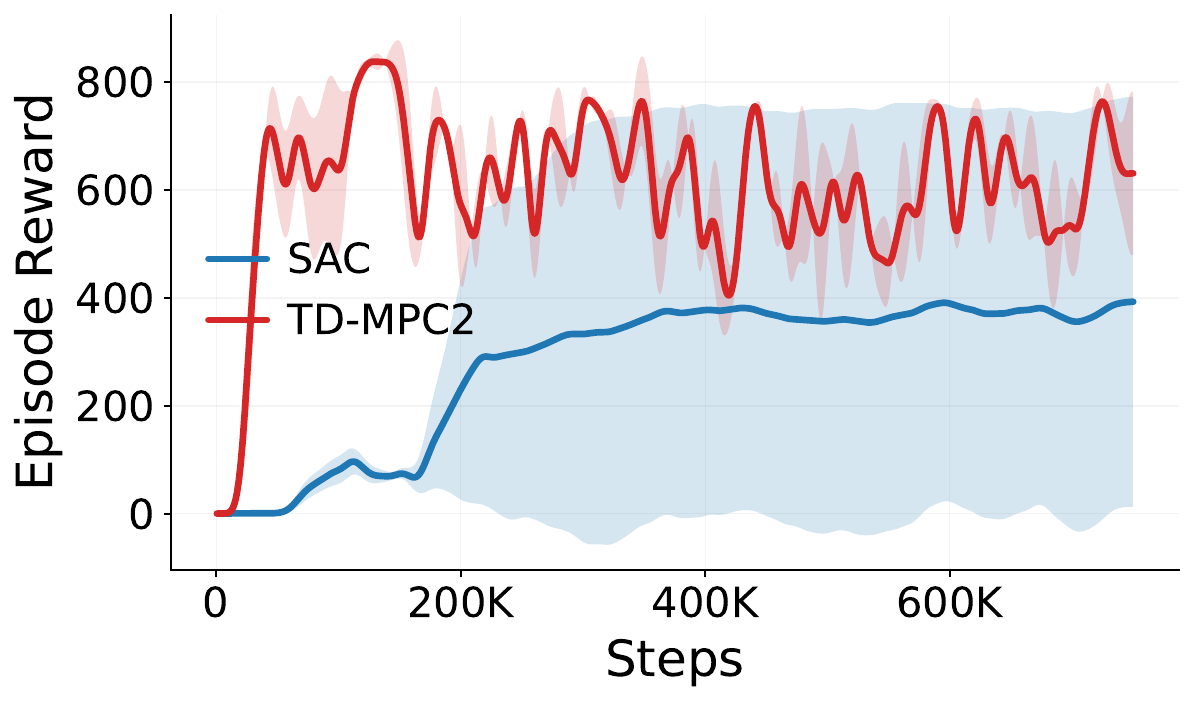}
        \caption{Pendulum Swingup}
        \label{fig:res_pendulum}
    \end{subfigure}
    \hfill
    \begin{subfigure}[b]{0.32\textwidth}
        \centering
        \includegraphics[width=\textwidth]{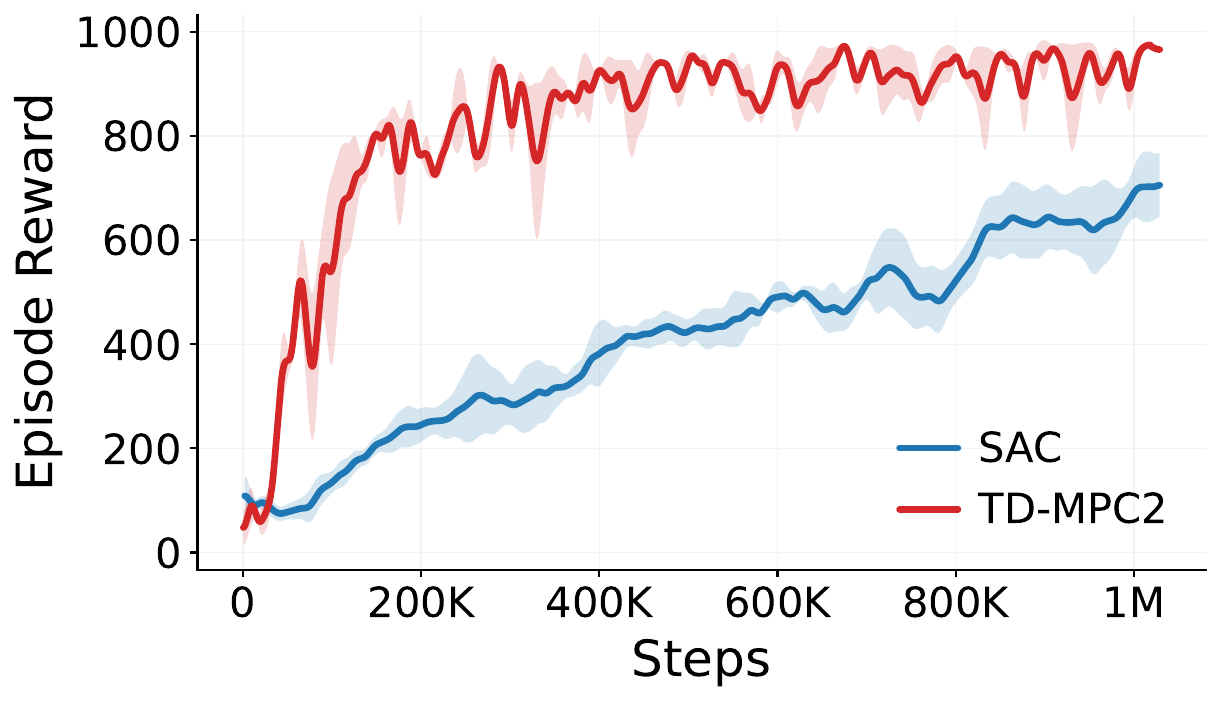}
        \caption{Finger Hard}
        \label{fig:res_finger}
    \end{subfigure}
    \hfill
    \begin{subfigure}[b]{0.32\textwidth}
        \centering
        \includegraphics[width=\textwidth]{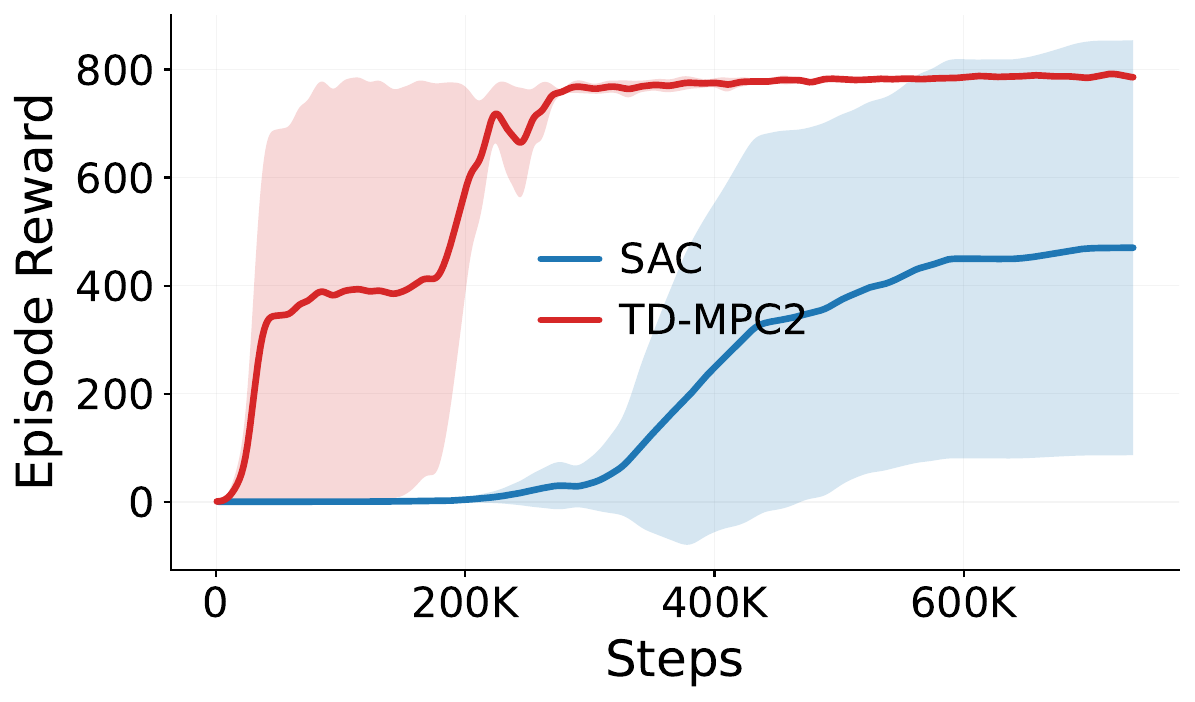}
        \caption{Cartpole Swingup}
        \label{fig:res_ph1}
    \end{subfigure}

    \vspace{0.75em} 

    \begin{subfigure}[b]{0.32\textwidth}
        \centering
        \includegraphics[width=\textwidth]{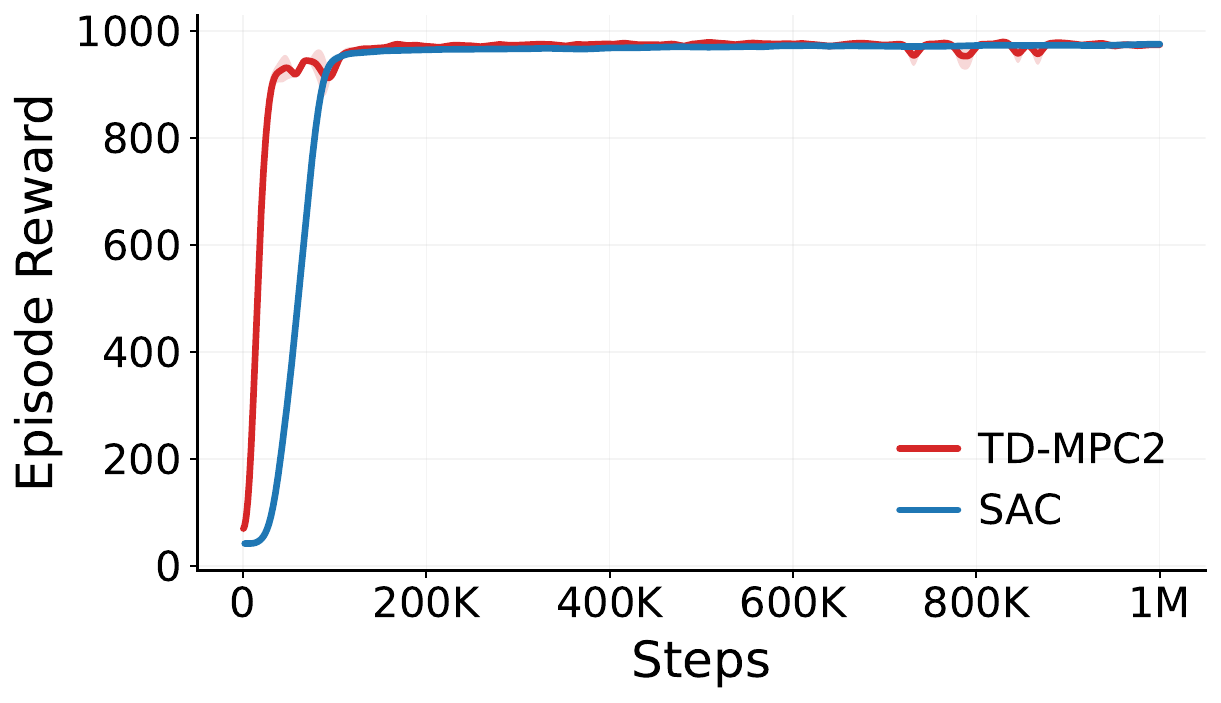}
        \caption{Walker Walk}
        \label{fig:res_ph2}
    \end{subfigure}
    \hfill
    \begin{subfigure}[b]{0.32\textwidth}
        \centering
        \includegraphics[width=\textwidth]{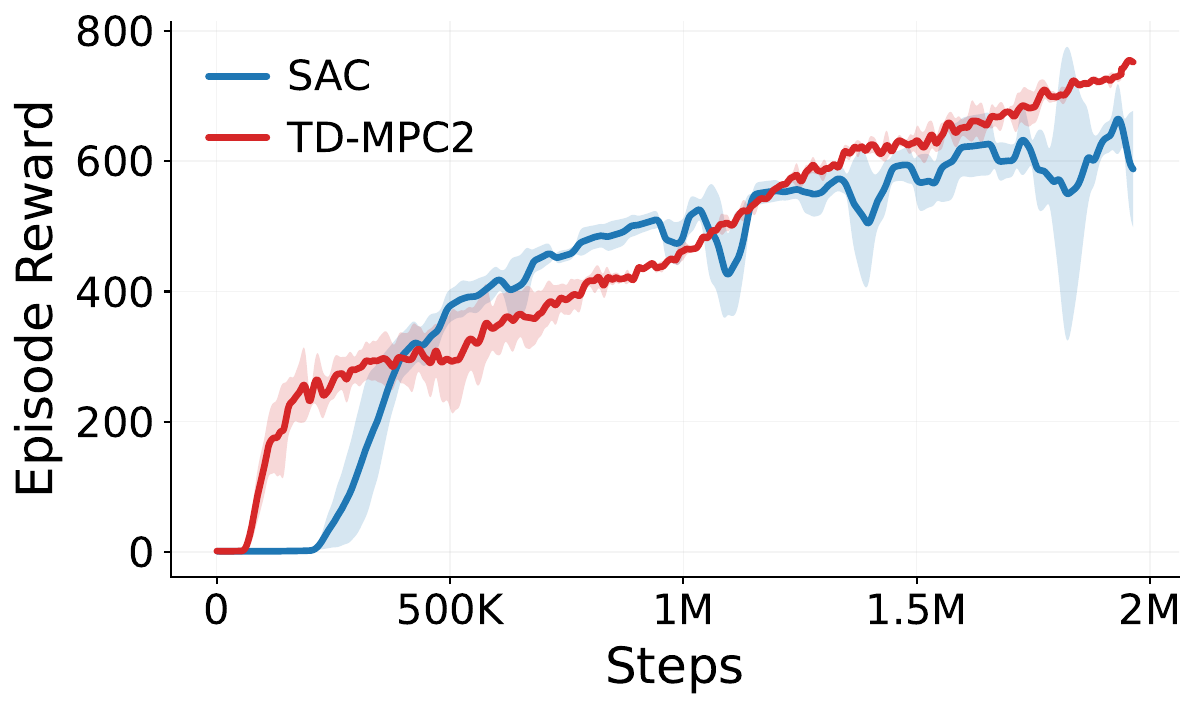}
        \caption{Humanoid Walk}
        \label{fig:res_ph3}
    \end{subfigure}
    \hfill
    \begin{subfigure}[b]{0.32\textwidth}
        \centering
        \includegraphics[width=\textwidth]{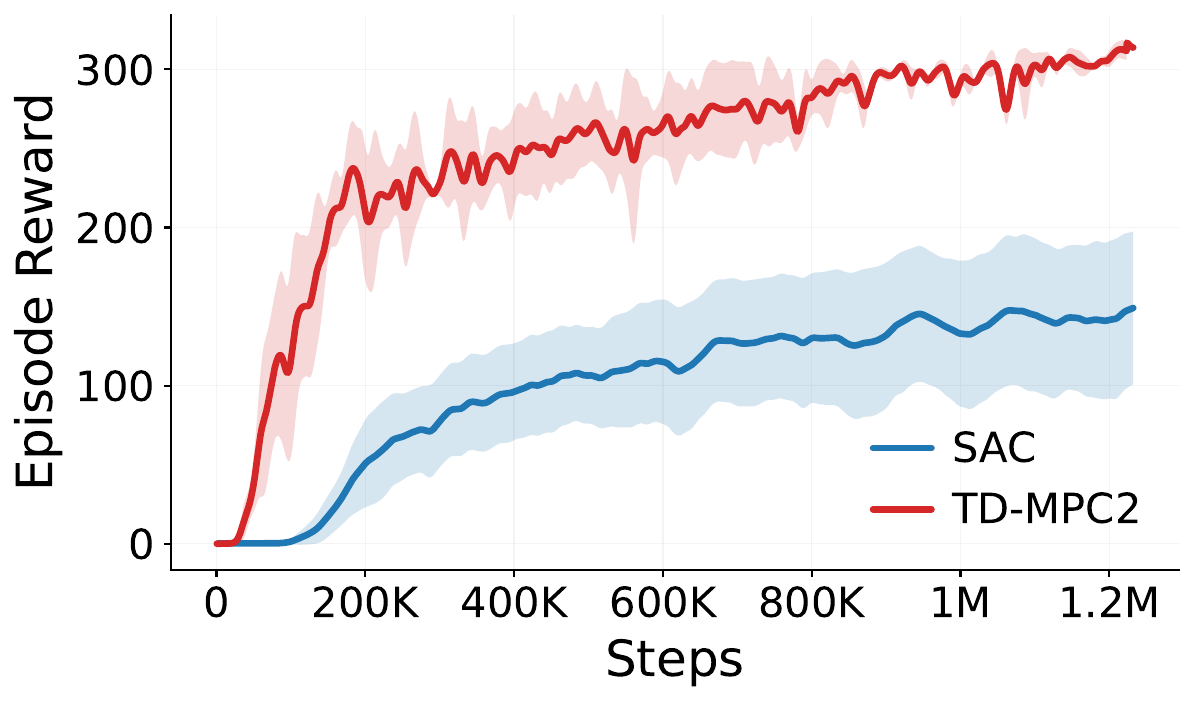}
        \caption{Hopper Hop}
        \label{fig:res_ph4}
    \end{subfigure}

    \caption{Benchmarking \texttt{stable-worldmodel} against standard baselines across diverse environments. Solid lines depict the mean episode reward over 5 random seeds, while shaded areas denote the standard deviation. TD-MPC2 consistently achieves faster convergence in continuous control tasks.}
    \label{fig:benchmark_results_grid}
\end{figure}

\begin{figure}[t]
    \centering
    \includegraphics[width=0.5\textwidth]{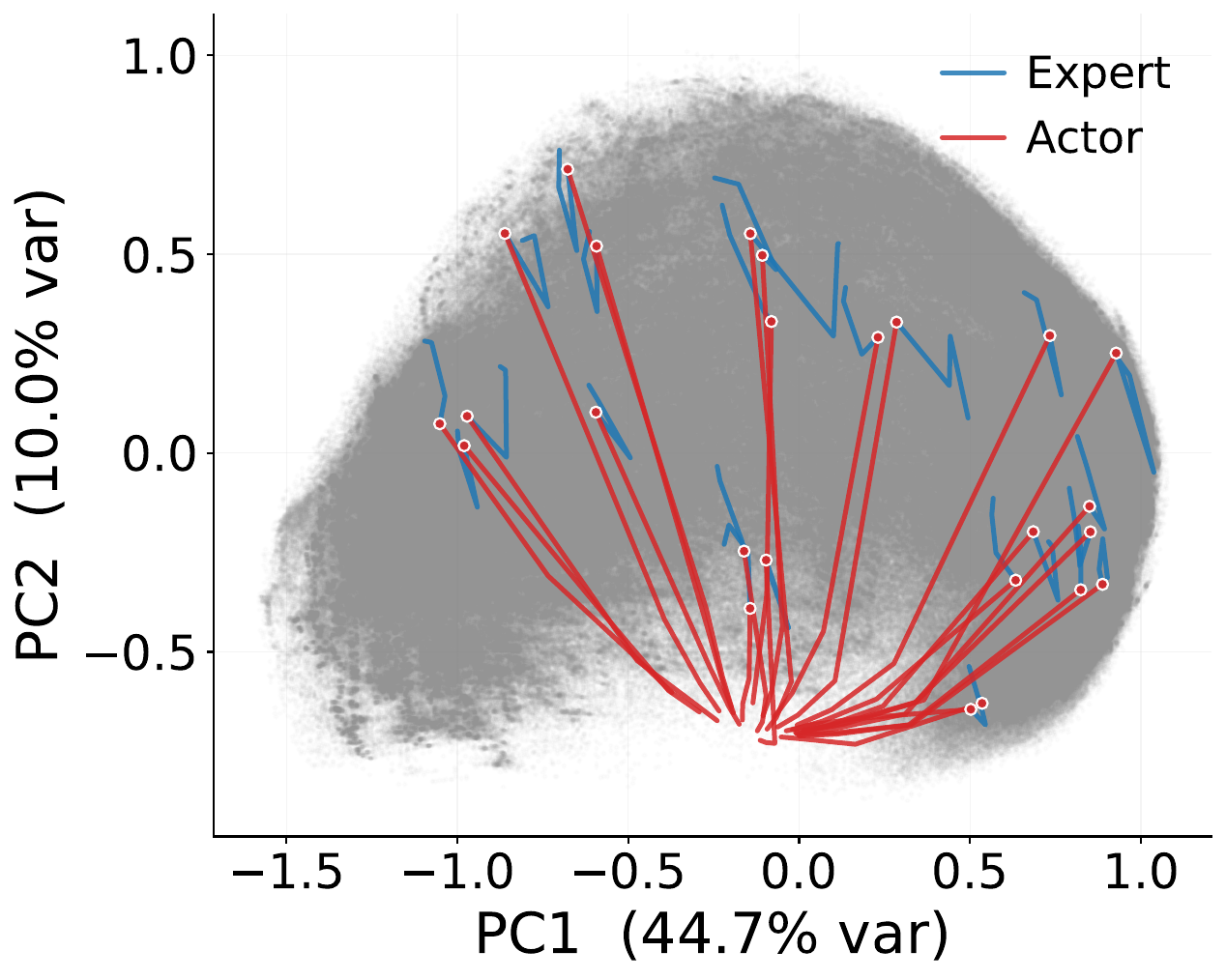
    }
    \caption{PCA projection of TD-MPC2's latent state space on Push-T. Gray points show the distribution of expert states from the offline dataset. Blue trajectories are rolled out from the expert policy and remain within the data manifold, while red trajectories are produced by TD-MPC2's actor and quickly drift outside the support of the training distribution. This visualizes the out-of-distribution drift that drives the planner's poor offline performance reported in Sec.~\ref{sec:exp_planning}.}

    \label{fig:pca-tdmpc}
\end{figure}

Finally, Tab.~\ref{tab:online_training_rewards} and Fig.~\ref{fig:benchmark_results_grid} validate the continuous control solvers in the online regime. While Sec.~\ref{sec:exp_planning} reports that TD-MPC2 struggles to plan effectively offline on Push-T due to out-of-distribution action sampling, these results show strong online performance: across several DeepMind Control Suite tasks, TD-MPC2 reaches competitive asymptotic rewards and matches the sample efficiency of a Soft Actor-Critic (SAC) baseline. This confirms that the algorithm is correctly implemented within \texttt{swm} and that its offline limitations are algorithmic rather than an implementation artifact. Fig.~\ref{fig:pca-tdmpc} illustrates this offline failure mode directly: a PCA projection of the latent state space shows that, unlike expert rollouts which stay within the training distribution, trajectories produced by TD-MPC2's actor quickly drift away from the data manifold, fooling the predictor and leading to poor planning performance.

\begin{table}[t]
    \centering
    \caption{Online training rewards on DeepMind Control Suite tasks. Mean and standard deviation computed over 25 evaluation episodes.}
    \label{tab:online_training_rewards}
    \begin{tabular}{l c c c c}
        \toprule
        \textbf{Environment} & \multicolumn{2}{c}{\textbf{TD-MPC2}} & \multicolumn{2}{c}{\textbf{SAC}} \\
        \cmidrule(lr){2-3} \cmidrule(lr){4-5}
        & \textbf{Mean $\pm$ Std} & \textbf{Min / Max} & \textbf{Mean $\pm$ Std} & \textbf{Min / Max} \\
        \midrule
        cartpole\_swingup  & $\mathbf{767 \pm 1}$ & 766 / 768 & $764 \pm 4$ & 760 / 766 \\
        cheetah\_run       & $\mathbf{2397 \pm 9}$ & 2356 / 2405 & $2369 \pm 20$ & 2308 / 2387 \\
        finger\_turn\_hard & $\mathbf{897 \pm 262}$ & 0 / 1000 & $893 \pm 264$ & 0 / 1000 \\
        hopper\_hop        & $\mathbf{355 \pm 8}$ & 340 / 370 & $265 \pm 16$ & 222 / 284 \\
        humanoid\_walk     & $\mathbf{749 \pm 30}$ & 685 / 800 & $543 \pm 283$ & 2 / 898 \\
        pendulum\_swingup  & $\mathbf{870 \pm 86}$ & 686 / 1000 & $839 \pm 91$ & 698 / 988 \\
        quadruped\_walk    & $956 \pm 24$ & 876 / 990 & $\mathbf{959 \pm 22}$ & 893 / 992 \\
        reacher\_hard      & $969 \pm 12$ & 943 / 997 & $\mathbf{976 \pm 13}$ & 939 / 994 \\
        walker\_walk       & $\mathbf{977 \pm 9}$ & 960 / 990 & $975 \pm 19$ & 893 / 994 \\
        \bottomrule
    \end{tabular}
\end{table}

\section{Research Opportunities}

\texttt{stable-worldmodel} is intended as an open platform for world-model research and evaluation. In the remainder of this section, we outline several promising research directions that \texttt{swm} makes immediately tractable for the community.

\paragraph{Achieving performant zero-shot world models.} Out-of-distribution generalization and zero-shot evaluation are central motivations behind \texttt{swm}, and the platform directly exposes the tools needed to study them: factors of variation, environment wrappers, and a unified baseline suite. As shown in Sec.~\ref{sec:experiments}, current approaches remain brittle even under simple visual perturbations such as color shifts of the scene. Closing this gap on visual and physical robustness is required for safe real-world deployment of world models, yet it will demand substantial further progress. We hope \texttt{swm} lowers the barrier for the community to pursue it.

\paragraph{Unlocking long-horizon planning.} Reliable long-horizon planning remains a major limitation of current world models: as we have shown in Fig.~\ref{fig:combined_pushT}, prediction errors compound over rollouts, and most published evaluations rely on short horizons sampled from in-distribution trajectories. \texttt{swm} ships with a broad set of planning solvers and reference baselines, together with goal sampling at controllable temporal offsets, making it straightforward to stress-test models well beyond the short, expert-policy regimes used today. We hope this would motivate research on horizon scaling, hierarchical, and multi-step
planning.

\paragraph{Scaling World Models.} Scaling laws have driven much of the recent progress in language modeling~\citep{hoffmann2022training}. Whether analogous regularities and their associated emergent behaviours hold for world models is an open and exciting question. \texttt{swm} is designed with this regime in mind: its high-throughput Lance-based data layer alleviates the I/O bottlenecks that typically starve accelerators during multimodal training, as shown in Fig.~\ref{fig:pusht-data}, while its standardized collection and evaluation pipelines make it straightforward to compare models across data, parameter, and compute scales under identical protocols. We hope this turns scaling studies of world models into a concrete research agenda.


\newpage

\end{document}